\RequirePackage{fix-cm}
\documentclass[twocolumn]{svjour3}          
\smartqed  

\usepackage[cmex10]{amsmath}
\usepackage{amsfonts}
\usepackage{amssymb}
\usepackage{bm}
\usepackage{graphicx}
\usepackage{graphics}
\usepackage[applemac]{inputenc}
\usepackage{cite}
\usepackage{mdwtab}
\usepackage{subfigure}
\usepackage{color}
\usepackage{xspace}
\usepackage{url}
\usepackage{tabularx}
\usepackage{epstopdf}
\usepackage{adjustbox}
\usepackage{multirow}

\usepackage{color}
\usepackage{booktabs}
\usepackage{bbding}
\usepackage{pifont}
\usepackage{array}
\usepackage[percent]{overpic}
\usepackage{caption}
\usepackage{enumerate}

\usepackage{natbib} 

\usepackage{appendix}

\newcommand{\cmark}{\ding{51}}%
\newcommand{\xmark}{\ding{55}}%

\newcommand*{\eg}{\emph{e.g.}\@\xspace}
\newcommand*{\ie}{\emph{i.e.}\@\xspace}
\newcommand*{\cf}{\emph{c.f.}\@\xspace}

\newcommand*{\wrt}{\emph{w.r.t}\@\xspace}

\makeatother

\newcommand{\Fig}{Fig.\xspace}

\newcommand{\Sec}{Sec.\xspace}
\newcommand{\Tab}{Tab.\xspace}

\newcolumntype{C}[1]{>{\centering}m{#1}}
\newcolumntype{L}[1]{>{\raggedright\let\newline\\\arraybackslash\hspace{0pt}}m{#1}}
\newcolumntype{C}[1]{>{\centering\let\newline\\\arraybackslash\hspace{0pt}}m{#1}}
\newcolumntype{R}[1]{>{\raggedleft\let\newline\\\arraybackslash\hspace{0pt}}m{#1}}

\newcommand{\MOTChallenge}{{\it MOTChallenge}\xspace}
\newcommand{\MOTNEW}{{\it MOT16}\xspace}
\newcommand{\MOTOLD}{{\it MOT15}\xspace}
\newcommand{\MOTTWENTY}{{\it MOT20}\xspace}
\newcommand{\MOTLAST}{{\it MOT17}\xspace}

\newcommand{\MOTLASTDET}{{\it MOT17Det}\xspace}

\newcommand{\dismeas}{d}			
\newcommand{\simthresh}{t_d}  		

\newcommand{\PAR}[1]{\vskip4pt \noindent {\bf #1~}}

\newcommand{\numtrold}{73\xspace}
\newcommand{\numtrnew}{74\xspace}
\newcommand{\numtrlast}{57\xspace}
\newcommand{\numtroldtotal}{692\xspace}

\newcommand{\numusr}{1833\xspace}

\graphicspath{{figures/},{figures/moving/},{figures/static/},{figures/annotations/}}

\definecolor{darkgreen}{rgb}{0,.75,0}
\definecolor{gray40}{gray}{.40}

\newcommand{\done}[1]{\textcolor{green}{\textbf{DONE}}}

\makeatletter
\renewcommand\subsection{\@startsection{subsection}{2}{\z@}%
    {-21dd plus-8pt minus-4pt}{10.5dd}
     {\normalsize\bfseries\boldmath\upshape}}
\makeatother

\begin{document}

\title{MOTChallenge: A Benchmark for Single-Camera Multiple Target Tracking}

\author{
    Patrick Dendorfer \and
    Aljo\u{s}a O\u{s}ep \and
    Anton Milan \and
    Konrad Schindler \and
    Daniel Cremers \and
    Ian Reid \and
    Stefan Roth \and
    Laura Leal-Taix\'{e}
}


\institute{
    P. Dendorfer, A. O\u{s}ep,  D. Cremers, L. Leal-Taix\'{e}\at
    Technical University Munich, Germany \\
    \email{\{patrick.dendorfer, aljosa.osep, cremers, leal.taixe\} @tum.de}  
    \and
    A. Milan \at
    Amazon Research, Germany \\
    (Work done prior to joining Amazon)\\
    \email{antmila@amazon.com}
    \and
    K. Schindler \at
    ETH Z\"{u}rich, Switzerland \\
    \email{schindler@ethz.ch}            
    \and
    I. Reid \at
    The University of Adelaide, Australia \\
    \email{ian.reid@adelaide.edu.au}            
    \and
    S. Roth \at
    Technical University of Darmstadt, Germany \\
    \email{stefan.roth@vision.tu-darmstadt.de}
}

\date{Received: date / Accepted: date}

\maketitle

\begin{abstract}
Standardized benchmarks have been crucial in pushing the performance of computer vision algorithms, especially since the advent of deep learning.
Although leaderboards should not be over-claimed, they often
provide the most objective measure of performance and are therefore important guides
for research.

We present \MOTChallenge, a benchmark for single-camera Multiple Object Tracking (MOT) launched in late 2014, 
to collect existing and new data and create a framework
for the standardized evaluation of multiple object tracking methods. 
The benchmark is focused on multiple people tracking, since
pedestrians are by far the most studied object in the tracking community, with applications ranging from robot navigation to self-driving cars.
This paper collects the first three releases of the benchmark: (i) \MOTOLD , along with numerous state-of-the-art results that were submitted in the last years, (ii) \MOTNEW, which contains new challenging videos, and (iii) \MOTLAST, that extends \MOTNEW sequences with more precise labels and evaluates tracking performance on three different object detectors.
The second and third release not only offers a significant increase
in the number of labeled boxes, but also provide labels for multiple object classes beside pedestrians, as well as the level of visibility for every single object of interest.

We finally provide a categorization of state-of-the-art trackers and a broad error analysis. This will help newcomers understand the related work and research trends in the MOT community, and hopefully shed some light on potential future research directions.

\end{abstract}

\section{Introduction}
\label{sec:introduction}

Evaluating and comparing single-camera multi-target tracking methods is not trivial 
for numerous reasons~\citep{Milan13cvprw}. 
Firstly, unlike for other tasks, such as image denoising, the ground truth, \ie, the perfect solution one aims to achieve, is
difficult to define clearly. Partially visible, occluded, or cropped targets, 
reflections in mirrors or windows, and objects that very closely resemble targets all impose intrinsic ambiguities, such that even humans may not agree on one particular ideal solution. Secondly, many different
evaluation metrics with free parameters and ambiguous definitions often lead to conflicting quantitative results across the literature. Finally, the lack of pre-defined test and training data makes it difficult to compare
different methods fairly.

Even though multi-target tracking is a crucial problem in scene understanding, 
until recently it still lacked large-scale benchmarks to provide a fair comparison between 
tracking methods. 
Typically, methods are tuned for each sequence, reaching over 90\% accuracy in well-known sequences like 
PETS~\citep{Ferryman10avss}. Nonetheless, the real challenge for a tracking system is to be able to perform well on a variety of sequences with different level of crowdedness, camera motion, illumination, etc., without overfitting the set of parameters
to a specific video sequence. 

To address this issue, we released the \MOTChallenge benchmark in 2014, which consisted of
three main components:
(1) a (re-)collection of publicly available and new datasets, (2) a centralized evaluation method, and (3) 
an infrastructure that allows for crowdsourcing of new data, new 
evaluation methods and even new annotations. 
The first release of the dataset named \MOTOLD consists of 11 sequences for training and 11 for testing, with a total of 11286 frames or 996 seconds of video. 3D information was also provided for 4 of those sequences.
Pre-computed object detections, annotations (only for the training sequences), and a common evaluation method for 
all datasets were provided to all participants, which allowed for all results to be compared fairly. 


Since October 2014, over $1,\!000$ methods have been publicly tested on the \MOTChallenge benchmark, and over \numusr users have registered, see Fig.~\ref{fig:motchallenge}. 
In particular, $760$ methods have been tested on \MOTOLD, $1,\!017$ on \MOTNEW and $692$ on \MOTLAST; $132$, $213$ and $190$ (respectively) were published on the public leaderboard. 
This established \MOTChallenge as the first standardized large-scale tracking benchmark for single-camera multiple people tracking.

Despite its success, the first tracking benchmark, \MOTOLD, was lacking in a few aspects:
\begin{itemize}
    \item The annotation protocol was not consistent across all sequences since some of the ground truth was collected from various online sources;
    \item the distribution of crowd density was not balanced for training and test sequences;
    \item some of the sequences were well-known (\eg, PETS09-S2L1) and methods were overfitted to them, which made them not ideal for testing
    purposes;
    \item the provided public detections did not show good performance on the benchmark, which made some participants switch to other pedestrian detectors.
\end{itemize}  

To resolve the aforementioned shortcomings, we introduced the second benchmark, \MOTNEW. It consists of a set of 14 sequences with crowded scenarios, recorded from different viewpoints, with/without camera motion, and it covers a diverse set of weather and illumination conditions.
Most importantly, the annotations for \emph{all} sequences were carried out by qualified researchers from scratch following a strict protocol and finally double-checked to ensure a high annotation accuracy. 
In addition to pedestrians, we also annotated classes such as vehicles, sitting people, and occluding objects. 
With this fine-grained level of annotation, it was possible to accurately compute the degree of occlusion and cropping of all bounding boxes, which was also provided with the benchmark. 

For the third release, \MOTLAST, we (1) further improved the annotation consistency over the sequences\footnote{We thank the numerous contributors and users of MOTChallenge that pointed us to issues with annotations.} and (2) proposed a new evaluation protocol with public detections. 
In \MOTLAST, we provided 3 sets of public detections, obtained using three different object detectors. Participants were required to evaluate their trackers using all three detections sets, and results were then averaged to obtain the final score. The main idea behind this new protocol was to establish the robustness of the trackers when fed with detections of different quality.
Besides, we released a separate subset for evaluating object detectors, \MOTLASTDET. 

In this work, we categorize and analyze \numtrold published trackers that have been evaluated on \MOTOLD, \numtrnew trackers on \MOTNEW, and \numtrlast on \MOTLAST\vspace{0.005cm}
\footnote{In this paper, we only consider published trackers that were on the leaderboard on April 17th, 2020, and used the provided set of public detections. 
For this analysis, we focused on peer-reviewed methods, \ie, published at a conference or a journal, and excluded entries for which we could not find corresponding publications due to lack of information provided by the authors.}.
Having results on such a large number of sequences allows us to perform a thorough analysis of trends in tracking, currently best-performing methods, and special failure cases. 
We aim to shed some light on potential research directions for the near future in order to further improve tracking performance.

\noindent In summary, this paper has two main goals:
\begin{itemize}
    \item To present the \MOTChallenge benchmark for a fair evaluation of multi-target tracking methods, along with its first releases: \MOTOLD, \MOTNEW, and \MOTLAST;
    \item to analyze the performance of \numtrold state-of-the-art trackers on \MOTOLD, \numtrnew trackers on \MOTNEW, and \numtrlast on \MOTLAST to analyze trends in MOT over the years. We analyze the main weaknesses of current trackers and discuss promising research directions for the community to advance the field of multi-target tracking. 
\end{itemize}

The benchmark with all datasets, ground truth, detections, submitted results, current ranking and submission guidelines can be found at:
\begin{center}
    \url{http://www.motchallenge.net/}
\end{center}

\section{Related work}
\label{sec:related-work}

\noindent{\bf Benchmarks and challenges.}
In the recent past, the computer vision community has developed centralized benchmarks for numerous tasks including object detection~\citep{Everingham15ijcv}, pedestrian detection \citep{caltechpedestrians}, 3D reconstruction~\citep{Seitz06CVPR}, 
optical flow~\citep{Baker:11:ijcv,Geiger12kitti}, visual odometry~\citep{Geiger12kitti}, single-object short-term tracking~\citep{VOC2014}, and stereo estimation~\citep{Scharstein02IJCV,Geiger12kitti}.     
Despite potential pitfalls of such benchmarks~\citep{Torralba11CVPR}, they 
have proven to be extremely helpful to advance the state of the art in the respective area.

For single-camera multiple target tracking, in contrast, there has been very limited work on standardizing quantitative
evaluation. One of the few exceptions is the well-known PETS dataset~\citep{Ferryman10avss} addressing primarily surveillance
applications. The 2009 version consists of 3 subsets S: S1 targeting person count and density estimation, S2 targeting people tracking, and S3 targeting flow analysis and event recognition. 
The simplest sequence for tracking (S2L1) consists of a scene with few pedestrians, 
and for that sequence, state-of-the-art methods perform extremely well 
with accuracies of over 90\% given a good set of initial detections~\citep{Milan14pami, Henriques11iccv, Zamir12eccv}. 
Therefore, methods started to focus on tracking objects in the most challenging sequence, \ie, with the highest crowd
density, but hardly ever on the complete dataset. Even for this widely used benchmark, we observe that tracking results are commonly obtained inconsistently, involving using different subsets of the available data, inconsistent model training that is often prone to overfitting, varying evaluation scripts, and different detection inputs.
Results are thus not easily comparable.  
Hence, the questions that arise are: (i) are these sequences already too easy for current tracking methods?, (ii) do methods simply overfit?, and (iii) are existing methods poorly evaluated?

The PETS team organizes a workshop approximately once a year to which researchers can submit their results, and methods are evaluated under the same conditions. Although this is indeed a fair comparison, the fact that submissions are evaluated only once a year means that
the use of this benchmark for high impact conferences like ICCV or CVPR remains challenging.
Furthermore, the sequences tend to be focused only on surveillance scenarios and lately on specific tasks such as vessel tracking. Surveillance videos have a low frame rate, fixed camera viewpoint, and low pedestrian density. The ambition of \MOTChallenge is to tackle more general scenarios including varying viewpoints, illumination conditions, different frame rates, and levels of crowdedness.

A well-established and useful way of organizing datasets is through standardized challenges. These are usually in the form of web servers that host the data and through which results are uploaded by the users.
Results are then evaluated in a centralized way by the server and afterward presented online to the public, making a comparison with any other method immediately possible.

There are several datasets organized in this fashion: the Labeled Faces in the Wild~\citep{huangtech2007} for unconstrained face recognition, the 
PASCAL VOC~\citep{Everingham15ijcv} for object detection and the ImageNet large scale visual recognition challenge~\citep{Russakovsky15ijcv}. 

The KITTI benchmark~\citep{Geiger12kitti} was introduced for challenges in autonomous driving, which includes stereo/flow, odometry, road and lane estimation, object detection, and orientation estimation, as well as tracking. 
Some of the sequences include crowded pedestrian crossings, making the dataset quite challenging, but the camera position is located at a fixed height for all sequences. 

Another work that is worth mentioning is~\citep{alahicvpr2014}, in which the authors collected a large amount of data containing 42 million pedestrian trajectories. Since annotation of such a large collection of data is infeasible, they use a denser set of cameras to create the ``ground-truth'' trajectories. Though we do not aim at collecting such a large amount of data, the goal of our benchmark is somewhat similar: to push research in tracking forward by generalizing the test data to a larger set that is 
highly variable and hard to overfit. 

DETRAC~\citep{detracarxiv2015} is a benchmark for vehicle tracking, following a similar submission system to the one we proposed with \MOTChallenge. This benchmark consists of a total of 100 sequences, 60\% of which are used for training. 
Sequences are recorded from a high viewpoint (surveillance scenarios) with the goal of vehicle tracking.  

\bigskip


\noindent{\bf Evaluation.} 
A critical question with any dataset is how to measure the performance of the algorithms. 
In the case of multiple object tracking, the CLEAR-MOT metrics \citep{Stiefelhagen06CLE} have emerged as the standard measures.
By measuring the intersection over union of bounding boxes and matching those from ground-truth annotations and results, measures of accuracy and precision can be computed. 
Precision measures how well the persons are localized, while accuracy evaluates how many distinct errors such as missed targets, ghost trajectories, or identity switches are made.

Alternatively, trajectory-based measures by \citet{Wu06CVPR} evaluate how many trajectories were mostly tracked, mostly lost, and partially tracked, relative to the track lengths. These are mainly used to assess track coverage. 
The IDF1 metric~\citep{ristani16ECCV} was introduced for MOT evaluation in a multi-camera setting. Since then it has been adopted for evaluation in the standard single-camera setting in our benchmark. Contrary to MOTA, the ground truth to predictions mapping is established at the level of entire tracks instead of on frame by frame level, and therefore, measures long-term tracking quality. In~\Sec~\ref{sec:experiments} we report IDF1 performance in conjunction with MOTA. A detailed discussion on the measures can be found in \Sec~\ref{sec:evaluation}. 

A key parameter in both families of metrics is the intersection over union threshold which determines whether a predicted bounding box was matched to an annotation.
It is fairly common to observe methods compared under different thresholds, varying from 25\% to 50\%. There are often many other variables and implementation details that differ between evaluation scripts, which may affect results significantly. 
Furthermore, the evaluation script is not the only factor. Recently, a thorough study~\citep{Mathias14ECCV} on face detection benchmarks showed that annotation policies vary greatly among  datasets. 
For example, bounding boxes can be defined tightly around the object, or more loosely to account for pose variations. 
The size of the bounding box can greatly affect results since the intersection over union depends directly on it. 

Standardized benchmarks are preferable for comparing methods in a fair and principled way.
Using the same ground-truth data and evaluation methodology is the only way to guarantee that the only part being evaluated is the tracking method that delivers the results. This is the main goal of the \MOTChallenge benchmark.

\section{History of MOTChallenge}
\begin{figure}[t]
\centering
{\includegraphics[width=0.99\linewidth]{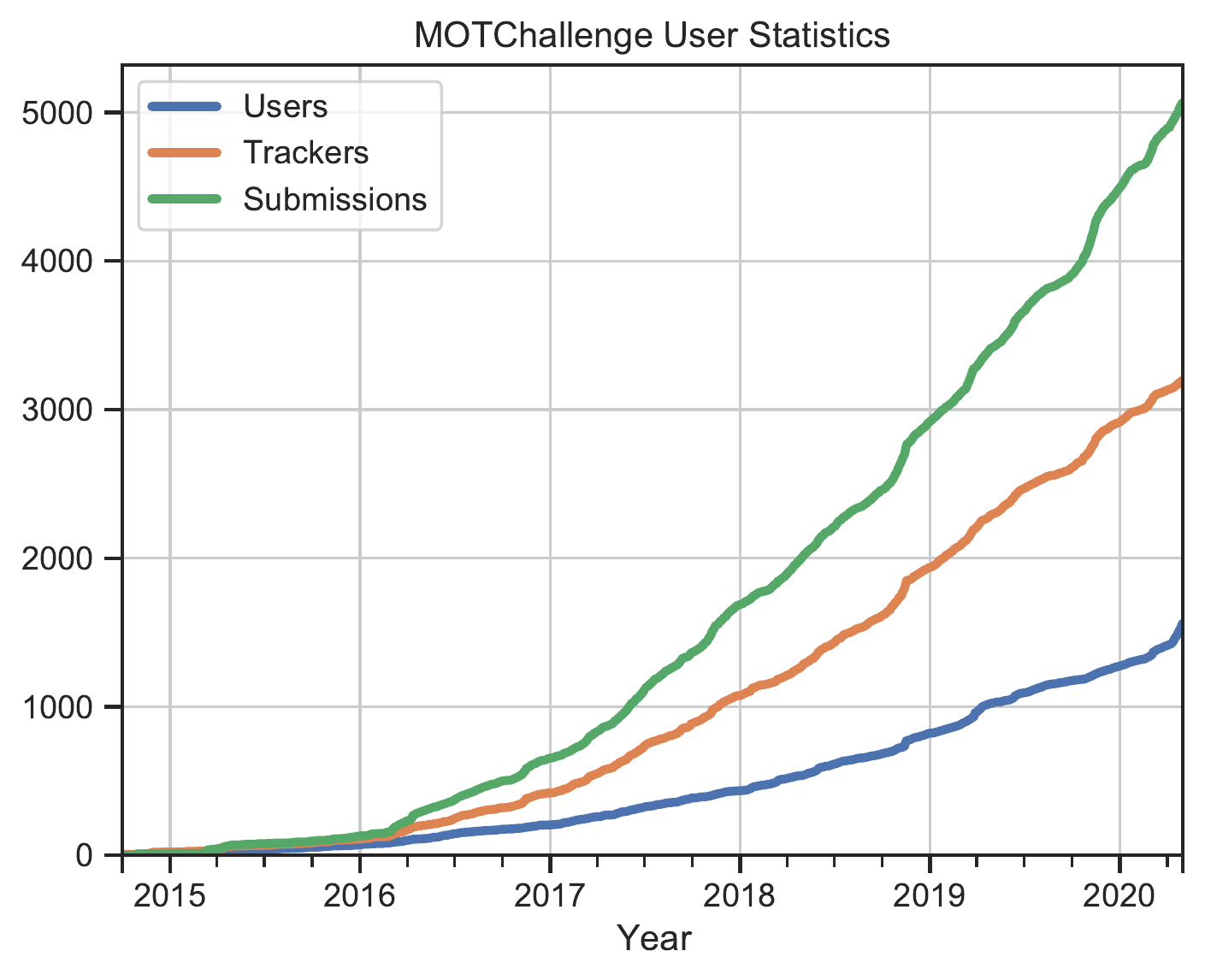}}
\caption{Evolution of \MOTChallenge submissions, number of users registered and trackers created.}
\label{fig:motchallenge}
\end{figure}
\begin{figure*}[htbp]
    \centering
    \subfigure[][Detection performance of~\citep{Dollar14pami}]{
    \includegraphics[width=0.24\linewidth]{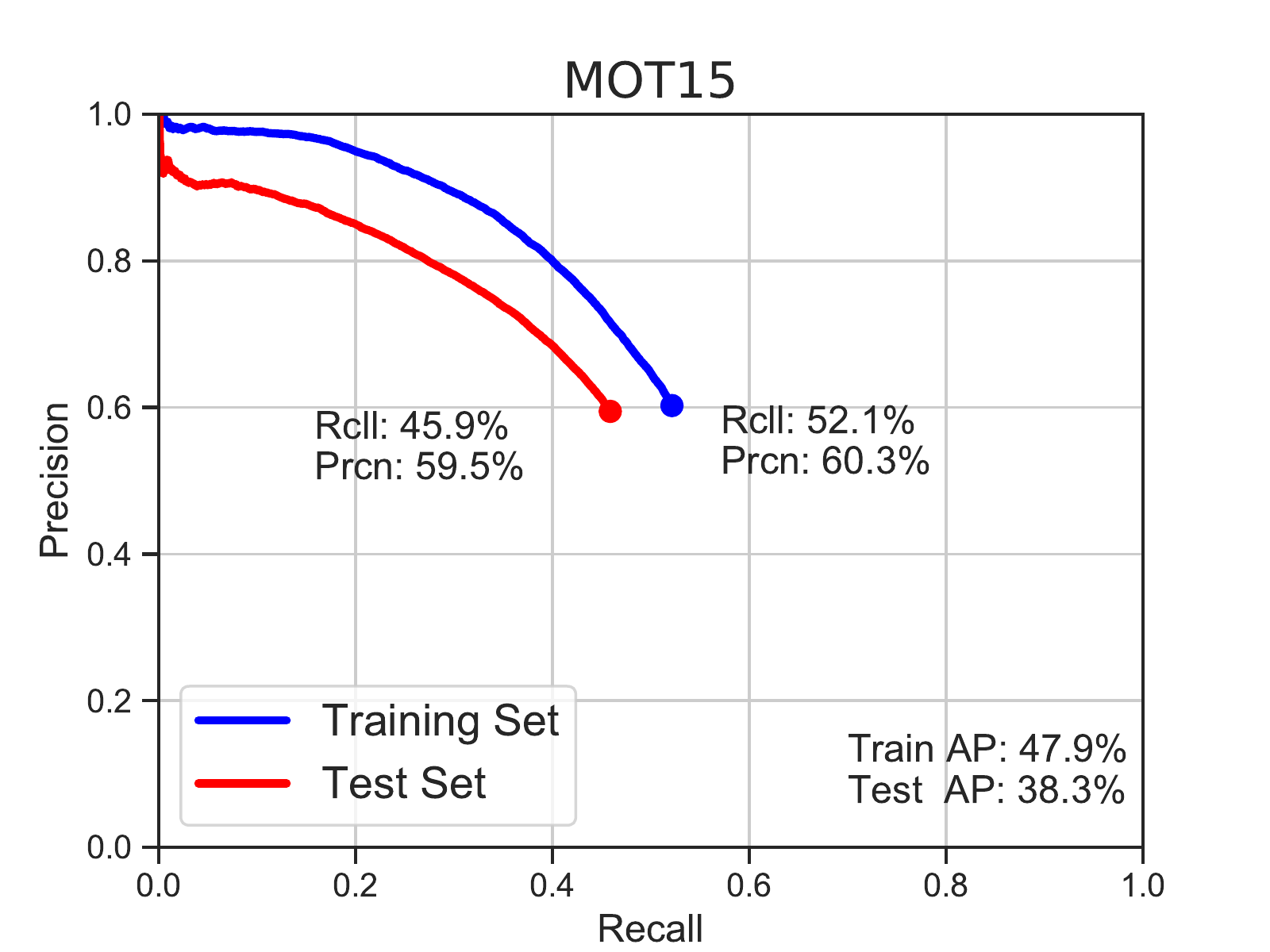}} \subfigure[][ADL-Rundle-8]{
    \includegraphics[width=0.23\linewidth]{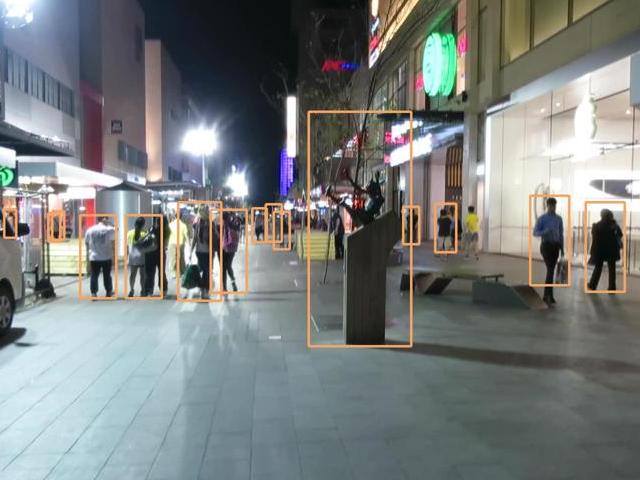}} 
    \subfigure[][Venice-1]{
    \includegraphics[width=0.23\linewidth]{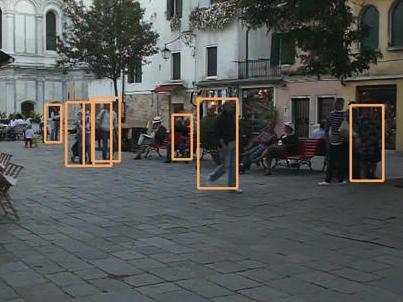}} 
    \subfigure[][KITTI-16]{
    \includegraphics[width=0.23\linewidth]{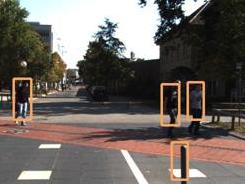}} 
    \caption{(a) The performance of the provided detection bounding boxes evaluated on the training (blue) and the test (red) set. The circle indicates
    the operating point (\ie, the input detection set) for the trackers. (b-d) Exemplar detection results.}
    \label{fig:det-performanceMOT15}
\end{figure*}

The first benchmark was released in October 2014 and it consists of 11 sequences for training and 11 for testing, where the testing sequences have not been available publicly.
We also provided a set of detections and evaluation scripts.
Since its release, \numtroldtotal tracking results were submitted to the benchmark, which has quickly become the standard for evaluating multiple pedestrian tracking methods in high impact conferences such as ICCV, CVPR, and ECCV.
Together with the release of the new data, we organized the 1st Workshop on Benchmarking Multi-Target Tracking (BMTT) in conjunction with the IEEE Winter Conference on Applications of Computer Vision (WACV) in 2015.\footnote{\url{https://motchallenge.net/workshops/bmtt2015/}}

After the success of the first release of sequences, we created a 2016 edition, with 14 longer and more crowded sequences and a more accurate annotation policy which we describe in this manuscript~(Sec.~\ref{sec:anno-rules}). 
For the release of \MOTNEW, we organized the second workshop\footnote{\url{https://motchallenge.net/workshops/bmtt2016/}} in conjunction with the European Conference in Computer Vision (ECCV) in 2016. 

For the third release of our dataset, \MOTLAST, we improved the annotation consistency over the \MOTNEW sequences and provided three public sets of detections, on which trackers need to be evaluated. 
For this release, we organized a Joint Workshop on Tracking and Surveillance in conjunction with the Performance Evaluation of Tracking and Surveillance (PETS)~\citep{Ferryman09pets, Ferryman10avss} workshop and the Conference on Vision and Pattern Recognition (CVPR) in 2017\footnote{\url{https://motchallenge.net/workshops/bmtt-pets2017/}}.

In this paper, we focus on the \MOTOLD, \MOTNEW, and \MOTLAST benchmarks because numerous methods have already submitted their results to these challenges for several years that allow us to analyze these methods and to draw conclusions about research trends in multi-object tracking.

Nonetheless, work continues on the benchmark, with frequent releases of new challenges and datasets. 
The latest pedestrian tracking dataset was first presented at the 4th \MOTChallenge workshop\footnote{\url{https://motchallenge.net/workshops/bmtt2019/}} (CVPR 2019), an ambitious tracking challenge with eight new sequences \citep{dendorfer2019cvpr19}. With the feedback of the workshop the sequences were revised and re-published as the \MOTTWENTY~\citep{dendorfer20arxiv} benchmark. This challenge focuses on very crowded scenes, where the object density can reach up to 246 pedestrians per frame. The diverse sequences show indoor and outdoor scenes, filmed either during day or night. 
With more than $2M$ bounding boxes and $3,\!833$ tracks,  \MOTTWENTY constitutes a new level of complexity and challenges the performance of tracking methods in very dense scenarios. At the time of this article, only 11 submissions for \MOTTWENTY had been received, hence a discussion of the results is not yet significant nor informative, and is left for future work.

The future vision of \MOTChallenge is to establish it as a general platform for benchmarking multi-object tracking, expanding beyond pedestrian tracking. To this end, we recently added a public benchmark for multi-camera 3D zebrafish tracking~\citep{Pedersen_2020_CVPR}, and a benchmark for the large-scale Tracking any Object (TAO) dataset~\citep{Dave:2020:ECCV}. This dataset consists of 2,907 videos, covering 833 classes by 17,287 tracks.  

In Fig.~\ref{fig:motchallenge}, we plot the evolution of the number of users, submissions, and trackers created since \MOTChallenge was released to the public in 2014. Since our 2nd workshop was announced at ECCV, we have experienced steady growth in the number of users as well as submissions.
\section{MOT15 Release}
\label{sec:datasets15}
\newcommand{\thumbwidth}{0.135\linewidth}
\newcommand{\thumbheight}{0.08\linewidth}
\begin{figure*}[ht!]
    \centering
    \includegraphics[width=\thumbwidth,height=\thumbheight]{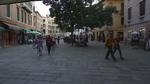}
    \includegraphics[width=\thumbwidth,height=\thumbheight]{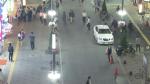}
    \includegraphics[width=\thumbwidth,height=\thumbheight]{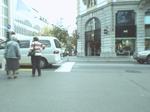}
    \includegraphics[width=\thumbwidth,height=\thumbheight]{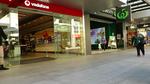}
    \includegraphics[width=\thumbwidth,height=\thumbheight]{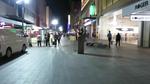}
    \includegraphics[width=\thumbwidth,height=\thumbheight]{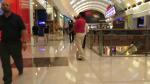}
    \includegraphics[width=\thumbwidth,height=\thumbheight]{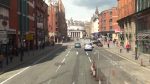}\\[1.em]
    \includegraphics[width=\thumbwidth,height=\thumbheight]{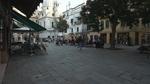}
    \includegraphics[width=\thumbwidth,height=\thumbheight]{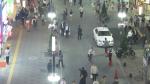}
    \includegraphics[width=\thumbwidth,height=\thumbheight]{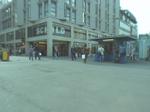}
    \includegraphics[width=\thumbwidth,height=\thumbheight]{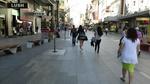}
    \includegraphics[width=\thumbwidth,height=\thumbheight]{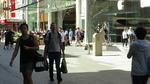}
    \includegraphics[width=\thumbwidth,height=\thumbheight]{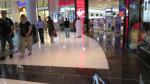}
    \includegraphics[width=\thumbwidth,height=\thumbheight]{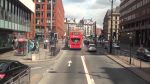}
    \caption{An overview of the \MOTNEW/\MOTLAST dataset. Top: Training sequences. Bottom: test sequences.}
    \label{fig:MOT16/17thumbnails}
\end{figure*}
One of the key aspects of any benchmark is data collection. 
The goal of \MOTChallenge is not only to compile yet another dataset with completely new data but rather to: (1) create a common framework to test tracking methods on, and 
(2) gather existing and new challenging sequences with very different characteristics  (frame rate, pedestrian density, illumination, or point of view) in order to challenge researchers to develop more general tracking methods that can deal with all types of sequences.
In~\Tab~\ref{tab:dataoverview15} of the Appendix we show an overview of the sequences included in the benchmark. 
\subsection{Sequences}

We have compiled a total of 22 sequences that combine different videos from several sources~\citep{Geiger12kitti,  Andriluka:2010:CVPR, Ess08cvpr, Ferryman10avss, Benfold:2011:CVPR} and new data collected from us. 
We use half of the data for training and a half for testing, and the annotations of the testing sequences 
are not released to the public to avoid (over)fitting of  methods to 
specific sequences. 
Note, the test data contains over 10 minutes of footage and 61440 annotated bounding boxes, therefore, it is hard for researchers to over-tune their algorithms on such a large amount of data. 
This is one of the major strengths of the benchmark. 
%

%

We collected 6 new challenging sequences, 4 filmed from a static camera and 2 from a moving camera held at pedestrian's height. %
Three sequences are particularly challenging: a night sequence filmed from a moving camera and two outdoor sequences with a high density of pedestrians. 
The moving camera together with the low illumination creates a lot of motion blur, making this sequence extremely challenging. %
A smaller subset of the benchmark including only these six new sequences were presented at the 1st Workshop on Benchmarking Multi-Target Tracking\footnote{\url{https://motchallenge.net/workshops/bmtt2015/}}, where the top-performing method reached MOTA (tracking accuracy) of only 12.7\%. This confirms the difficulty of the new sequences.\footnote{The challenge results are available at \url{http://motchallenge.net/results/WACV_2015_Challenge/}.}


\subsection{Detections}
\label{sec:detections}

To detect pedestrians in all images of the \MOTOLD edition, we use the object detector of~\citet{Dollar14pami}, which is based on aggregated channel features (ACF). 
We rely on the default parameters and the pedestrian model trained on the INRIA dataset~\citep{Dalal05CVPR}, rescaled with a factor of $0.6$ to enable the detection of smaller pedestrians. 
The detector performance along with three sample frames is depicted in~\Fig~\ref{fig:det-performanceMOT15}, for both the training and the test set of the benchmark. 
Recall does not reach 100\% because of the non-maximum suppression applied.

We cannot (nor necessarily want to) prevent anyone from using a different set of detections.
However, we require that this is noted as part of the tracker's description and is also displayed in the rating table.

\subsection{Weaknesses of \MOTOLD}

By the end of 2015, it was clear that a new release was due for the \MOTChallenge benchmark. The main weaknesses of \MOTOLD were the following:

\begin{itemize}
    \item {\it Annotations:} we collected annotations online for the existing sequences, while we manually annotated the new sequences. Some of the collected annotations were not accurate enough, especially in scenes with moving cameras.
    \item {\it Difficulty:} generally, we wanted to include some well-known sequences, \eg, PETS2009, in the \MOTOLD benchmark. However, these sequences have turned out to be too simple for state-of-the-art trackers why we concluded to create a new and more challenging benchmark.
\end{itemize}

To overcome these weaknesses, we created \MOTNEW, a collection of all-new challenging sequences (including our new sequences from \MOTOLD) and creating
annotations following a more strict protocol (see Sec.~\ref{sec:anno-rules} of the Appendix).
\section{MOT16 and MOT17 Releases}
\label{sec:datasets16}
\begin{figure*}[t]
    \centering
    \subfigure[][DPM v5]{
    \includegraphics[width=0.23\linewidth]{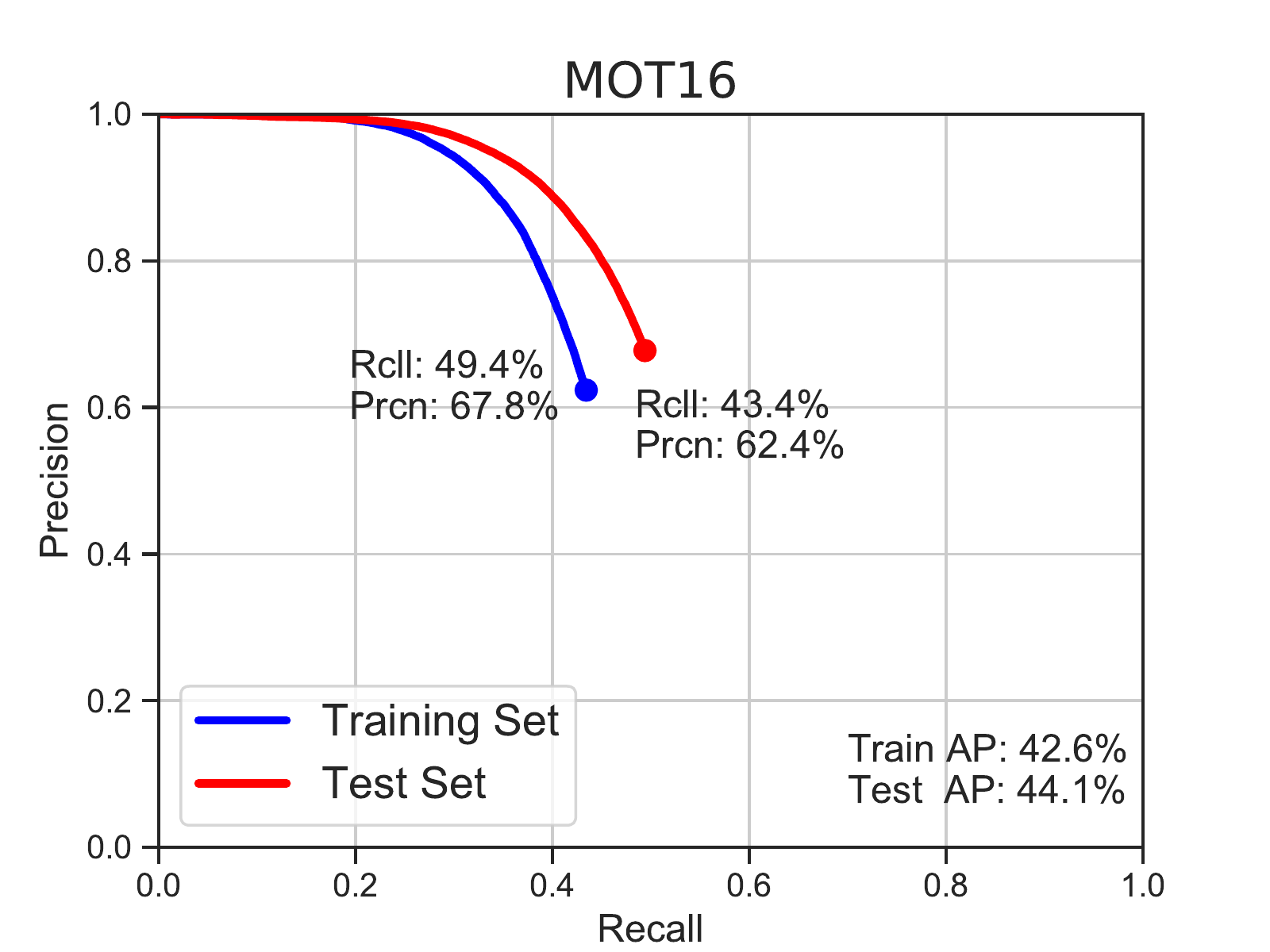}}
    \hfill
     \subfigure[][DPM v5]{
    \includegraphics[width=.23\linewidth]{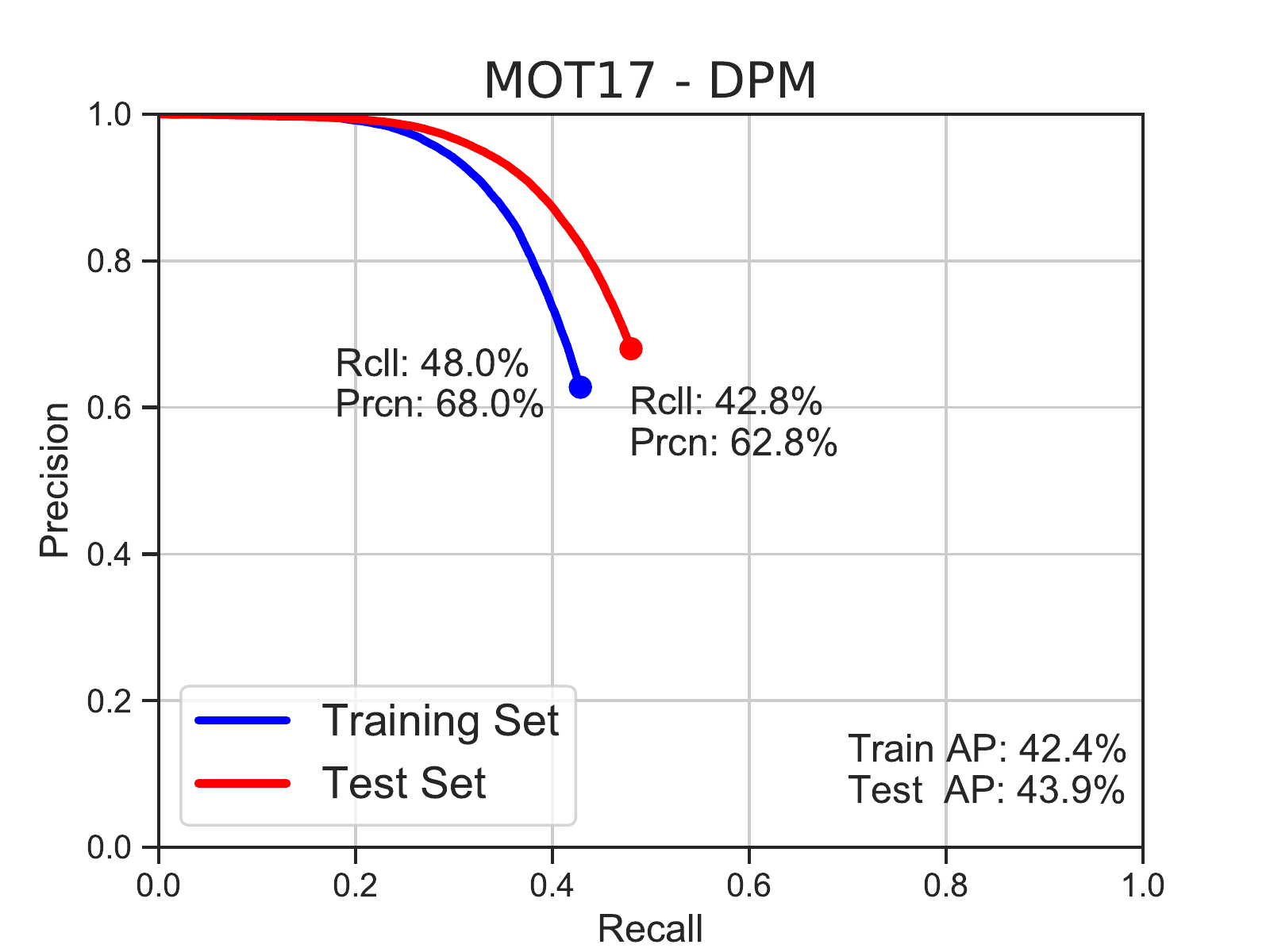}}
    \hfill
     \subfigure[][Faster-RCNN]{
   \includegraphics[width=.23\linewidth]{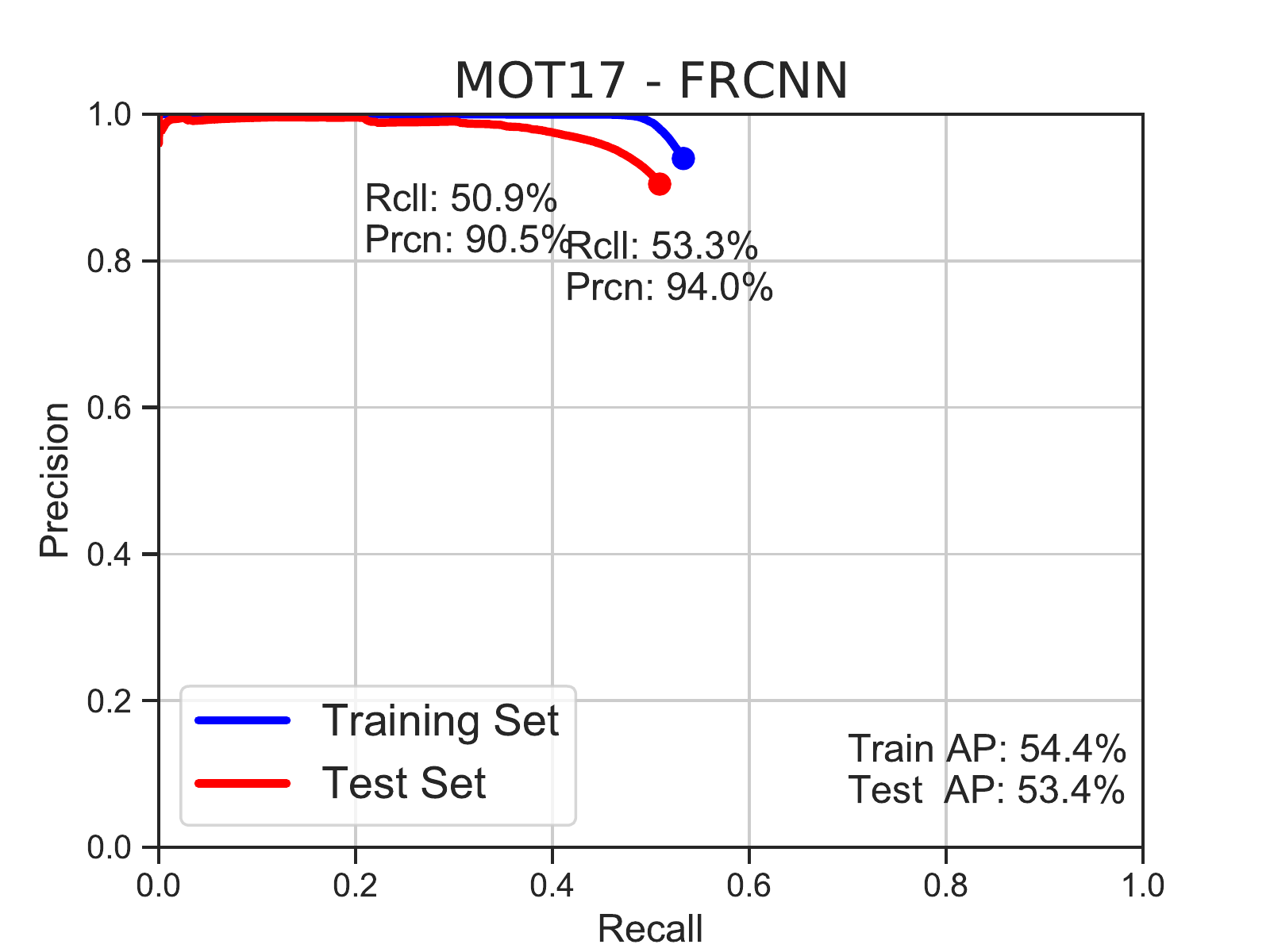}}
   \hfill
    \subfigure[][SDP]{
     \includegraphics[width=.23\linewidth]{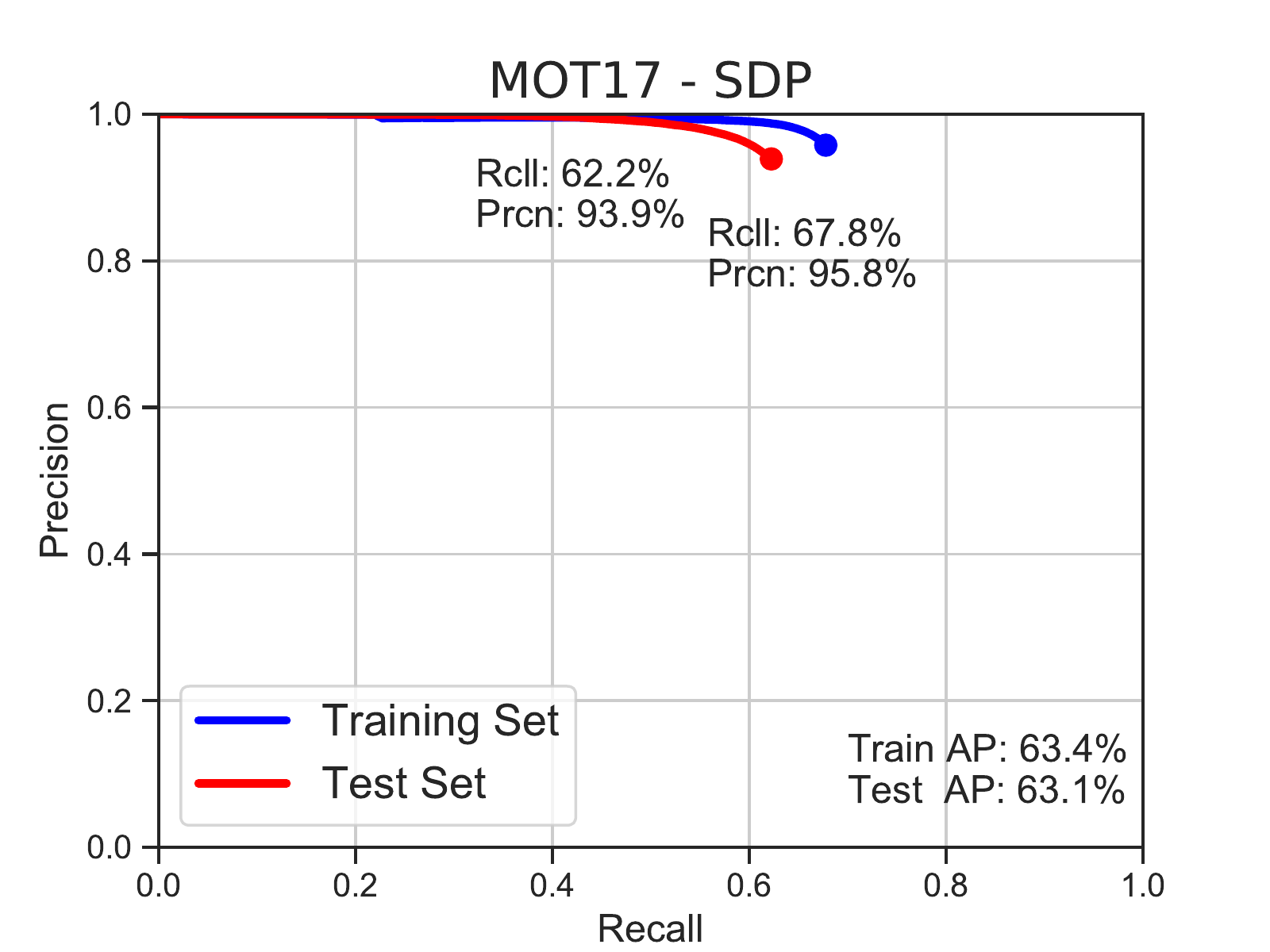}}
    \caption{The performance of three popular pedestrian detectors evaluated on the training (blue) and the test (red) set. The circle indicates the operating point (i.e. the input detection set) for the trackers of \MOTNEW and \MOTLAST.}
    \label{fig:det-performance1617}

\end{figure*}

Our ambition for the release of \MOTNEW was to compile a benchmark with new and more challenging sequences compared to \MOTOLD. 
Figure~\ref{fig:MOT16/17thumbnails} presents an overview of the benchmark training and test sequences (detailed information about the sequences is presented in~\Tab~\ref{tab:dataoverview16} in the Appendix).

\MOTLAST consists of the same sequences as \MOTNEW, but contains two important changes: (i) the annotations are further improved, \ie, increasing the accuracy of the bounding boxes, adding missed pedestrians, annotating additional occluders, following the comments received by many anonymous benchmark users, as well as the second round of sanity checks, (ii) the evaluation system significantly differs from \MOTLAST, including the evaluation of tracking methods using three different detectors in order to show the robustness to varying levels of noisy detections.

\subsection{MOT16 sequences}

We compiled a total of 14 sequences, of which we use half for training and a half for testing. 
The annotations of the testing sequences are not publicly available.
The sequences can be classified according to moving/static camera, viewpoint, and illumination conditions (\Fig~\ref{fig:2Ddata16} in Appendix). 
The new data contains almost 3 times more bounding boxes for training and testing than \MOTOLD. 
Most sequences are filmed in high resolution, and the mean crowd density is 3 times higher when compared to the first benchmark release. 
Hence, the new sequences present a more challenging benchmark than \MOTOLD for the tracking community.

\subsection{Detections}
\label{sec:detections16}
We evaluate several state-of-the-art detectors on our benchmark, and summarize the main findings in~\Fig~\ref{fig:det-performance1617}. 
To evaluate the performance of the detectors for the task of tracking, we evaluate them using all bounding boxes considered for the tracking evaluation, including partially visible or occluded objects. Consequently, the recall and average precision (AP) is lower than the results obtained by evaluating solely on visible objects, as we do for the detection challenge.

\noindent{\bf MOT16 detections.} We first train the deformable part-based model (DPM) v5~\citep{felzenszwalbijcv2006} and find that it outperforms other detectors such as Fast-RNN~\citep{Girshick:2015:ICCV} and ACF~\citep{Dollar14pami} for the task of detecting persons on \MOTNEW. Hence, for that benchmark, we provide DPM detections as public detections.

\noindent{\bf MOT17 detections.}
For the new \MOTLAST release, we use Faster-RCNN~\citep{Ren:2015:NIPS} and a detector with scale-dependent pooling (SDP)~\citep{SDP_detector}, both of which outperform the previous DPM method. 
After a discussion held in one of the \MOTChallenge workshops, we agreed to provide all three detections as public detections, effectively changing the way \MOTChallenge evaluates trackers.
The motivation is to challenge trackers further to be more general and work with detections of varying quality. These detectors have different characteristics, as can be seen in in~\Fig~\ref{fig:det-performance1617}. Hence, a tracker that can work with all three inputs is going to be inherently more robust. 
The evaluation for \MOTLAST is, therefore, set to evaluate the output of trackers on all three detection sets, averaging their performance for the final ranking.
A detailed breakdown of detection bounding box statistics on individual sequences is provided in~\Tab~\ref{tab:detMOT16} in the Appendix.

\section{Evaluation}
\label{sec:evaluation}
\MOTChallenge is also a platform for a fair comparison of state-of-the-art tracking methods. 
By providing authors with standardized ground-truth data, evaluation metrics, scripts, as well as a set of precomputed detections, all methods are compared under the same conditions, thereby isolating the performance of the tracker from other factors.
In the past, a large number of metrics for quantitative evaluation of 
multiple target tracking have been proposed~\citep{Smith05CVPRW,Stiefelhagen06CLE,Bernardin08CLE,Schuhmacher08ACM,Wu06CVPR,LiCVPR2009}. 
Choosing ``the right'' one is largely application dependent and the quest for a unique, general evaluation measure is still ongoing. 
On the one hand, it is desirable to summarize the performance into a single number to enable a direct comparison between methods.
On the other hand, one might want to provide more informative performance estimates by detailing the types of errors the algorithms make, which precludes a clear ranking.

Following a recent trend~\citep{Milan14pami,Bae14CVPR,Wen14CVPR}, we employ three sets of tracking performance measures that have been established in the literature: (i) the frame-to-frame based \emph{CLEAR-MOT} metrics proposed by~\citet{Stiefelhagen06CLE}, (ii) track quality measures proposed by~\citet{Wu06CVPR}, and (iii) trajectory-based IDF1 proposed by~\citet{ristani16ECCV}. 

These evaluation measures give a complementary view on tracking performance. The main representative of CLEAR-MOT measures, Multi-Object Tracking Accuracy (MOTA), is evaluated based on frame-to-frame matching between track predictions and ground truth. It explicitly penalizes identity switches between consecutive frames, thus evaluating tracking performance only locally. This measure tends to put more emphasis on object detection performance compared to temporal continuity. 
In contrast, track quality measures~\citep{Wu06CVPR} and IDF1~\cite{ristani16ECCV}, perform prediction-to-ground-truth matching on a trajectory level and over-emphasize the temporal continuity aspect of the tracking performance. In this section, we first introduce the matching between predicted track and ground-truth annotation before we present the final measures.
All evaluation scripts used in our benchmark are publicly available\footnote{\url{http://motchallenge.net/devkit}}.

\subsection{Multiple Object Tracking Accuracy}
\label{sec:mota}

MOTA summarizes three sources of errors with a single performance measure:
\begin{equation}
    \text{MOTA} = 
    1 - \frac
    {\sum_t{(\text{FN}_t + \text{FP}_t + \text{IDSW}_t})}
    {\sum_t{\text{GT}_t}},
    \label{eq:mota}
\end{equation}
where $t$ is the frame index and $GT$ is the number of ground-truth objects. 
where ${FN}$ are the false negatives, i.e., the number of ground truth objects that were not detected by the method. ${FP}$ are the false positives, i.e., the number of objects that were falsely detected by the method but do not exist in the ground-truth. ${IDSW}$ is the number of identity switches, i.e., how many times a given trajectory changes from one ground-truth object to another. The computation of these values as well as other implementation details of the evaluation tool are detailed in Appendix Sec. \ref{sec:eval_appendix}.
We report the percentage MOTA $(-\infty, 100]$ in our benchmark. 
Note, that MOTA can also be negative in cases where the number of errors made by the tracker exceeds the number of all objects in the scene.

\PAR{Justification.} 
We note that MOTA has been criticized in the literature for not having different sources of errors properly balanced. However, to this day, MOTA is still considered to be the most expressive measure for single-camera MOT evaluation. It was widely adopted for ranking methods in more recent tracking benchmarks, such as PoseTrack~\citep{PoseTrack}, KITTI tracking~\citep{Geiger12kitti}, and the newly released Lyft \citep{Lyft}, Waymo~\citep{Waymo}, and ArgoVerse~\citep{Argoverse} benchmarks. 
We adopt MOTA for ranking, however, we recommend taking alternative evaluation measures~\citep{Wu06CVPR, ristani16ECCV} into the account when assessing the tracker's performance.

\PAR{Robustness.} One incentive behind compiling this benchmark was to reduce dataset bias by keeping the data as diverse as possible. 
The main motivation is to challenge state-of-the-art approaches and analyze their performance in unconstrained environments and on unseen data. 
Our experience shows that most methods can be heavily overfitted on one particular dataset, and may not be general enough to handle an entirely different setting without a major change in parameters or even in the model.

\subsection{Multiple Object Tracking Precision}
\label{sec:motp}

The Multiple Object Tracking Precision is the average dissimilarity between all true positives and their corresponding ground-truth targets. 
For bounding box overlap, this is computed as: 
\begin{equation}
    \text{MOTP} = 
    \frac
    {\sum_{t,i}{d_{t,i}}}
    {\sum_t{c_t}},
    \label{eq:motp}
\end{equation}
where $c_t$ denotes the number of matches in frame $t$ and $d_{t,i}$ is the bounding box overlap of target $i$ with its assigned ground-truth object in frame $t$. 
MOTP thereby gives the average overlap of $\simthresh$ between all correctly matched hypotheses and their respective objects and ranges between 
$\simthresh := 50\%$ and $100\%$.

It is important to point out that MOTP is a 
measure of localisation precision, \emph{not} to be confused with the 
\emph{positive predictive value} or \emph{relevance} in the context of 
precision / recall curves used, \eg, in object detection.

In practice, it quantifies the localization precision of the detector, 
and therefore, it provides little information about the actual performance of the tracker.

\subsection{Identification Precision, Identification Recall, and F1 Score}
\label{sec:idf1}

CLEAR-MOT evaluation measures provide event-based tracking assessment. 
In contrast, the IDF1 measure~\citep{ristani16ECCV} is an identity-based measure that emphasizes the track identity preservation capability over the entire sequence. 
In this case, the predictions-to-ground-truth mapping is established by solving a bipartite matching problem, connecting pairs with the largest temporal overlap. 
After the matching is established, we can compute the number of True Positive IDs (IDTP), False Negative IDs (IDFN), and False Positive IDs (IDFP), that generalise the concept of per-frame TPs, FNs and FPs to tracks. Based on these quantities, we can express the Identification Precision (IDP) as:
\begin{equation}
IDP = \frac{IDTP}{IDTP + IDFP},
\end{equation} 
and Identification Recall (IDR) as: 
\begin{equation}
IDR = \frac{IDTP}{IDTP + IDFN}.
\end{equation}
Note that IDP and IDR are the fraction of computed (ground-truth) detections that are correctly identified. IDF1 is then expressed as a ratio of correctly identified detections over the average number of ground-truth and computed detections and balances identification precision and recall through their harmonic mean:
\begin{equation}
IDF1 = \frac{2 \cdot IDTP}{2 \cdot IDTP + IDFP + IDFN}.
\end{equation}

\subsection{Track quality measures}
\label{sec:track-measures}

The final measures that we report on our benchmark are qualitative, and evaluate the percentage of the ground-truth trajectory that is recovered by a tracking algorithm.
Each ground-truth trajectory can be consequently classified as mostly tracked (MT), partially tracked (PT), and mostly lost (ML). 
As defined in~\citep{Wu06CVPR}, a target is mostly tracked if it is successfully tracked for at least $80\%$ of its life span, and considered lost in case it is covered for less than $20\%$ of its total length.
The remaining tracks are considered to be partially tracked. 
A higher number of MT and a few ML is desirable. 
Note, that it is irrelevant for this measure whether the ID remains the same throughout the track. 
We report MT and ML as a ratio of mostly tracked and mostly lost targets to the total number of ground-truth trajectories.

In certain situations, one might be interested in obtaining long, persistent tracks without trajectory gaps. 
To that end, the number of track fragmentations (FM) counts how many times a ground-truth trajectory is interrupted (untracked). 
A fragmentation event happens each time a trajectory changes its status from tracked to untracked and is resumed at a later point. 
Similarly to the ID switch ratio (\cf~\Sec~\ref{sec:tracker-assignment}), we also provide the relative number of fragmentations as FM / Recall.

\begin{figure*}[htbp]

 \centering
 \begin{minipage}{0.49\linewidth}
    \includegraphics[width=0.9\linewidth]{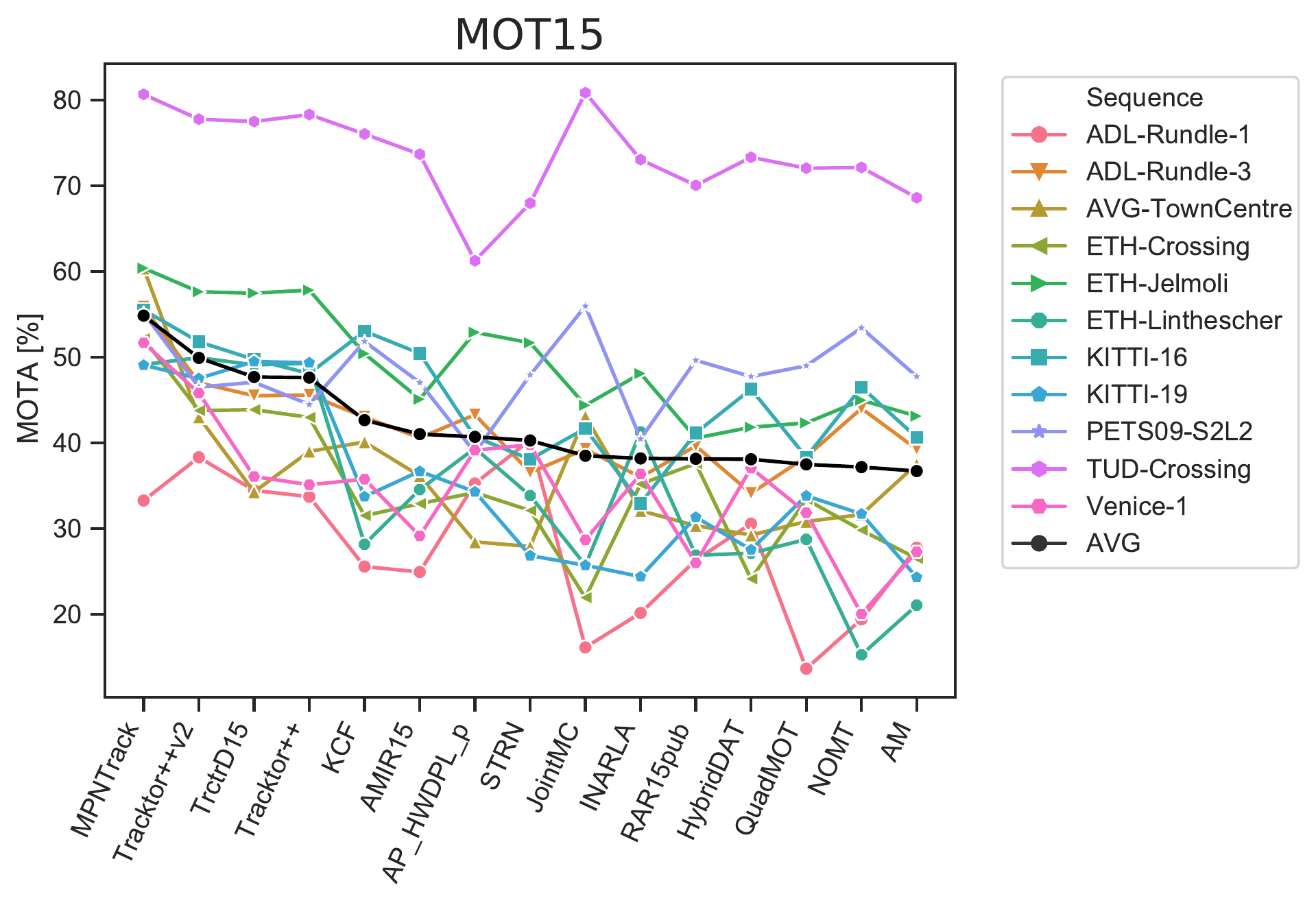}
    \end{minipage}
    \hfill    
    \begin{minipage}{0.49\linewidth}
      \includegraphics[width=0.9\linewidth]{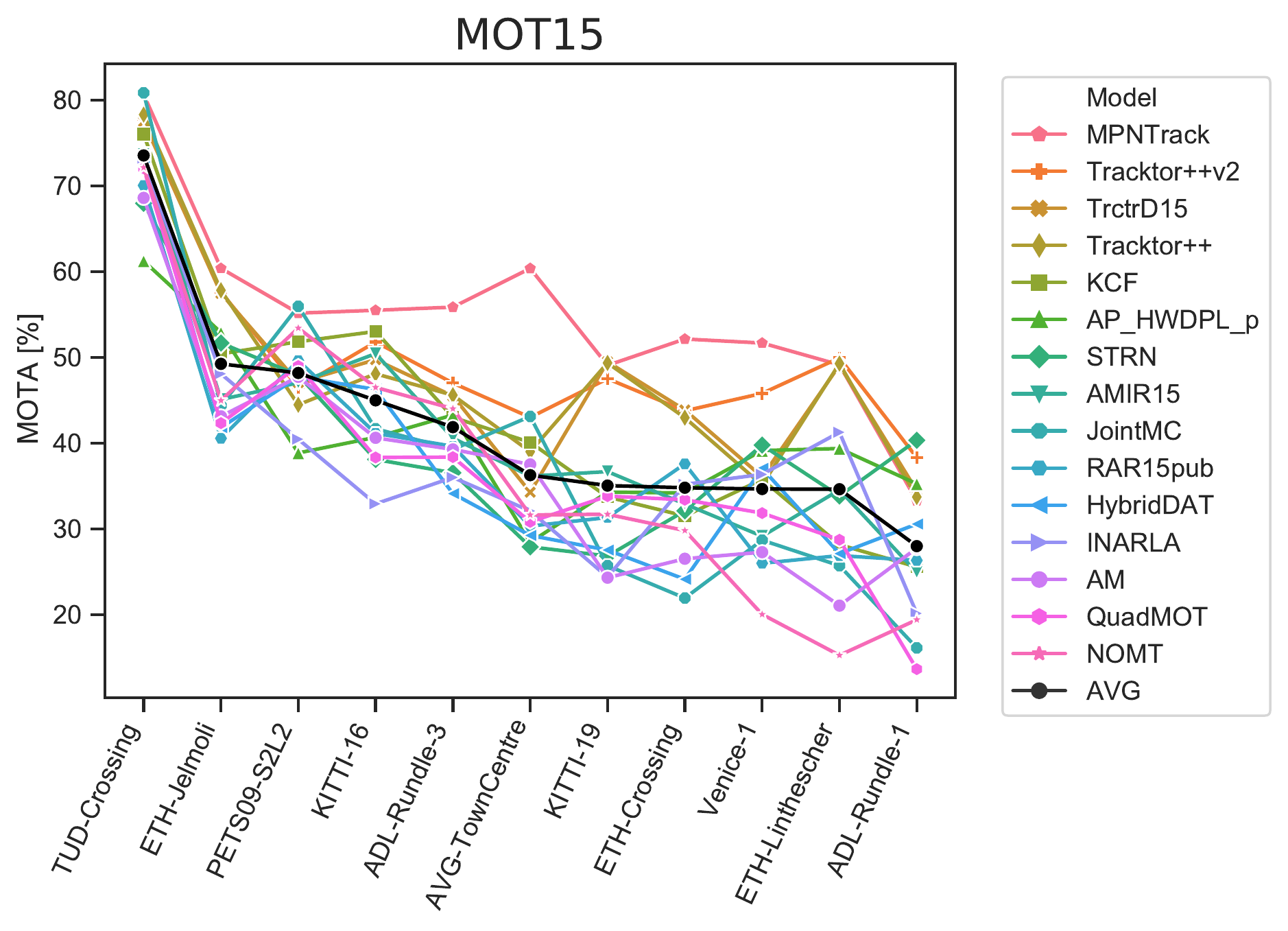}
      \end{minipage}

    \begin{minipage}{0.49\linewidth}
    \includegraphics[width=0.9\linewidth]{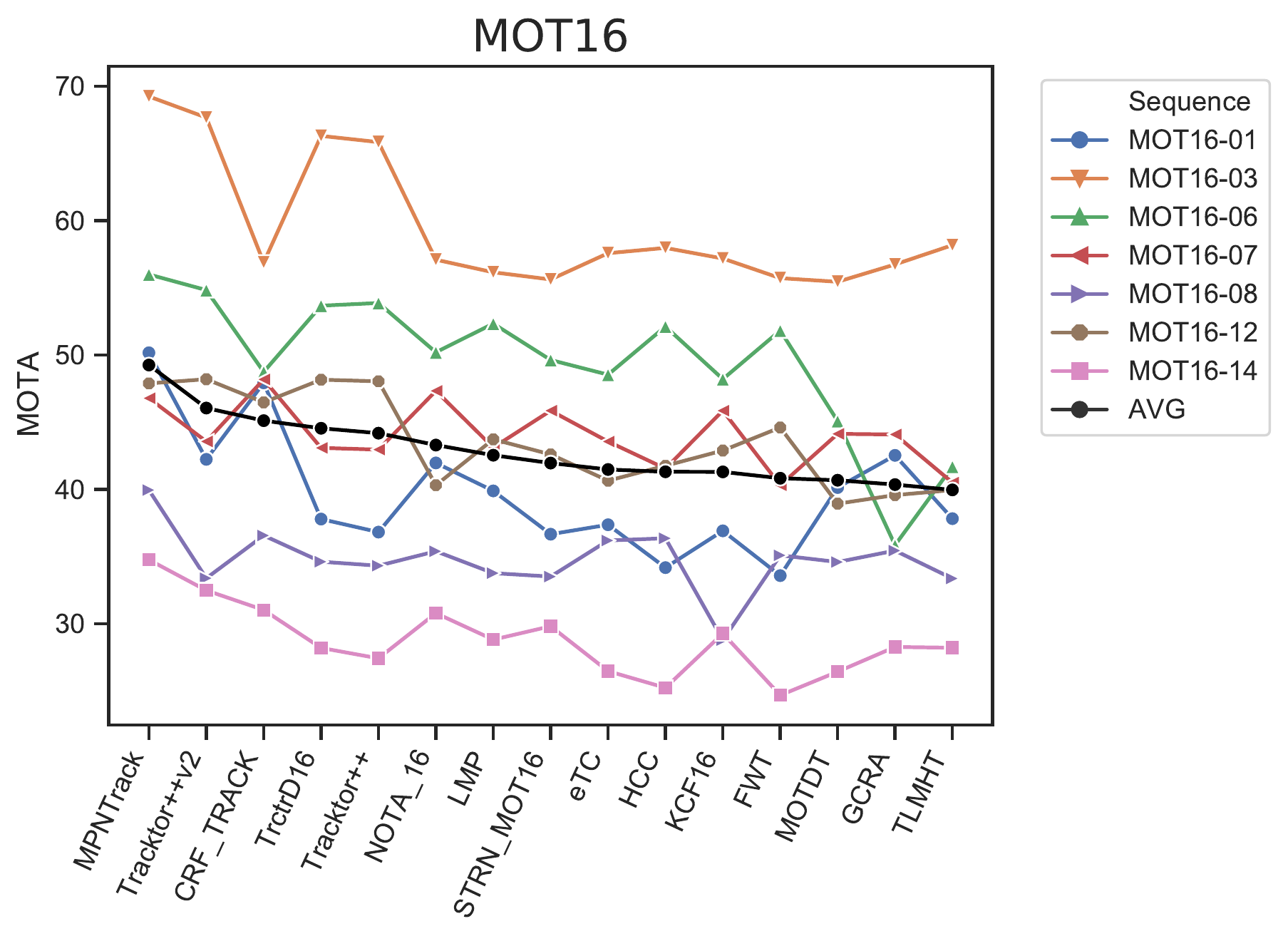}
    \end{minipage}
    \hfill
    \begin{minipage}{0.49\linewidth}
  \includegraphics[width=0.9\linewidth]{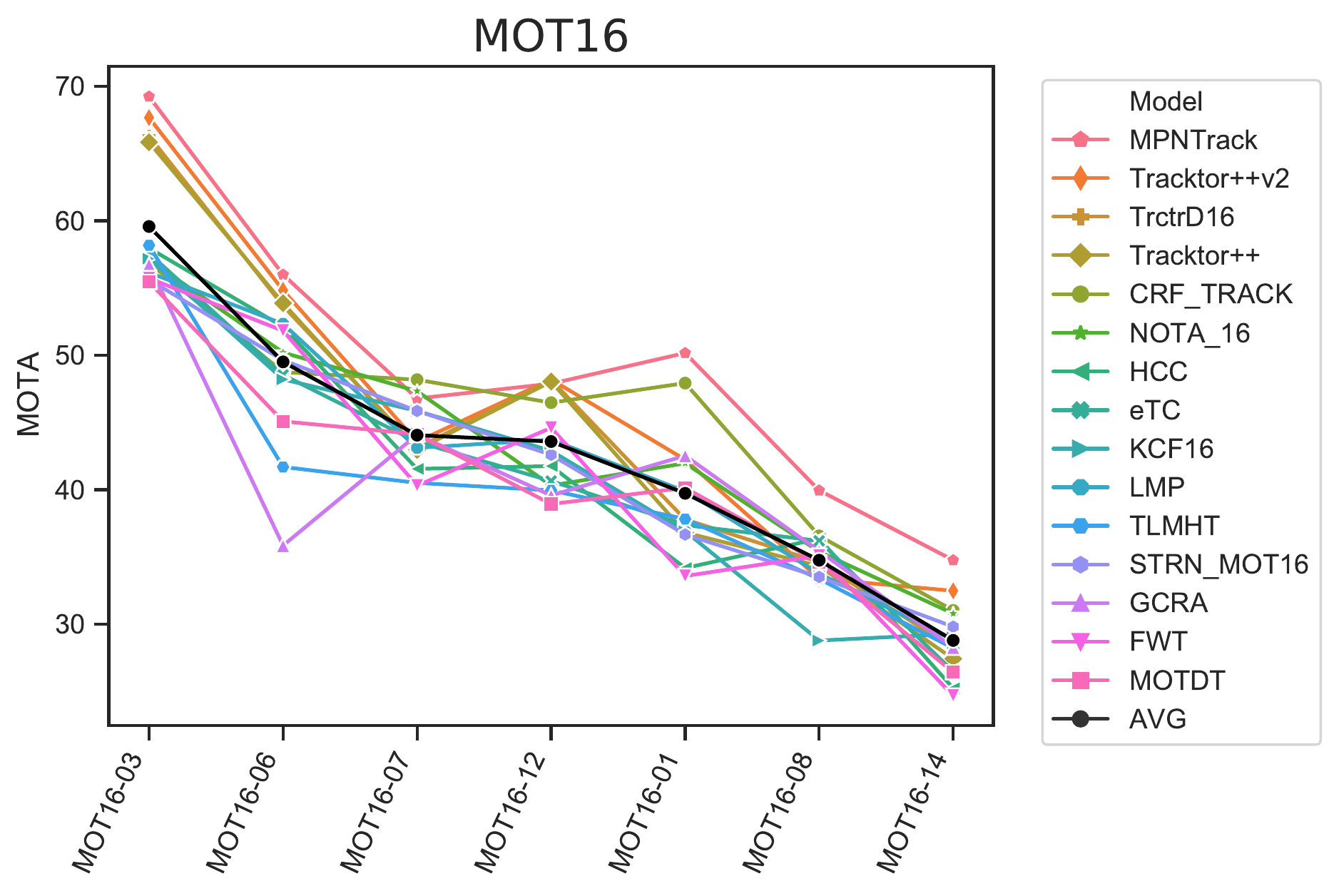}
  \end{minipage}
  
     \begin{minipage}{1.\linewidth}
    \centering \includegraphics[width=0.73\linewidth]{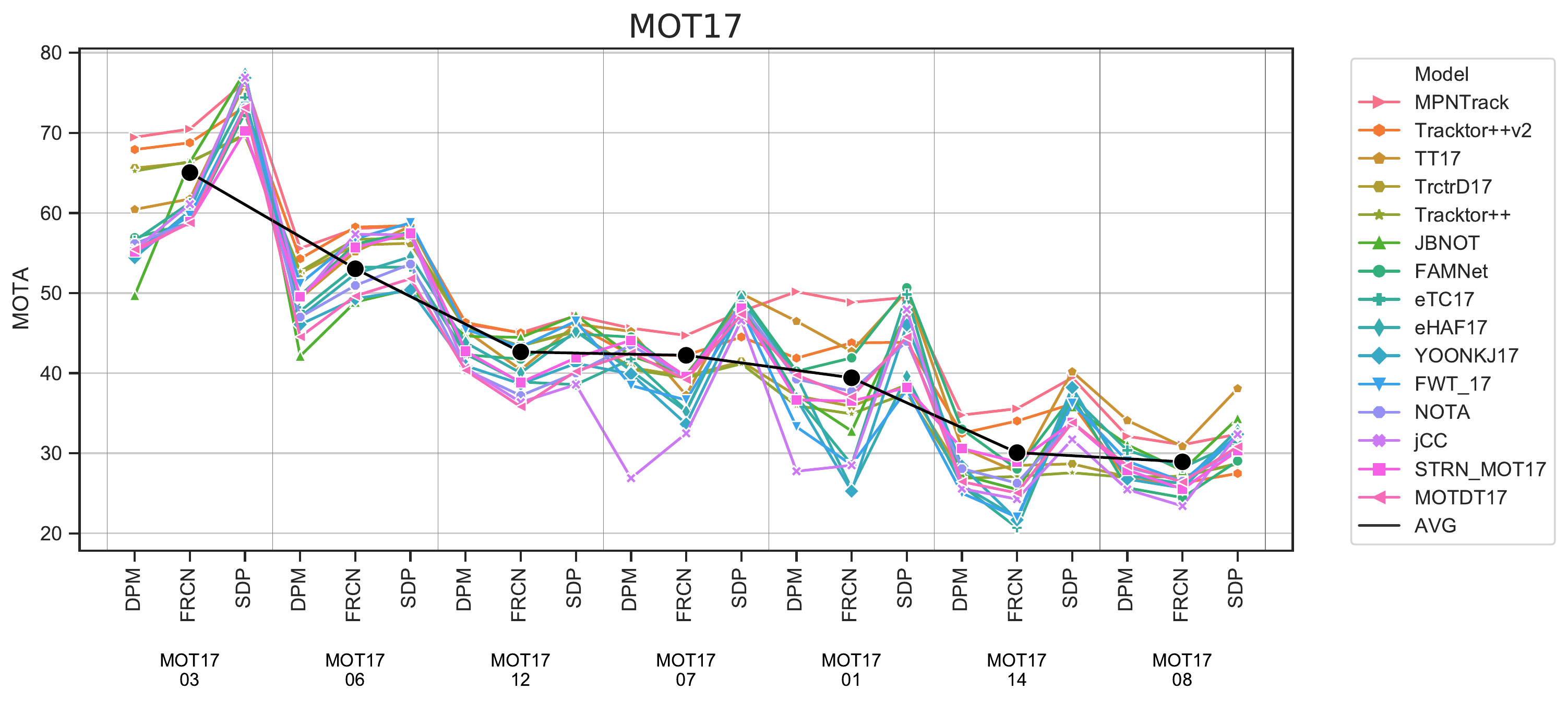}
      \end{minipage}
    \caption{Graphical overview of the top 15 trackers of all benchmarks. The entries are ordered from easiest sequence / best performing method, to hardest sequence / poorest performance, respectively. The mean performance across all sequences / submissions is depicted with a thick black line.}
    \label{fig:results-detailed}
   
\end{figure*}

\section{Analysis of State-of-the-Art Trackers}
\label{sec:experiments}

\begin{table*}[htb]
\tiny
\caption{The \MOTOLD leaderboard. Performance of several trackers according to different metrics.}
\label{tab:mot15}
\smallskip
\centering
\begin{tabular*}{\linewidth}{@{\extracolsep{\stretch{1}}}l rrrrrrrrrrrr @{}}
\toprule
       Method &  MOTA &  IDF1 &  MOTP &  FAR &   MT &   ML &     FP &     FN &   IDSW &    FM &  IDSWR &   FMR \\
\midrule
      MPNTrack~\citep{mpntrack} & 51.54 & 58.61 & 76.05 & 1.32 &  225 &  187 &   7620 &  21780 &    375 &   872 &   5.81 & 13.51 \\
      \hline
 Tracktor++v2~\citep{tracktor} & 46.60 & 47.57 & 76.36 & 0.80 &  131 &  201 &   4624 &  26896 &   1290 &  1702 &  22.94 & 30.27 \\
     TrctrD15~\citep{trctrd} & 44.09 & 45.99 & 75.26 & 1.05 &  124 &  192 &   6085 &  26917 &   1347 &  1868 &  23.97 & 33.24 \\
  Tracktor++~\citep{tracktor} & 44.06 & 46.73 & 75.03 & 1.12 &  130 &  189 &   6477 &  26577 &   1318 &  1790 &  23.23 & 31.55 \\
  \hline
          KCF~\citep{kcf} & 38.90 & 44.54 & 70.56 & 1.27 &  120 &  227 &   7321 &  29501 &    720 &  1440 &  13.85 & 27.70 \\
  AP\textunderscore HWDPL\textunderscore p~\citep{ap_hwdpl_p} & 38.49 & 47.10 & 72.56 & 0.69 &   63 &  270 &   4005 &  33203 &    586 &  1263 &  12.75 & 27.48 \\
         STRN~\citep{strn} & 38.06 & 46.62 & 72.06 & 0.94 &   83 &  241 &   5451 &  31571 &   1033 &  2665 &  21.25 & 54.82 \\
      AMIR15~\citep{amir} & 37.57 & 46.01 & 71.66 & 1.37 &  114 &  193 &   7933 &  29397 &   1026 &  2024 &  19.67 & 38.81 \\
      JointMC~\citep{jointmc} & 35.64 & 45.12 & 71.90 & 1.83 &  167 &  283 &  10580 &  28508 &    457 &   969 &   8.53 & 18.08 \\
     RAR15pub~\citep{rar15pub} & 35.11 & 45.40 & 70.94 & 1.17 &   94 &  305 &   6771 &  32717 &    381 &  1523 &   8.15 & 32.58 \\
    HybridDAT~\citep{hybriddat} & 34.97 & 47.72 & 72.57 & 1.46 &   82 &  304 &   8455 &  31140 &    358 &  1267 &   7.26 & 25.69 \\
      INARLA~\citep{inarla} & 34.69 & 42.06 & 70.72 & 1.71 &   90 &  216 &   9855 &  29158 &   1112 &  2848 &  21.16 & 54.20 \\
          STAM~\citep{stam16} & 34.33 & 48.26 & 70.55 & 0.89 &   82 &  313 &   5154 &  34848 &    348 &  1463 &   8.04 & 33.80 \\
      QuadMOT~\citep{quadmot} & 33.82 & 40.43 & 73.42 & 1.37 &   93 &  266 &   7898 &  32061 &    703 &  1430 &  14.70 & 29.91 \\
         NOMT~\citep{nomt} & 33.67 & 44.55 & 71.94 & 1.34 &   88 &  317 &   7762 &  32547 &    442 &   823 &   9.40 & 17.50 \\
        DCCRF~\citep{dccrf} & 33.62 & 39.08 & 70.91 & 1.02 &   75 &  271 &   5917 &  34002 &    866 &  1566 &  19.39 & 35.07 \\
         TDAM~\citep{tdam} & 33.03 & 46.05 & 72.78 & 1.74 &   96 &  282 &  10064 &  30617 &    464 &  1506 &   9.25 & 30.02 \\
   CDA\textunderscore DDALpb~\citep{cda_ddal} & 32.80 & 38.79 & 70.70 & 0.86 &   70 &  304 &   4983 &  35690 &    614 &  1583 &  14.65 & 37.77 \\
      MHT\textunderscore DAM~\citep{mht_dam} & 32.36 & 45.31 & 71.83 & 1.57 &  115 &  316 &   9064 &  32060 &    435 &   826 &   9.10 & 17.27 \\
          LFNF~\citep{lfnf} & 31.64 & 33.10 & 72.03 & 1.03 &   69 &  301 &   5943 &  35095 &    961 &  1106 &  22.41 & 25.79 \\
    GMPHD\textunderscore OGM~\citep{gmphd_ogm} & 30.72 & 38.82 & 71.64 & 1.13 &   83 &  275 &   6502 &  35030 &   1034 &  1351 &  24.05 & 31.43 \\
     PHD\textunderscore GSDL~\citep{phd_gsdl} & 30.51 & 38.82 & 71.20 & 1.13 &   55 &  297 &   6534 &  35284 &    879 &  2208 &  20.65 & 51.87 \\
          MDP~\citep{mdp} & 30.31 & 44.68 & 71.32 & 1.68 &   94 &  277 &   9717 &  32422 &    680 &  1500 &  14.40 & 31.76 \\
          \hline
      MCF\textunderscore PHD~\citep{mcf_phd} & 29.89 & 38.18 & 71.70 & 1.54 &   86 &  317 &   8892 &  33529 &    656 &   989 &  14.44 & 21.77 \\
      CNNTCM~\citep{cnntcm} & 29.64 & 36.82 & 71.78 & 1.35 &   81 &  317 &   7786 &  34733 &    712 &   943 &  16.38 & 21.69 \\
        RSCNN~\citep{rscnn} & 29.50 & 36.97 & 73.07 & 2.05 &   93 &  262 &  11866 &  30474 &    976 &  1176 &  19.36 & 23.33 \\
      TBSS15~\citep{tbss} & 29.21 & 37.23 & 71.28 & 1.05 &   49 &  316 &   6068 &  36779 &    649 &  1508 &  16.17 & 37.57 \\
         SCEA~\citep{scea} & 29.08 & 37.15 & 71.11 & 1.05 &   64 &  341 &   6060 &  36912 &    604 &  1182 &  15.13 & 29.61 \\
  SiameseCNN~\citep{Leal:2016:CVPRW} & 29.04 & 34.27 & 71.20 & 0.89 &   61 &  349 &   5160 &  37798 &    639 &  1316 &  16.61 & 34.20 \\
  HAM\textunderscore INTP15~\citep{ham_sadf17} & 28.62 & 41.45 & 71.13 & 1.30 &   72 &  317 &   7485 &  35910 &    460 &  1038 &  11.07 & 24.98 \\
    GMMA\textunderscore intp~\citep{gmma_intp} & 27.32 & 36.59 & 70.92 & 1.36 &   47 &  311 &   7848 &  35817 &    987 &  1848 &  23.67 & 44.31 \\
         oICF~\citep{oicf} & 27.08 & 40.49 & 69.96 & 1.31 &   46 &  351 &   7594 &  36757 &    454 &  1660 &  11.30 & 41.32 \\
          TO~\citep{to} & 25.66 & 32.74 & 72.17 & 0.83 &   31 &  414 &   4779 &  40511 &    383 &   600 &  11.24 & 17.61 \\
      LP\textunderscore SSVM~\citep{lp_ssvm} & 25.22 & 34.05 & 71.68 & 1.45 &   42 &  382 &   8369 &  36932 &    646 &   849 &  16.19 & 21.28 \\
     HAM\textunderscore SADF~\citep{ham_sadf17} & 25.19 & 37.80 & 71.38 & 1.27 &   41 &  420 &   7330 &  38275 &    357 &   745 &   9.47 & 19.76 \\
          ELP~\citep{elp} & 24.99 & 26.21 & 71.17 & 1.27 &   54 &  316 &   7345 &  37344 &   1396 &  1804 &  35.60 & 46.00 \\
      AdTobKF~\citep{adtobkf} & 24.82 & 34.50 & 70.78 & 1.07 &   29 &  375 &   6201 &  39321 &    666 &  1300 &  18.50 & 36.11 \\
        LINF1~\citep{linf1} & 24.53 & 34.82 & 71.33 & 1.01 &   40 &  466 &   5864 &  40207 &    298 &   744 &   8.62 & 21.53 \\
      TENSOR~\citep{tensor} & 24.32 & 24.13 & 71.58 & 1.15 &   40 &  336 &   6644 &  38582 &   1271 &  1304 &  34.16 & 35.05 \\
        TFMOT~\citep{jcmin_mot} & 23.81 & 32.30 & 71.35 & 0.78 &   35 &  447 &   4533 &  41873 &    404 &   792 &  12.69 & 24.87 \\
      JPDA\textunderscore m~\citep{jpda_m} & 23.79 & 33.77 & 68.17 & 1.10 &   36 &  419 &   6373 &  40084 &    365 &   869 &  10.50 & 25.00 \\
      MotiCon~\citep{moticon} & 23.07 & 29.38 & 70.87 & 1.80 &   34 &  375 &  10404 &  35844 &   1018 &  1061 &  24.44 & 25.47 \\
  DEEPDA\textunderscore MOT~\citep{deepda_mot} & 22.53 & 25.92 & 70.92 & 1.27 &   46 &  447 &   7346 &  39092 &   1159 &  1538 &  31.86 & 42.28 \\
     SegTrack~\citep{segtrack} & 22.51 & 31.48 & 71.65 & 1.36 &   42 &  461 &   7890 &  39020 &    697 &   737 &  19.10 & 20.20 \\
     EAMTTpub~\citep{eamtt_17} & 22.30 & 32.84 & 70.79 & 1.37 &   39 &  380 &   7924 &  38982 &    833 &  1485 &  22.79 & 40.63 \\
    SAS\textunderscore MOT15~\citep{sas_mot} & 22.16 & 27.15 & 71.10 & 0.97 &   22 &  444 &   5591 &  41531 &    700 &  1240 &  21.60 & 38.27 \\
      OMT\textunderscore DFH~\citep{omt_dfh} & 21.16 & 37.34 & 69.94 & 2.29 &   51 &  335 &  13218 &  34657 &    563 &  1255 &  12.92 & 28.79 \\
  MTSTracker~(Nguyen Thi Lan Anh et\,al, \citeyear{mtstracker}) & 20.64 & 31.87 & 70.32 & 2.62 &   65 &  266 &  15161 &  32212 &   1387 &  2357 &  29.16 & 49.55 \\
  TC\textunderscore SIAMESE~\citep{tc_siamese} & 20.22 & 32.59 & 71.09 & 1.06 &   19 &  487 &   6127 &  42596 &    294 &   825 &   9.59 & 26.90 \\
  \hline
        DCO\textunderscore X~\citep{dco_x} & 19.59 & 31.45 & 71.39 & 1.84 &   37 &  396 &  10652 &  38232 &    521 &   819 &  13.79 & 21.68 \\
          CEM~\citep{Milan14pami} & 19.30 &  N/A & 70.74 & 2.45 &   61 &  335 &  14180 &  34591 &    813 &  1023 &  18.60 & 23.41 \\
     RNN\textunderscore LSTM~\citep{rnn_lstm} & 18.99 & 17.12 & 70.97 & 2.00 &   40 &  329 &  11578 &  36706 &   1490 &  2081 &  37.01 & 51.69 \\
         RMOT~\citep{yoonwacv2015} & 18.63 & 32.56 & 69.57 & 2.16 &   38 &  384 &  12473 &  36835 &    684 &  1282 &  17.08 & 32.01 \\
     TSDA\textunderscore OAL~\citep{tsda_oal} & 18.61 & 36.07 & 69.68 & 2.83 &   68 &  305 &  16350 &  32853 &    806 &  1544 &  17.32 & 33.18 \\
     GMPHD\textunderscore 15~\citep{gmphd_15} & 18.47 & 28.38 & 70.90 & 1.36 &   28 &  399 &   7864 &  41766 &    459 &  1266 &  14.33 & 39.54 \\
         SMOT~\citep{smot} & 18.23 &  0.00 & 71.23 & 1.52 &   20 &  395 &   8780 &  40310 &   1148 &  2132 &  33.38 & 61.99 \\
     ALExTRAC~\citep{BewleyICRA2016} & 16.95 & 17.30 & 71.18 & 1.60 &   28 &  378 &   9233 &  39933 &   1859 &  1872 &  53.11 & 53.48 \\
          TBD~\citep{tbd} & 15.92 &  0.00 & 70.86 & 2.58 &   46 &  345 &  14943 &  34777 &   1939 &  1963 &  44.68 & 45.23 \\
         GSCR~\citep{gscr} & 15.78 & 27.90 & 69.38 & 1.31 &   13 &  440 &   7597 &  43633 &    514 &  1010 &  17.73 & 34.85 \\
      TC\textunderscore ODAL~\citep{Bae14CVPR} & 15.13 &  0.00 & 70.53 & 2.24 &   23 &  402 &  12970 &  38538 &    637 &  1716 &  17.09 & 46.04 \\
      DP\textunderscore NMS~\citep{Pirsiavash11CVPR} & 14.52 & 19.69 & 70.76 & 2.28 &   43 &  294 &  13171 &  34814 &   4537 &  3090 & 104.69 & 71.30 \\
\bottomrule
\end{tabular*}
\end{table*}

\begin{table*}[!ht]
\tiny
\caption{The \MOTNEW leaderboard. Performance of several trackers according to different metrics.}
\label{tab:mot16}
\smallskip
\centering
\begin{tabular*}{\linewidth}{@{\extracolsep{\stretch{1}}}l rrrrrrrrrrrr @{}}
\toprule
       Method &  MOTA &  IDF1 &  MOTP &  FAR &   MT &   ML &     FP &      FN &  IDSW &    FM &  IDSWR &    FMR \\
\midrule
     MPNTrack~\citep{mpntrack} & 58.56 & 61.69 & 78.88 & 0.84 &  207 &  258 &   4949 &   70252 &   354 &   684 &   5.76 &  11.13 \\
 Tracktor++v2~\citep{tracktor} & 56.20 & 54.91 & 79.20 & 0.40 &  157 &  272 &   2394 &   76844 &   617 &  1068 &  10.66 &  18.46 \\
     TrctrD16~\citep{trctrd} & 54.83 & 53.39 & 77.47 & 0.50 &  145 &  281 &   2955 &   78765 &   645 &  1515 &  11.36 &  26.67 \\
   Tracktor++~\citep{tracktor} & 54.42 & 52.54 & 78.22 & 0.55 &  144 &  280 &   3280 &   79149 &   682 &  1480 &  12.05 &  26.15 \\
       \hline
      NOTA\textunderscore 16~\citep{nota} & 49.83 & 55.33 & 74.49 & 1.22 &  136 &  286 &   7248 &   83614 &   614 &  1372 &  11.34 &  25.34 \\
          HCC~\citep{hcc} & 49.25 & 50.67 & 79.00 & 0.90 &  135 &  303 &   5333 &   86795 &   391 &   535 &   7.46 &  10.21 \\
          eTC~\citep{etc} & 49.15 & 56.11 & 75.49 & 1.42 &  131 &  306 &   8400 &   83702 &   606 &   882 &  11.20 &  16.31 \\
        KCF16~\citep{kcf} & 48.80 & 47.19 & 75.66 & 0.99 &  120 &  289 &   5875 &   86567 &   906 &  1116 &  17.25 &  21.25 \\
          LMP~\citep{lmp} & 48.78 & 51.26 & 79.04 & 1.12 &  138 &  304 &   6654 &   86245 &   481 &   595 &   9.13 &  11.29 \\
         TLMHT~\citep{tlmht} & 48.69 & 55.29 & 76.43 & 1.12 &  119 &  338 &   6632 &   86504 &   413 &   642 &   7.86 &  12.22 \\
   STRN\textunderscore MOT16~\citep{strn} & 48.46 & 53.90 & 73.75 & 1.53 &  129 &  265 &   9038 &   84178 &   747 &  2919 &  13.88 &  54.23 \\
         GCRA~\citep{gcra} & 48.16 & 48.55 & 77.50 & 0.86 &   98 &  312 &   5104 &   88586 &   821 &  1117 &  15.97 &  21.73 \\
          FWT~\citep{fwt} & 47.77 & 44.28 & 75.51 & 1.50 &  145 &  290 &   8886 &   85487 &   852 &  1534 &  16.04 &  28.88 \\
        MOTDT~\citep{motdt} & 47.63 & 50.94 & 74.81 & 1.56 &  115 &  291 &   9253 &   85431 &   792 &  1858 &  14.90 &  34.96 \\
       NLLMPa~\citep{nllmpa} & 47.58 & 47.34 & 78.51 & 0.99 &  129 &  307 &   5844 &   89093 &   629 &   768 &  12.30 &  15.02 \\
       EAGS16~\citep{eags16} & 47.41 & 50.13 & 75.95 & 1.41 &  131 &  324 &   8369 &   86931 &   575 &   913 &  10.99 &  17.45 \\
        JCSTD~\citep{jcstd} & 47.36 & 41.10 & 74.43 & 1.36 &  109 &  276 &   8076 &   86638 &  1266 &  2697 &  24.12 &  51.39 \\
         ASTT ~\citep{astt}& 47.24 & 44.27 & 76.08 & 0.79 &  124 &  316 &   4680 &   90877 &   633 &   814 &  12.62 &  16.23 \\
       eHAF16~\citep{ehaf} & 47.22 & 52.44 & 75.69 & 2.13 &  141 &  325 &  12586 &   83107 &   542 &   787 &   9.96 &  14.46 \\
         AMIR~\citep{amir} & 47.17 & 46.29 & 75.82 & 0.45 &  106 &  316 &   2681 &   92856 &   774 &  1675 &  15.77 &  34.14 \\
       JointMC (MCjoint)~\citep{jointmc} & 47.10 & 52.26 & 76.27 & 1.13 &  155 &  356 &   6703 &   89368 &   370 &   598 &   7.26 &  11.73 \\
     YOONKJ16~\citep{yoonkj} & 46.96 & 50.05 & 75.76 & 1.33 &  125 &  317 &   7901 &   88179 &   627 &   945 &  12.14 &  18.30 \\
      NOMT\textunderscore 16~\citep{nomt} & 46.42 & 53.30 & 76.56 & 1.65 &  139 &  314 &   9753 &   87565 &   359 &   504 &   6.91 &   9.70 \\
          JMC~\citep{jmc} & 46.28 & 46.31 & 75.68 & 1.08 &  118 &  301 &   6373 &   90914 &   657 &  1114 &  13.10 &  22.22 \\
    DD\textunderscore TAMA16~\citep{dd_tama16} & 46.20 & 49.43 & 75.42 & 0.87 &  107 &  334 &   5126 &   92367 &   598 &  1127 &  12.12 &  22.84 \\
      DMAN\textunderscore 16~\citep{dman} & 46.08 & 54.82 & 73.77 & 1.34 &  132 &  324 &   7909 &   89874 &   532 &  1616 &  10.49 &  31.87 \\
       STAM16~\citep{stam16} & 45.98 & 50.05 & 74.92 & 1.16 &  111 &  331 &   6895 &   91117 &   473 &  1422 &   9.46 &  28.43 \\
     RAR16pub~\citep{rar15pub} & 45.87 & 48.77 & 74.84 & 1.16 &  100 &  318 &   6871 &   91173 &   648 &  1992 &  12.96 &  39.85 \\
   MHT\textunderscore DAM\textunderscore 16~\citep{mht_dam} & 45.83 & 46.06 & 76.34 & 1.08 &  123 &  328 &   6412 &   91758 &   590 &   781 &  11.88 &  15.72 \\
         MTDF~\citep{mtdf} & 45.72 & 40.07 & 72.63 & 2.03 &  107 &  276 &  12018 &   84970 &  1987 &  3377 &  37.21 &  63.24 \\
   INTERA\textunderscore MOT~\citep{intera_mot} & 45.40 & 47.66 & 74.41 & 2.27 &  137 &  294 &  13407 &   85547 &   600 &   930 &  11.30 &  17.52 \\
         EDMT~\citep{edmt} & 45.34 & 47.86 & 75.94 & 1.88 &  129 &  303 &  11122 &   87890 &   639 &   946 &  12.34 &  18.27 \\
      DCCRF16~\citep{dccrf} & 44.76 & 39.67 & 75.63 & 0.95 &  107 &  321 &   5613 &   94133 &   968 &  1378 &  20.01 &  28.49 \\
         TBSS~\citep{tbss} & 44.58 & 42.64 & 75.18 & 0.70 &   93 &  333 &   4136 &   96128 &   790 &  1419 &  16.71 &  30.01 \\
    OTCD\textunderscore 1\textunderscore 16~\citep{otcd_1} & 44.36 & 45.62 & 75.36 & 0.97 &   88 &  361 &   5759 &   94927 &   759 &  1787 &  15.83 &  37.28 \\
    QuadMOT16~\citep{quadmot} & 44.10 & 38.27 & 76.40 & 1.08 &  111 &  341 &   6388 &   94775 &   745 &  1096 &  15.52 &  22.83 \\
   CDA\textunderscore DDALv2~\citep{cda_ddal} & 43.89 & 45.13 & 74.69 & 1.09 &   81 &  337 &   6450 &   95175 &   676 &  1795 &  14.14 &  37.55 \\
       LFNF16~\citep{lfnf} & 43.61 & 41.62 & 76.63 & 1.12 &  101 &  347 &   6616 &   95363 &   836 &   938 &  17.53 &  19.67 \\
      oICF\textunderscore 16~\citep{oicf} & 43.21 & 49.33 & 74.31 & 1.12 &   86 &  368 &   6651 &   96515 &   381 &  1404 &   8.10 &  29.83 \\
   MHT\textunderscore bLSTM6~\citep{mht_blstm} & 42.10 & 47.84 & 75.85 & 1.97 &  113 &  337 &  11637 &   93172 &   753 &  1156 &  15.40 &  23.64 \\
     LINF1\textunderscore 16~\citep{linf1} & 41.01 & 45.69 & 74.85 & 1.33 &   88 &  389 &   7896 &   99224 &   430 &   963 &   9.43 &  21.13 \\
   PHD\textunderscore GSDL16~\citep{phd_gsdl} & 41.00 & 43.14 & 75.90 & 1.10 &   86 &  315 &   6498 &   99257 &  1810 &  3650 &  39.73 &  80.11 \\
   GMPHD\textunderscore ReId~\cite{gmphd_reid} & 40.42 & 49.71 & 75.25 & 1.11 &   85 &  329 &   6572 &  101266 &   792 &  2529 &  17.81 &  56.88 \\
       AM\textunderscore ADM~\citep{am_adm} & 40.12 & 43.79 & 75.45 & 1.44 &   54 &  351 &   8503 &   99891 &   789 &  1736 &  17.45 &  38.40 \\
       \hline
    EAMTT\textunderscore pub~\citep{eamtt_17} & 38.83 & 42.43 & 75.15 & 1.37 &   60 &  373 &   8114 &  102452 &   965 &  1657 &  22.03 &  37.83 \\
         OVBT~\citep{BanBMTT2016} & 38.40 & 37.82 & 75.39 & 1.95 &   57 &  359 &  11517 &   99463 &  1321 &  2140 &  29.07 &  47.09 \\
        GMMCP~\citep{gmmcp} & 38.10 & 35.50 & 75.84 & 1.12 &   65 &  386 &   6607 &  105315 &   937 &  1669 &  22.18 &  39.51 \\
    LTTSC-CRF~\citep{lttsc-crf} & 37.59 & 42.06 & 75.94 & 2.02 &   73 &  419 &  11969 &  101343 &   481 &  1012 &  10.83 &  22.79 \\
    JCmin\textunderscore MOT~\citep{jcmin_mot} & 36.65 & 36.16 & 75.86 & 0.50 &   57 &  413 &   2936 &  111890 &   667 &   831 &  17.27 &  21.51 \\
       HISP\textunderscore T~\citep{hisp_t} & 35.87 & 28.93 & 76.07 & 1.08 &   59 &  380 &   6412 &  107918 &  2594 &  2298 &  63.56 &  56.31 \\
      LP2D\textunderscore 16~\citep{moticon} & 35.74 & 34.18 & 75.84 & 0.86 &   66 &  385 &   5084 &  111163 &   915 &  1264 &  23.44 &  32.39 \\
   GM\textunderscore PHD\textunderscore DAL~\citep{gmphd_dal} & 35.13 & 26.58 & 76.59 & 0.40 &   53 &  390 &   2350 &  111886 &  4047 &  5338 & 104.75 & 138.17 \\
       TBD\textunderscore 16~\citep{tbd} & 33.74 &  0.00 & 76.53 & 0.98 &   55 &  411 &   5804 &  112587 &  2418 &  2252 &  63.22 &  58.88 \\
   GM\textunderscore PHD\textunderscore N1T~\citep{gm_phd_n1t} & 33.25 & 25.47 & 76.84 & 0.30 &   42 &  425 &   1750 &  116452 &  3499 &  3594 &  96.85 &  99.47 \\
    CEM\textunderscore 16~\citep{Milan14pami} &  33.19 &  N/A & 75.84 & 1.16 &    59 &  413 &      6837 &  114322 &     642 &     731 &   17.21 & 19.60 \\
    GMPHD\textunderscore HDA~\citep{gmphd_15} & 30.52 & 33.37 & 75.42 & 0.87 &   35 &  453 &   5169 &  120970 &   539 &   731 &  16.02 &  21.72 \\
      SMOT\textunderscore 16~\citep{smot} & 29.75 &  N/A & 75.18 & 2.94 &   40 &  362 &  17426 &  107552 &  3108 &  4483 &  75.79 & 109.32 \\
      \hline
    JPDA\textunderscore m\textunderscore 16~\citep{jpda_m} & 26.17 &   N/A& 76.34 & 0.62 &   31 &  512 &   3689 &  130549 &   365 &   638 &  12.85 &  22.47 \\
    DP\textunderscore NMS\textunderscore 16~\citep{Pirsiavash11CVPR} & 26.17 & 31.19 & 76.34 & 0.62 &   31 &  512 &   3689 &  130557 &   365 &   638 &  12.86 &  22.47 \\
       
\bottomrule
\end{tabular*}
\end{table*}

\begin{table*}[htbp]
\tiny
\caption{The \MOTLAST leaderboard. Performance of several trackers according to different metrics.}
\label{tab:mot17}
\smallskip
\centering
\begin{tabular*}{\linewidth}{@{\extracolsep{\stretch{1}}}l rrrrrrrrrrrr @{}}
%
\toprule
       Method &  MOTA &  IDF1 &  MOTP &  FAR &   MT &    ML &     FP &      FN &   IDSW &     FM &  IDSWR &    FMR \\
\midrule
     MPNTrack~\citep{mpntrack} & 58.85 & 61.75 & 78.62 & 0.98 &  679 &   788 &  17413 &  213594 &   1185 &   2265 &  19.07 &  36.45 \\
 Tracktor++v2~\citep{tracktor} & 56.35 & 55.12 & 78.82 & 0.50 &  498 &   831 &   8866 &  235449 &   1987 &   3763 &  34.10 &  64.58 \\
     TrctrD17~\citep{trctrd} & 53.72 & 53.77 & 77.23 & 0.66 &  458 &   861 &  11731 &  247447 &   1947 &   4792 &  34.68 &  85.35 \\
   Tracktor++~\citep{tracktor} & 53.51 & 52.33 & 77.98 & 0.69 &  459 &   861 &  12201 &  248047 &   2072 &   4611 &  36.98 &  82.28 \\
        JBNOT~\citep{jbnot} & 52.63 & 50.77 & 77.12 & 1.78 &  465 &   844 &  31572 &  232659 &   3050 &   3792 &  51.90 &  64.53 \\
       FAMNet~\citep{famnet} & 52.00 & 48.71 & 76.48 & 0.80 &  450 &   787 &  14138 &  253616 &   3072 &   5318 &  55.80 &  96.60 \\
        eTC17~\citep{etc} & 51.93 & 58.13 & 76.34 & 2.04 &  544 &   836 &  36164 &  232783 &   2288 &   3071 &  38.95 &  52.28 \\
       eHAF17~\citep{ehaf} & 51.82 & 54.72 & 77.03 & 1.87 &  551 &   893 &  33212 &  236772 &   1834 &   2739 &  31.60 &  47.19 \\
     YOONKJ17~\citep{yoonkj} & 51.37 & 53.98 & 77.00 & 1.64 &  500 &   878 &  29051 &  243202 &   2118 &   3072 &  37.23 &  53.99 \\
       FWT\textunderscore 17~\citep{fwt} & 51.32 & 47.56 & 77.00 & 1.36 &  505 &   830 &  24101 &  247921 &   2648 &   4279 &  47.24 &  76.33 \\
         NOTA~\citep{nota} & 51.27 & 54.46 & 76.68 & 1.13 &  403 &   833 &  20148 &  252531 &   2285 &   5798 &  41.36 & 104.95 \\
          JointMC (jCC)~\citep{jointmc} & 51.16 & 54.50 & 75.92 & 1.46 &  493 &   872 &  25937 &  247822 &   1802 &   2984 &  32.13 &  53.21 \\
   STRN\textunderscore MOT17~\citep{strn} & 50.90 & 55.98 & 75.58 & 1.42 &  446 &   797 &  25295 &  249365 &   2397 &   9363 &  42.95 & 167.78 \\
      MOTDT17~\citep{motdt} & 50.85 & 52.70 & 76.58 & 1.36 &  413 &   841 &  24069 &  250768 &   2474 &   5317 &  44.53 &  95.71 \\
   MHT\textunderscore DAM\textunderscore 17~\citep{mht_dam} & 50.71 & 47.18 & 77.52 & 1.29 &  491 &   869 &  22875 &  252889 &   2314 &   2865 &  41.94 &  51.92 \\
      TLMHT\textunderscore 17~\citep{tlmht} & 50.61 & 56.51 & 77.65 & 1.25 &  415 &  1022 &  22213 &  255030 &   1407 &   2079 &  25.68 &  37.94 \\
       EDMT17~\citep{edmt} & 50.05 & 51.25 & 77.26 & 1.82 &  509 &   855 &  32279 &  247297 &   2264 &   3260 &  40.31 &  58.04 \\
        \hline
   GMPHDOGM17~\citep{gmphd_ogm} & 49.94 & 47.15 & 77.01 & 1.35 &  464 &   895 &  24024 &  255277 &   3125 &   3540 &  57.07 &  64.65 \\
       MTDF17~\citep{mtdf} & 49.58 & 45.22 & 75.48 & 2.09 &  444 &   779 &  37124 &  241768 &   5567 &   9260 &  97.41 & 162.03 \\
       PHD\textunderscore GM~(Sanchez-Matilla et\,al., \citeyear{phd_gm}) & 48.84 & 43.15 & 76.74 & 1.48 &  449 &   830 &  26260 &  257971 &   4407 &   6448 &  81.19 & 118.79 \\
    OTCD\textunderscore 1\textunderscore 17~\citep{otcd_1} & 48.57 & 47.90 & 76.91 & 1.04 &  382 &   970 &  18499 &  268204 &   3502 &   5588 &  66.75 & 106.51 \\
   HAM\textunderscore SADF17~\citep{ham_sadf17} & 48.27 & 51.14 & 77.22 & 1.18 &  402 &   981 &  20967 &  269038 &   1871 &   3020 &  35.76 &  57.72 \\
         DMAN~\citep{dman} & 48.24 & 55.69 & 75.69 & 1.48 &  454 &   902 &  26218 &  263608 &   2194 &   5378 &  41.18 & 100.94 \\
     AM\textunderscore ADM17~\citep{am_adm} & 48.11 & 52.07 & 76.69 & 1.41 &  316 &   934 &  25061 &  265495 &   2214 &   5027 &  41.82 &  94.95 \\
   PHD\textunderscore GSDL17~\citep{phd_gsdl} & 48.04 & 49.63 & 77.15 & 1.31 &  402 &   838 &  23199 &  265954 &   3998 &   8886 &  75.63 & 168.09 \\
    MHT\textunderscore bLSTM~\citep{mht_blstm} & 47.52 & 51.92 & 77.49 & 1.46 &  429 &   981 &  25981 &  268042 &   2069 &   3124 &  39.41 &  59.51 \\
         MASS~\citep{mass} & 46.95 & 45.99 & 76.11 & 1.45 &  399 &   856 &  25733 &  269116 &   4478 &  11994 &  85.62 & 229.31 \\
   GMPHD\textunderscore Rd17~\citep{gmphd_reid} & 46.83 & 54.06 & 76.41 & 2.17 &  464 &   784 &  38452 &  257678 &   3865 &   8097 &  71.14 & 149.03 \\
        IOU17~\citep{iou} & 45.48 & 39.40 & 76.85 & 1.13 &  369 &   953 &  19993 &  281643 &   5988 &   7404 & 119.56 & 147.84 \\
     LM\textunderscore NN\textunderscore 17~\citep{lm_nn_17} & 45.13 & 43.17 & 78.93 & 0.61 &  348 &  1088 &  10834 &  296451 &   2286 &   2463 &  48.17 &  51.90 \\
         FPSN~\citep{fpsn} & 44.91 & 48.43 & 76.61 & 1.90 &  388 &   844 &  33757 &  269952 &   7136 &  14491 & 136.82 & 277.84 \\
     HISP\textunderscore T17~\citep{hisp_dal} & 44.62 & 38.79 & 77.19 & 1.43 &  355 &   913 &  25478 &  276395 &  10617 &   7487 & 208.12 & 146.76 \\
    GMPHD\textunderscore DAL~\citep{gmphd_dal} & 44.40 & 36.23 & 77.42 & 1.08 &  350 &   927 &  19170 &  283380 &  11137 &  13900 & 223.74 & 279.25 \\
    SAS\textunderscore MOT17~\citep{sas_mot} & 44.24 & 57.18 & 76.42 & 1.66 &  379 &  1044 &  29473 &  283611 &   1529 &   2644 &  30.74 &  53.16 \\
    GMPHD\textunderscore SHA~\citep{gmphd_15} & 43.72 & 39.17 & 76.53 & 1.46 &  276 &  1012 &  25935 &  287758 &   3838 &   5056 &  78.33 & 103.18 \\
       SORT17~\citep{sort17} & 43.14 & 39.84 & 77.77 & 1.60 &  295 &   997 &  28398 &  287582 &   4852 &   7127 &  98.96 & 145.36 \\
     EAMTT\textunderscore 17~\citep{eamtt_17} & 42.63 & 41.77 & 76.03 & 1.73 &  300 &  1006 &  30711 &  288474 &   4488 &   5720 &  91.83 & 117.04 \\
   GMPHD\textunderscore N1Tr~\citep{gm_phd_n1t} & 42.12 & 33.87 & 77.66 & 1.03 &  280 &  1005 &  18214 &  297646 &  10698 &  10864 & 226.43 & 229.94 \\
   \hline
    GMPHD\textunderscore KCF~\citep{gmphd_kcf} & 39.57 & 36.64 & 74.54 & 2.87 &  208 &  1019 &  50903 &  284228 &   5811 &   7414 & 117.10 & 149.40 \\
       GM\textunderscore PHD~\citep{gm_phd} & 36.36 & 33.92 & 76.20 & 1.34 &   97 &  1349 &  23723 &  330767 &   4607 &  11317 & 111.34 & 273.51 \\
\bottomrule
\end{tabular*}
\end{table*}
\newcommand{\void}{---}
\newcommand{\fillme}{\textcolor{red}{\bf ??}}
\begin{table*}
    \tiny
    \centering
    \caption{\MOTOLD, \MOTNEW, \MOTLAST trackers and their characteristics. \\
    \textbf{Table legend:} \textit{App.} -- appearance model; \textit{OA} -- online target appearance adaptation; \textit{TR} -- target regression; \textit{ON} -- online method; \textit{(L)} -- learned. \\
    \textbf{Components:} \textit{MC} -- motion compensation module; \textit{OF} -- optical flow; \textit{Re-id} -- learned re-identification module; \textit{HoG} -- histogram of oriented gradients; \textit{NCC} -- normalized cross-correlation; \textit{IoU} -- intersection over union. \\
    \textbf{Association:} \textit{GMMCP} -- Generalized maximum multi-clique problem; \textit{MCF} -- Min-cost flow formulation~\citep{zhang08cvpr}; \textit{LP} -- linear programming; \textit{MHT} -- multi-hypothesis tracking~\citep{Reid:1979:MHT}; \textit{MWIS} -- maximum independent set problem; \textit{CRF} -- conditional random field formulation.}
    \begin{tabular}{@{} l  c c c c c c c  @{}} 
    \toprule
    Method & Box-box Affinity  & App. & Opt.  & Extra Inputs  & OA & TR & ON \\ \midrule
    %
    MPNTrack~\citep{mpntrack} & Appearance, geometry (L) & \cmark & MCF, LP & MC & \xmark & \cmark & \xmark \\
    DeepMOT~\citep{trctrd} & Re-id (L) & \cmark & \void & MC & \xmark & \cmark & \cmark \\

    TT17~\citep{zhang2020long} & Appearance, geometry (L) & \cmark & MHT/MWIS & \xmark & \xmark & \xmark & \xmark \\
    CRF\_TRACK~\citep{xiang2020end}& Appearance, geometry (L) & \cmark & CRF & re-id  & \xmark & \xmark & \xmark \\
    %

    \midrule
    Tracktor~\citep{tracktor} & Re-id (L) & \cmark & \void & MC, re-id & \xmark & \cmark & \cmark \\
    KCF~\citep{kcf} & Re-id (L) & \cmark & Multicut & re-id & \cmark & \cmark & \cmark \\
    STRN~\citep{strn} & Geometry, appearance (L) & \cmark & Hungarian algorithm & \void & \xmark & \xmark & \cmark \\
    JBNOT~\citep{jbnot} & Joint, body distances & \xmark & Frank-Wolfe algortihm & Body joint det. & \xmark & \xmark & \xmark \\
    FAMNet~\citep{famnet} & (L) & \cmark & Rank-1 tensor approx. & \void & \xmark & \cmark & \cmark \\
    %
    %
    \midrule
    %
    MHT\_bLSTM~\citep{mht_blstm} & Appearance, motion (L) & \cmark & MHT/MWIS & Pre-trained CNN & \cmark & \xmark & \cmark \\
    JointMC~\citep{jointmc} & DeepMatching (L), geometric & \cmark & Multicut & OF, non-nms dets & \xmark & \xmark & \xmark \\
    RAR~\citep{rar15pub} & Appearance, motion (L) & \cmark & Hungarian algorithm & \void & \xmark & \xmark & \cmark \\
    %
    %
    HCC~\citep{hcc} & Re-id (L) & \cmark & Multicut & External re-id & $\circ$ & \xmark & \xmark \\
    FWT~\citep{fwt} & DeepMatching, geometric & \cmark & Frank-Wolfe algorithm & Head detector & \xmark & \xmark & \xmark \\
    DMAN~\citep{dman} & Appearance (L), geometry & \cmark & \void & \void & \cmark & \cmark & \cmark \\
    eHAF~\citep{dman} & Appearance, motion & \cmark & MHT/MWIS & Super-pixels, OF & \xmark & \xmark & \xmark \\
    \midrule
    %
    %
    QuadMOT~\citep{quadmot} & Re-id (L), motion & \cmark & Min-max label prop. & \void & \cmark & \xmark & \cmark \\
    STAM~\citep{stam16} & Appearance (L), motion & \cmark & \void & \void & \cmark & \cmark & \cmark \\
    AMIR~\citep{amir} & Motion, appearance, interactions (L) &  \cmark & Hungarian algortihm & --- & \xmark & \xmark & \cmark \\
    LMP~\citep{lmp} & Re-id (L) & \cmark & Multicut & Non-nms det., re-id  & \cmark & \xmark & \xmark \\
    NLLMPa~\citep{nllmpa} & DeepMatching & \cmark & Multicut &  Non-NMS dets & \xmark &  \xmark & \xmark \\
    \midrule
    %
    LP\_SSVM~\citep{lp_ssvm} & Appearance, motion (L) & \cmark & MCF, greedy & \void & \xmark & \xmark & \xmark \\
    SiameseCNN~\citep{Leal:2016:CVPRW} & Appearance (L), geometry, motion & \cmark & MCF, LP & OF & \xmark & \xmark & \xmark \\
    SCEA~\citep{scea} & Appearance, geometry & \cmark & Clustering & \void & \xmark & \xmark & \cmark \\
    JMC~\citep{jmc} & DeepMatching & \cmark & Multicut &  Non-NMS dets & \xmark &  \xmark & \xmark  \\
    %
    %
    LINF1~\citep{linf1} & Sparse representation & \cmark & MCMC   & \void & \xmark &  \xmark & \xmark \\
    EAMTTpub~\citep{eamtt_17} & 2D distances & \xmark & Particle Filter  & Non-NMS dets &  \xmark & \xmark & \cmark \\
    OVBT~\citep{BanBMTT2016} & Dynamics from flow & \cmark & Variational EM  & OF & \xmark &  \xmark & \cmark\\
    LTTSC-CRF~\citep{lttsc-crf} & SURF & \cmark & CRF  & SURF & \xmark &\xmark & \xmark\\
    GMPHD\_HDA~\citep{gmphd_15} & HoG similarity, color histogram & \cmark & GM-PHD filter  &  HoG  &  \xmark & \xmark & \cmark \\
    DCO\_X~\citep{dco_x} & Motion, geometry & \cmark & CRF & \void & \xmark & \xmark & \xmark \\
    \midrule
    %
    ELP~\citep{elp} & Motion & \xmark & MCF, LP & \void & \xmark & \xmark & \xmark \\
    GMMCP~\citep{gmmcp} & Appearance, motion & \cmark & GMMCP/CRF & \void & \xmark & \xmark & \xmark \\
    MDP~\citep{mdp} & Motion (flow), geometry, appearance & \cmark & Hungarian algorithm & OF & \xmark & \xmark & \cmark \\
    MHT\_DAM~\citep{mht_dam} & (L) & \cmark & MHT/MWIS & --- & \cmark &  \xmark & \cmark \\
    NOMT~\citep{nomt} &  Interest point traj. &  \cmark &  CRF & OF & \xmark &  \xmark & \xmark\\
    JPDA\_m~\citep{jpda_m} & Mahalanobis distance & \xmark & LP & \void & \xmark  & \xmark & \xmark  \\
    SegTrack~\citep{segtrack} & Shape, geometry, motion & \cmark & CRF & OF, Super-pixels & \xmark & \xmark & \xmark \\
    \midrule
    %
    TBD~\citep {tbd} & IoU + NCC & \cmark & Hungarian algorithm  & \void & \xmark  &  \xmark & \xmark\\
    CEM~\citep{Milan14pami} & Motion & \xmark & Greedy sampling  & --- & \xmark  &  \xmark & \xmark\\
    MotiCon~\citep{moticon} & Motion descriptors & \xmark & MCF, LP & OF & \xmark & \xmark & \xmark \\
    \midrule
    %
    SMOT~\citep{smot} & Target dynamics & \xmark &   Hankel Least Squares  & \void & \xmark &  \xmark & \xmark \\
    \midrule
    %
    DP\_NMS~\citep{Pirsiavash11CVPR} & 2D image distances & \xmark & k-shortest paths  & \void & \xmark &  \xmark & \xmark\\
    LP2D~\citep{lealtaixe11iccv} & 2D image distances, IoU & \xmark & MCF, LP  & \void & \xmark &  \xmark & \xmark\\
    \bottomrule
    \end{tabular}
    \label{tab:characteristics}
    \vspace{-0.5em}
\end{table*}

We now present an analysis of recent multi-object tracking methods that submitted to the benchmark. This is divided into two parts:
(i) categorization of the methods, where our goal is to help young scientists to navigate the recent MOT literature, and (ii) error and runtime analysis, where we point out methods that have shown good performance on a wide range of scenes. We hope this can eventually lead to new promising research directions.

We consider all valid submissions to all three benchmarks that were published before April 17th, 2020, and used the provided set of public detections. 
For this analysis, we focus on methods that are peer-reviewed, \ie, published at a conference or a journal. 
We evaluate a total of 101 (public) trackers; \numtrold trackers were tested on \MOTOLD, \numtrnew on \MOTNEW and \numtrlast on \MOTLAST. 
A small subset of the submissions\footnote{The methods DP$\_$NMS, TC$\_$ODAL, TBD, SMOT, CEM, DCO$\_$X, and LP2D were taken as baselines for the benchmark.} were done by the benchmark organizers and not by the original authors of the respective method. 
Results for \MOTOLD are summarized in Tab.~\ref{tab:mot15}, for \MOTNEW in Tab.~\ref{tab:mot16} and for \MOTLAST in Tab.~\ref{tab:mot17}. The performance of the top 15 ranked trackers is demonstrated in Fig.~\ref{fig:results-detailed}.

\subsection{Trends in Tracking}

\PAR{Global optimization.} The community has long used the paradigm of tracking-by-detection for MOT, \ie, dividing the task into two steps: (i) object detection and (ii) data association, or temporal linking between detections.
The data association problem could be viewed as finding a set of disjoint paths in a graph, where nodes in the graph represent object detections, and links hypothesize feasible associations. Detectors usually produce multiple spatially-adjacent detection hypotheses, that are usually pruned using heuristic non-maximum suppression (NMS).

Before 2015, the community mainly focused on finding strong, preferably globally optimal methods to solve the data association problem. 
The task of linking detections into a consistent set of trajectories was often cast as, e.g., a graphical model and solved with k-shortest paths in DP$\_$NMS \citep{Pirsiavash11CVPR}, as a linear program solved with the simplex algorithm in LP2D \citep{lealtaixe11iccv}, as a Conditional Random Field in DCO$\_X$ \citep{dco_x}, SegTrack \citep{segtrack}, LTTSC-CRF \citep{lttsc-crf}, and GMMCP \citep{gmmcp}, using joint probabilistic data association filter (JPDA) \citep{jpda_m} or as a variational Bayesian model in OVBT \citep{BanBMTT2016}.  

A number of tracking approaches investigate the efficacy of using a Probability Hypothesis Density (PHD) filter-based tracking framework \citep{gmphd_ogm, phd_gsdl, mcf_phd, eamtt_17,gmphd_dal, gmphd_15, gm_phd_n1t,gmphd_reid}. This family of methods estimate states of multiple targets and data association simultaneously, reaching 30.72\% MOTA on \MOTOLD (GMPHD\_OGM), 41\% and 40.42\% on \MOTNEW (PHD\_GSDL and GMPHD\_ReId, respectively) and 49.94\% (GMPHD\_OGM) on \MOTLAST.

Newer methods \citep{Tang15CVPR} bypassed the need to pre-process object detections with NMS. They proposed a multi-cut optimization framework, which finds the connected components in a graph that represent feasible solutions, clustering all detections that correspond to the same target. 
This family of methods (JMC \citep{jmc}, LMP \citep{lmp}, NLLMPA \citep{nllmpa}, JointMC \citep{jointmc}, HCC \citep{hcc}) achieve 35.65\% MOTA on \MOTOLD (JointMC), 48.78\% and 49.25\% (LMP and HCC, respectively) on \MOTNEW and 51.16\% (JointMC) on \MOTLAST.

\PAR{Motion models.} A lot of attention has also been given to motion models, used as additional association affinity cues, \eg, SMOT \citep{smot}, CEM \citep{Milan14pami}, TBD \citep{tbd}, ELP \citep{elp} and MotiCon \citep{moticon}.
The pairwise costs for matching two detections were based on either simple distances or simple appearance models, such as color histograms.  
These methods achieve around 38\% MOTA on \MOTNEW (see \Tab~\ref{tab:mot16}) and 25\% on \MOTOLD (see \Tab~\ref{tab:mot15}). 

\PAR{Hand-crafted affinity measures.} After that, the attention shifted towards building robust pairwise similarity costs, mostly based on strong appearance cues or a combination of geometric and appearance cues.
This shift is clearly reflected in an improvement in tracker performance and the ability for trackers to handle more complex scenarios. 
For example, LINF1 \citep{linf1} uses sparse appearance models, and oICF \citep{oicf} use appearance models based on integral channel features. 
Top-performing methods of this class incorporate long-term interest point trajectories, \eg,~NOMT \citep{nomt}, and, more recently, learned models for sparse feature matching~JMC \citep{jmc} and JointMC \citep{jointmc} to improve pairwise affinity measures.
As can be seen in~\Tab~\ref{tab:mot15}, methods incorporating sparse flow or trajectories yielded a performance boost -- in particular, NOMT is a top-performing method published in 2015, achieving MOTA of 33.67\% on MOT15 and 46.42\% on MOT16. 
Interestingly, the first methods outperforming NOMT on \MOTNEW were published only in 2017 (AMIR \citep{amir} and NLLMP \citep{nllmpa}).

\PAR{Towards learning.} In 2015, we observed a clear trend towards utilizing learning to improve MOT. \\
LP\_SSVM \citep{lp_ssvm} demonstrates a significant performance boost by learning the parameters of linear cost association functions within a network flow tracking framework, especially when compared to methods using a similar optimization framework but hand-crafted association cues,~\eg~\citet{moticon}. 
The parameters are learned using structured SVM \citep{Taskar03NIPS}.
MDP \citep{mdp} goes one step further and proposes to learn track management policies (birth/death/association) by modeling object tracks as Markov Decision Processes \citep{Thrun05}. Standard MOT evaluation measures \citep{Stiefelhagen06CLE} are not differentiable. Therefore, this method relies on reinforcement learning to learn these policies. 
As can be seen in Tab.~\ref{tab:mot15}, this method outperforms the majority of methods published in 2015 by a large margin and surpasses 30\% MOTA on \MOTOLD.

In parallel, methods start leveraging the representational power of deep learning, initially by utilizing transfer learning.
MHT$\_$DAM \citep{mht_dam} learns to adapt appearance models online using multi-output regularized least squares. 
Instead of weak appearance features, such as color histograms, they extract base features for each object detection using a pre-trained convolutional neural network. 
With the combination of the powerful MHT tracking framework \citep{Reid:1979:MHT} and online-adapted features used for data association, this method surpasses MDP and attains over 32\% MOTA on \MOTOLD and 45\% MOTA on \MOTNEW. 
Alternatively, JMC \citep{jmc} and JointMC \citep{jointmc} use a pre-learned deep matching model to improve the pairwise affinity measures. 
All aforementioned methods leverage pre-trained models. 

\PAR{Learning appearance models.} The next clearly emerging trend goes in the direction of learning appearance models for data association in end-to-end fashion directly on the target (\ie, \MOTOLD, \MOTNEW, \MOTLAST) datasets.
SiameseCNN \citep{Leal:2016:CVPRW} trains a siamese convolutional neural network to learn spatio-temporal embeddings based on object appearance and estimated optical flow using contrastive loss \citep{Hadsell06CVPR}. 
The learned embeddings are then combined with contextual cues for robust data association. 
This method uses similar linear programming based optimization framework \citep{zhang08cvpr} compared to LP\_SSVM \citep{lp_ssvm}, however, it surpasses it significantly performance-wise, reaching 29\% MOTA on \MOTOLD. This demonstrates the efficacy of fine-tuning appearance models directly on the target dataset and utilizing convolutional neural networks. 
This approach is taken a step further with QuadMOT \citep{quadmot}, which similarly learns spatio-temporal embeddings of object detections.  However, they train their siamese network using quadruplet loss \citep{chen17cvpr} and learn to place embedding vectors of temporally-adjacent detections instances closer in the embedding space. These methods reach 33.42\% MOTA in \MOTOLD and 41.1\% on \MOTNEW.

The learning process, in this case, is supervised. 
Different from that, HCC \citep{hcc} learns appearance models in an unsupervised manner. To this end, they train their method using object trajectories obtained from the test set using offline correlation clustering-based tracking framework \citep{nllmpa}.  
TO \citep{to}, on the other hand, proposes to mine detection pairs over consecutive frames using single object trackers to learn affinity measures which are plugged into a network flow optimization tracking framework. Such methods have the potential to keep improving affinity models on datasets for which ground-truth labels are not available.
\PAR{Online appearance model adaptation.} The aforementioned methods only learn general appearance embedding vectors for object detection and do not adapt the tracking target appearance models online.  
Further performance is gained by methods that perform such adaptation online \citep{mht_dam, mht_blstm, dman, stam16}. 
MHT\_bLSTM \citep{mht_blstm} replaces the multi-output regularized least-squares learning framework of MHT\_DAM \citep{mht_dam} with a bi-linear LSTM and adapts both the appearance model as well as the convolutional filters in an online fashion. 
STAM \citep{stam16} and DMAN \citep{dman} employ an ensemble of single-object trackers (SOTs) that share a convolutional backbone and learn to adapt the appearance model of the targets online during inference. 
They employ a spatio-temporal attention model that explicitly aims to prevent drifts in appearance models due to occlusions and interactions among the targets. 
Similarly, KCF \citep{kcf} employs an ensemble of SOTs and updates the appearance model during tracking. To prevent drifts, they learn a tracking update policy using reinforcement learning. 
These methods achieve up to 38.9\% MOTA on \MOTOLD, 48.8\% on MOT16 (KCF), and 50.71\% on \MOTLAST (MHT\_DAM). Surprisingly, MHT\_DAM out-performs its bilinear-LSTM variant (MHT\_bLSTM achieves a MOTA of 47.52\%) on \MOTLAST.

\PAR{Learning to combine association cues.} A number of methods go beyond learning only the appearance model. 
Instead, these approaches learn to encode and combine heterogeneous association cues. SiameseCNN \citep{Leal:2016:CVPRW} uses gradient boosting to combine learned appearance embeddings with contextual features. AMIR \citep{amir} leverages recurrent neural networks in order to encode appearance, motion, pedestrian interactions and learns to combine these sources of information.  
STRN \citep{strn} proposes to leverage relational neural networks to learn to combine association cues, such as appearance, motion, and geometry.
RAR \citep{rar15pub} proposes recurrent auto-regressive networks for learning a generative appearance and motion model for data association. These methods achieve 37.57\% MOTA on \MOTOLD and 47.17\% on \MOTNEW.
\begin{figure*}[!ht]
    \centering
    \includegraphics[width=0.75\linewidth]{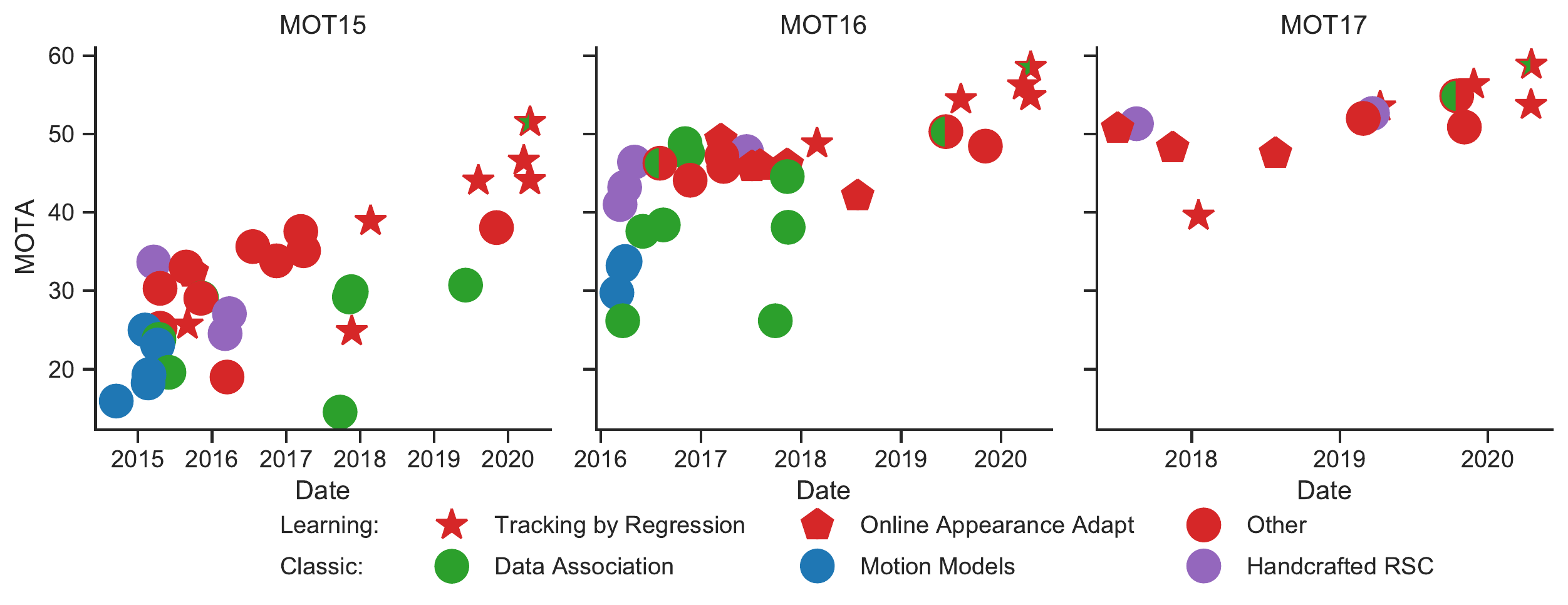}
    \caption{Overview of tracker performances measured by their date of submission time and model type category.}

    \label{fig:evolutionMOTA}

\end{figure*}
\PAR{Fine-grained detection.} A number of methods employ additional fine-grained detectors and incorporate their outputs into affinity measures, \eg, a head detector in the case of~FWT \citep{fwt}, or a body joint detectors in JBNOT \citep{jbnot}, which are shown to help significantly with occlusions. The latter attains 52.63\% MOTA on MOT17, which places it as the second-highest scoring method published in 2019. 

\PAR{Tracking-by-regression.} Several methods leverage ensembles of (trainable) single-object trackers (SOTs), used to regress tracking targets from the detected objects, utilized in combination with simple track management (birth/death) strategies.
    %
We refer to this family of models as MOT-by-SOT or tracking-by-regression. 
We note that this paradigm for MOT departs from the traditional view of the multi-object tracking problem in computer vision as a generalized assignment problem (or multi-dimensional assignment problem), \ie the problem of grouping object detections into a discrete set of tracks. 
Instead, methods based on target regression bring the focus back to the target state estimation. We believe the reasons for the success of these methods is two-fold: (i) rapid progress in learning-based SOT \citep{Held16ECCV, Li18CVPRb} that effectively leverages convolutional neural networks, and (ii) these methods can effectively utilize image evidence that is not covered by the given detection bounding boxes. 
Perhaps surprisingly, the most successful tracking-by-regression method, Tracktor \citep{tracktor}, does not perform online appearance model updates (\cf, STAM, DMAN \citep{stam16, dman} and KCF \citep{kcf}). 
Instead, it simply re-purposes the regression head of the Faster R-CNN \citep{Ren:2015:NIPS} detector, which is interpreted as the target regressor. 
This approach is most effective when combined with a motion compensation module and a learned re-identification module, attaining 46\% MOTA on MOT15 and 56\% on MOT16 and MOT17, outperforming methods published in 2019 by a large margin.

\begin{figure*}[!ht]
    \centering
     
    \vspace{1.5em}
    \includegraphics[width=0.32\linewidth]{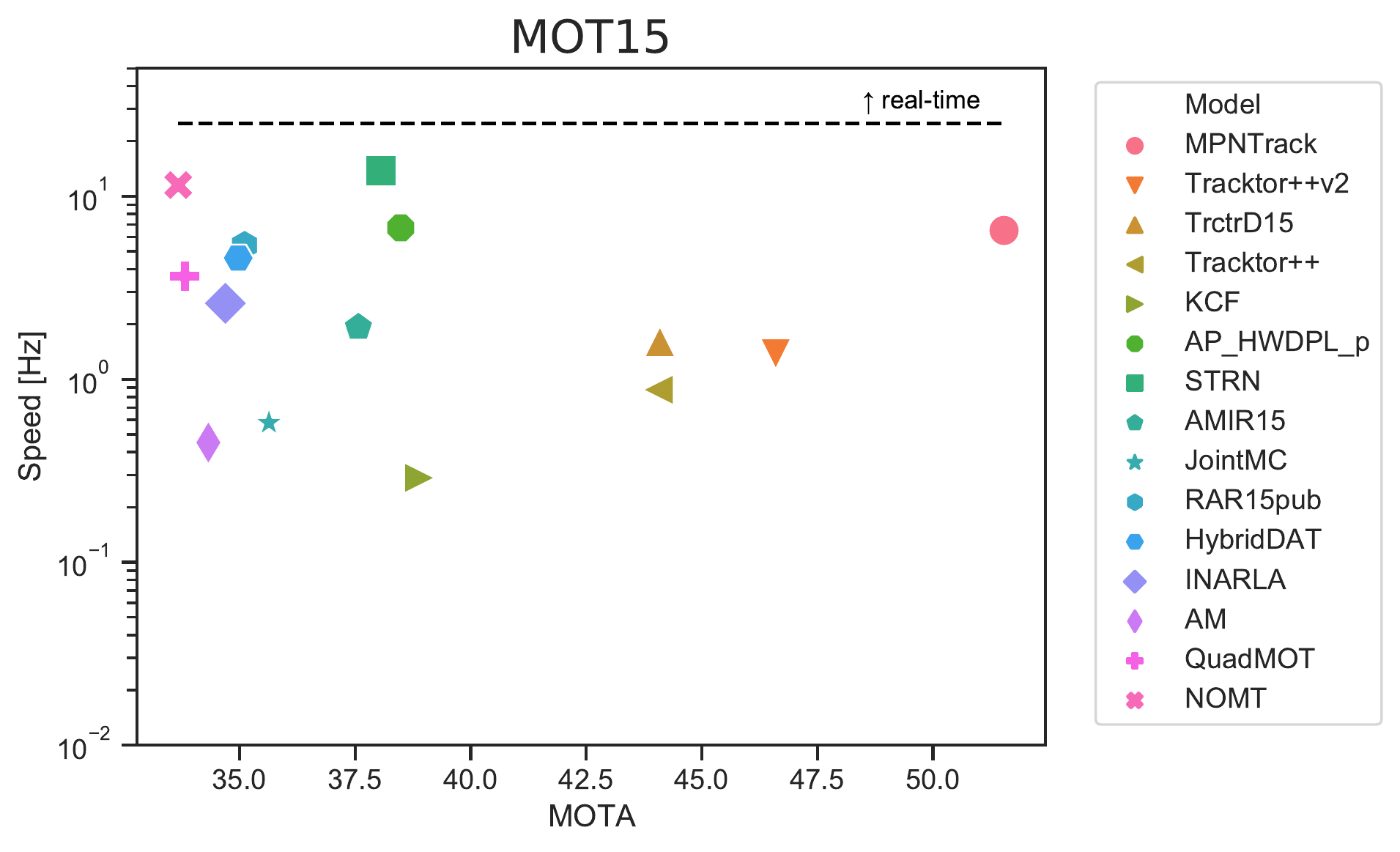}
    \hfill
    \includegraphics[width=0.32\linewidth]{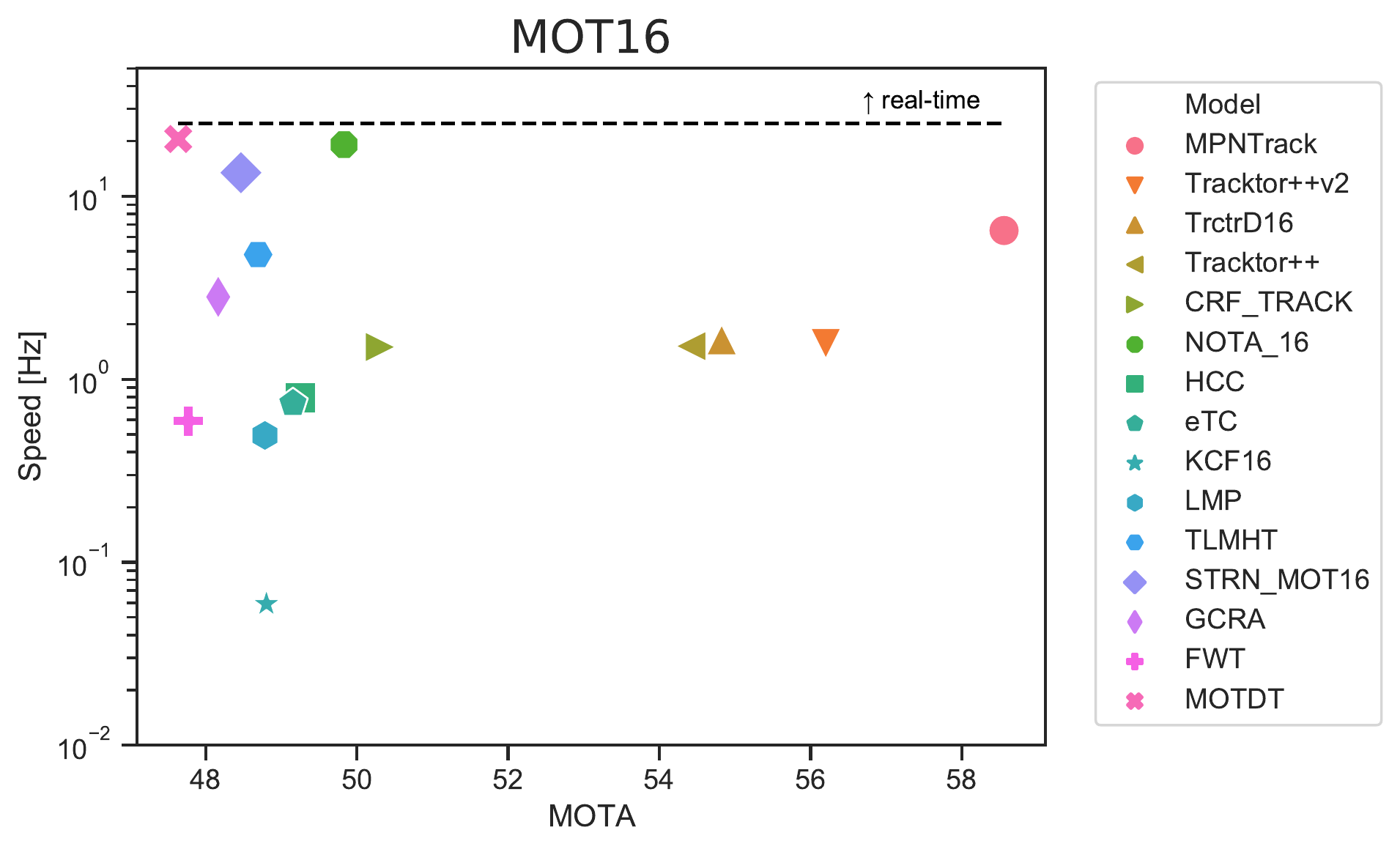}
    \hfill
    \includegraphics[width=0.32\linewidth]{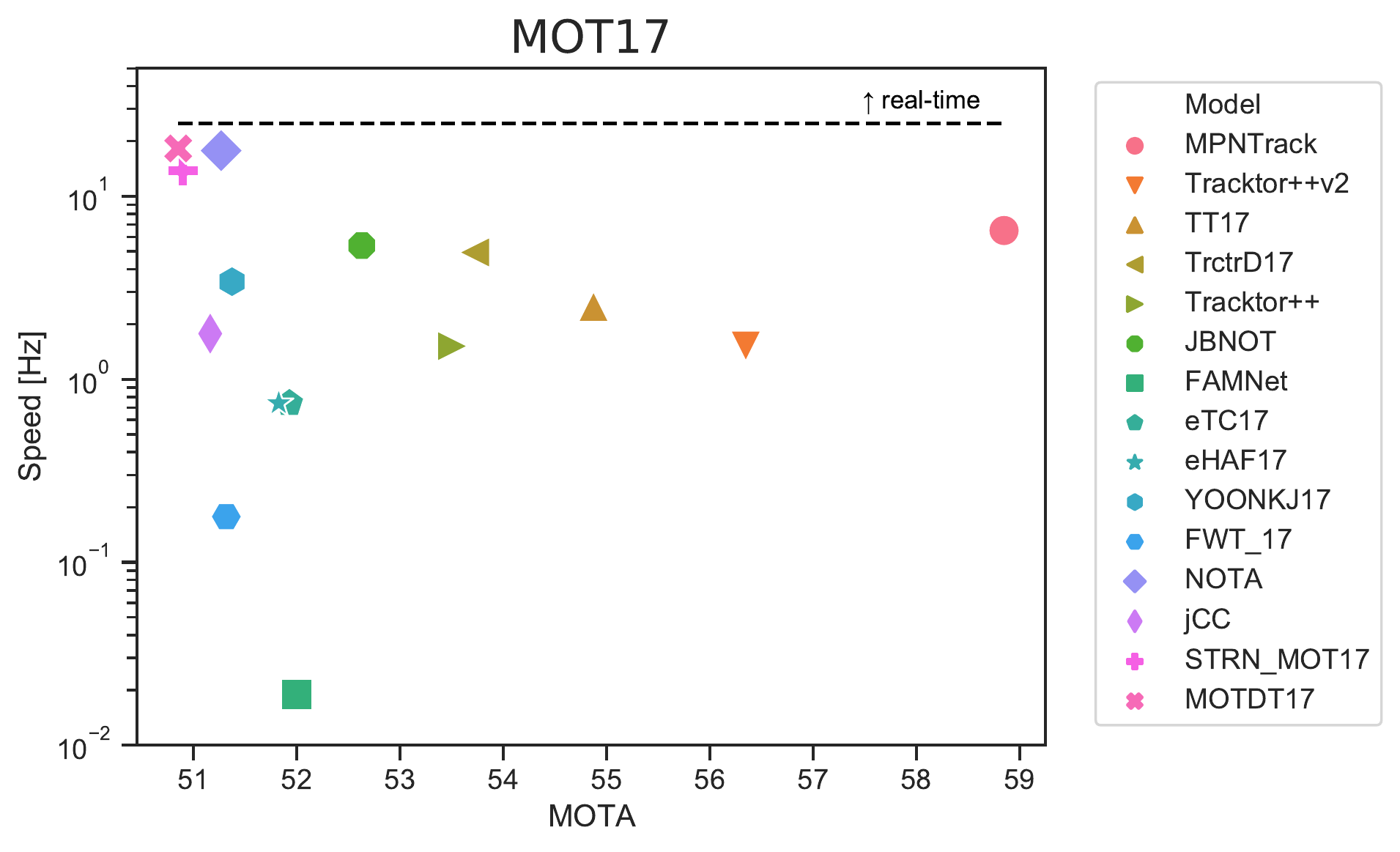}
  
    \caption{Tracker performance measured by MOTA versus processing efficiency in frames per second for \MOTOLD, \MOTNEW, and \MOTLAST on a log-scale. The latter is only indicative of the true value and has not been measured by the benchmark organizers. See text for details.}
    \label{fig:MOTA-runtime}
\end{figure*}

\PAR{Towards end-to-end learning.}
Even though tracking-by-regression methods brought substantial improvements, they are not able to cope with larger occlusions gaps. 
To combine the power of graph-based optimization methods with learning, MPNTrack \citep{mpntrack} proposes a method that leverages message-passing networks \citep{battaglia16nips} to directly learn to perform data association via edge classification. 
By combining the regression capabilities of Tracktor \citep{tracktor} with a learned discrete neural solver, MPNTrack establishes a new state of the art, effectively using the best of both worlds -- target regression and discrete data association.
This method is the first one to surpass MOTA above 50\% on \MOTOLD. On the \MOTNEW and \MOTLAST it attains a MOTA of 58.56\% and 58.85\%, respectively.
Nonetheless, this method is still not fully end-to-end trained, as it requires a projection step from the solution given by the graph neural network to the set of feasible solutions according to the network flow formulation and constraints. 

Alternatively, \citep{xiang2020end} uses MHT framework \citep{Reid:1979:MHT} to link tracklets, while iteratively re-evaluating appearance/motion models based on progressively merged tracklets. This approach is one of the top on \MOTLAST, achieving 54.87\% MOTA.

In the spirit of combining optimization-based methods with learning, \citet{zhang2020long} revisits CRF-based tracking models and learns unary and pairwise potential functions in an end-to-end manner. On \MOTNEW, this method attains MOTA of 50.31\%.

We do observe trends towards learning to perform end-to-end MOT. To the best of our knowledge, the first method attempting this is RNN\_LSTM \citep{rnn_lstm}, which jointly learns motion affinity costs and to perform bi-partite detection association using recurrent neural networks (RNNs).
FAMNet \citep{famnet} uses a single network to extract appearance features from images, learns association affinities, and estimates multi-dimensional assignments of detections into object tracks. The multi-dimensional assignment is performed via a differentiable network layer that computes rank-1 estimation of the assignment tensor, which allows for back-propagation of the gradient. They perform learning with respect to binary cross-entropy loss between predicted assignments and ground-truth. 

All aforementioned methods have one thing in common -- they optimize network parameters with respect to proxy losses that do not directly reflect tracking quality, most commonly measured by the CLEAR-MOT evaluation measures \citep{Stiefelhagen06CLE}. 
To evaluate MOTA, the assignment between track predictions and ground truth needs to be established; this is usually performed using the Hungarian algorithm \citep{Kuhn55NRLQ}, which contains non-differentiable operations. 
To address this discrepancy DeepMOT \citep{trctrd} proposes the missing link -- a differentiable matching layer that allows expressing a soft, differentiable variant of MOTA and MOTP. 

\PAR{Conclusion.} In summary, we observed that after an initial focus on developing algorithms for discrete data association \citep{zhang08cvpr, Pirsiavash11CVPR, gmmcp, lttsc-crf}, the focus shifted towards hand-crafting powerful affinity measures \citep{nomt, oicf, moticon}, followed by large improvements brought by learning powerful affinity models \citep{Leal:2016:CVPRW, quadmot, mdp, lp_ssvm}.

In general, the major outstanding trends we observe in the past years all leverage the representational power of deep learning for learning association affinities, learning to adapt appearance models online \citep{mht_blstm, stam16, dman, kcf} and learning to regress tracking targets \citep{stam16, dman, kcf, tracktor}. Figure~\ref{fig:evolutionMOTA} visualizes the promise of deep learning for tracking by plotting the performance of submitted models over time and by type.

The main common components of top-performing methods are: (i) learned single-target regressors (single-object trackers), such as \citep{Held16ECCV, Li18CVPRb}, and (ii) re-identification modules \citep{tracktor}. 
These methods fall short in bridging large occlusion gaps. 
To this end, we identified Graph Neural Network-based methods \citep{mpntrack} as a promising direction for future research. 
We observed the emergence of methods attempting to learn to track objects in end-to-end fashion instead of training individual modules of tracking pipelines \citep{famnet, trctrd, rnn_lstm}. 
We believe this is one of the key aspects to be addressed to further improve performance and expect to see more approaches leveraging deep learning for that purpose.

\begin{figure*}
    \centering
    \includegraphics[width=0.32\linewidth]{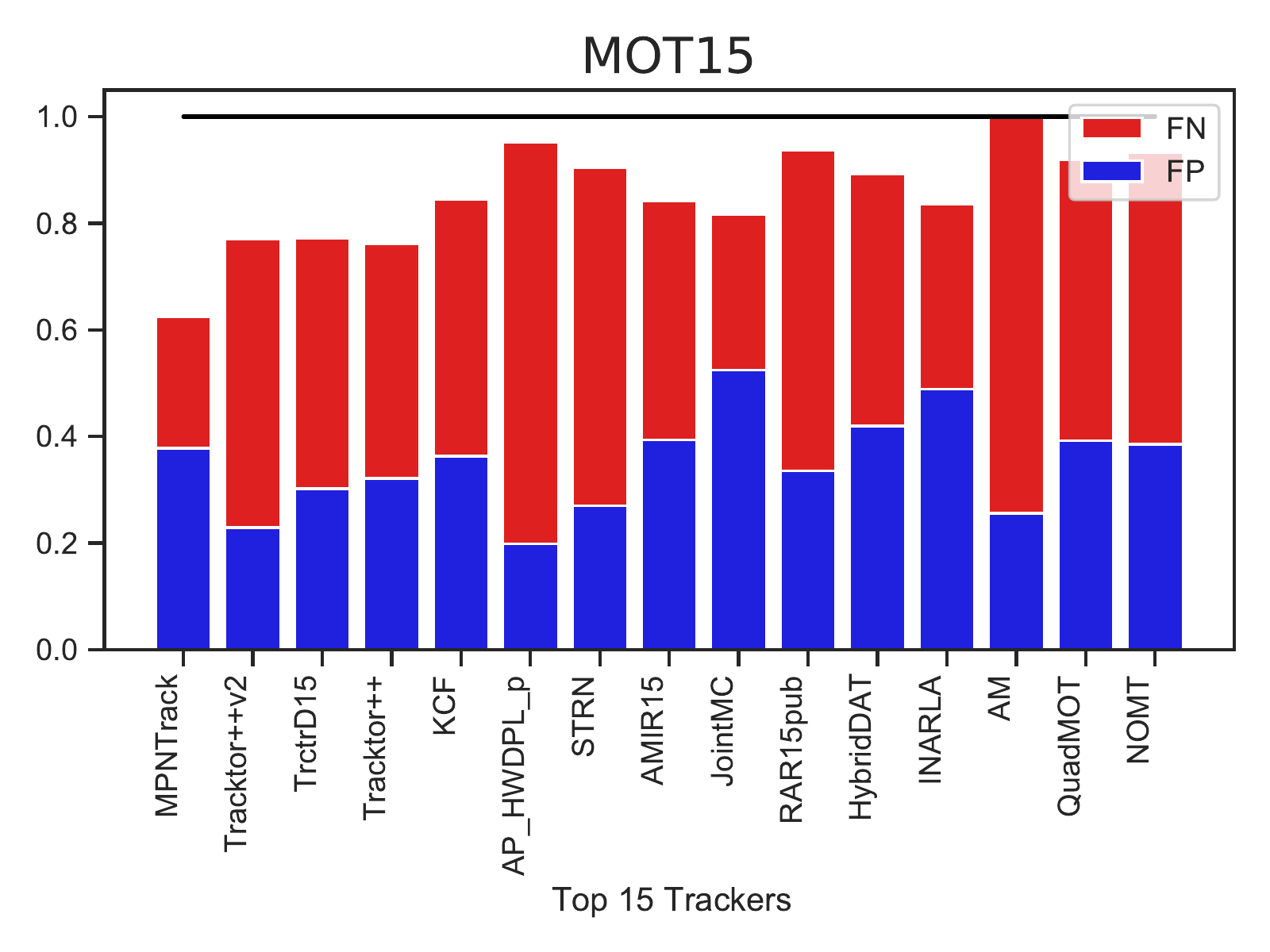}    
    \hfill
    \includegraphics[width=0.32\linewidth]{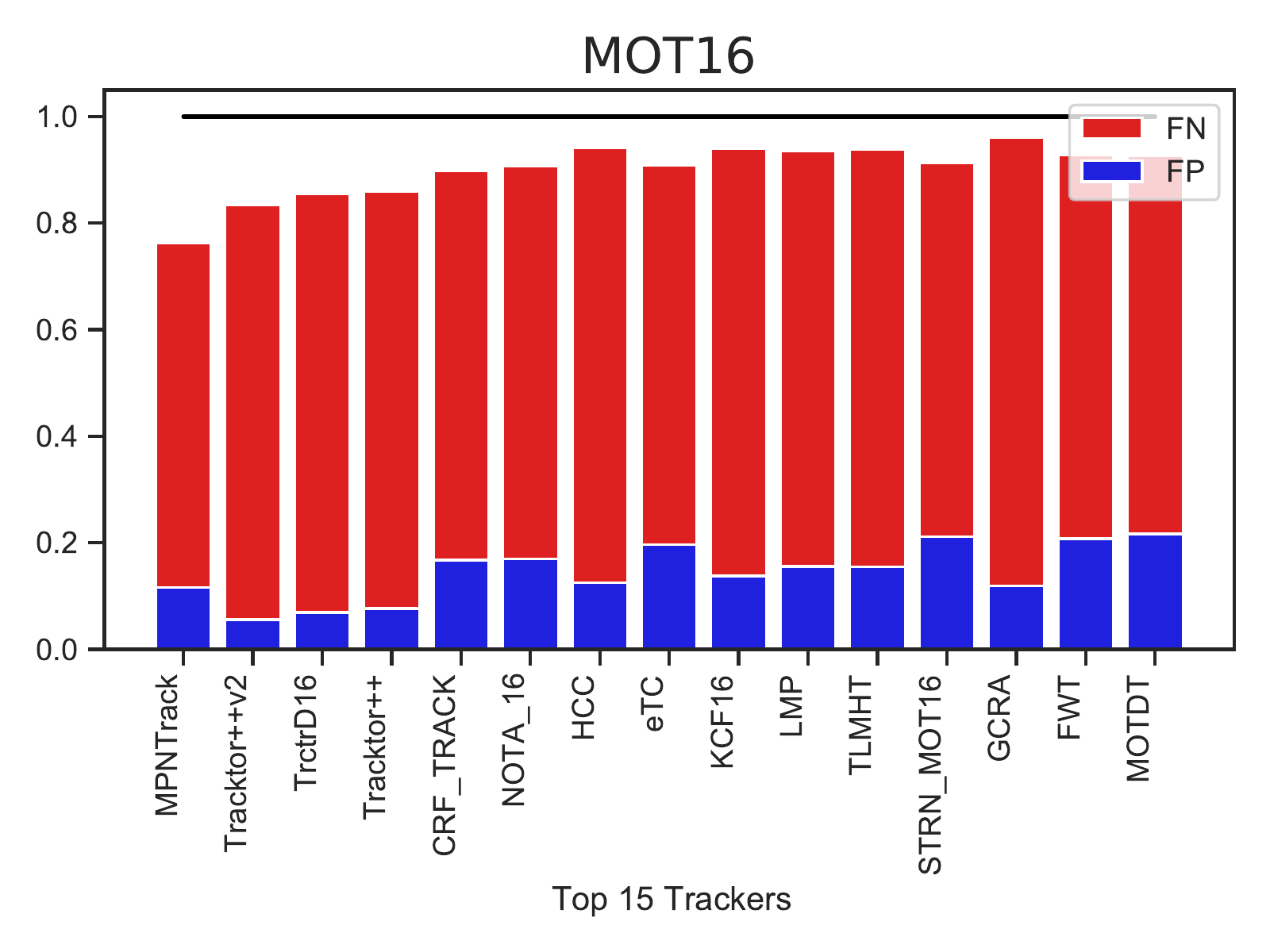} 
    \hfill
    \includegraphics[width=0.32\linewidth]{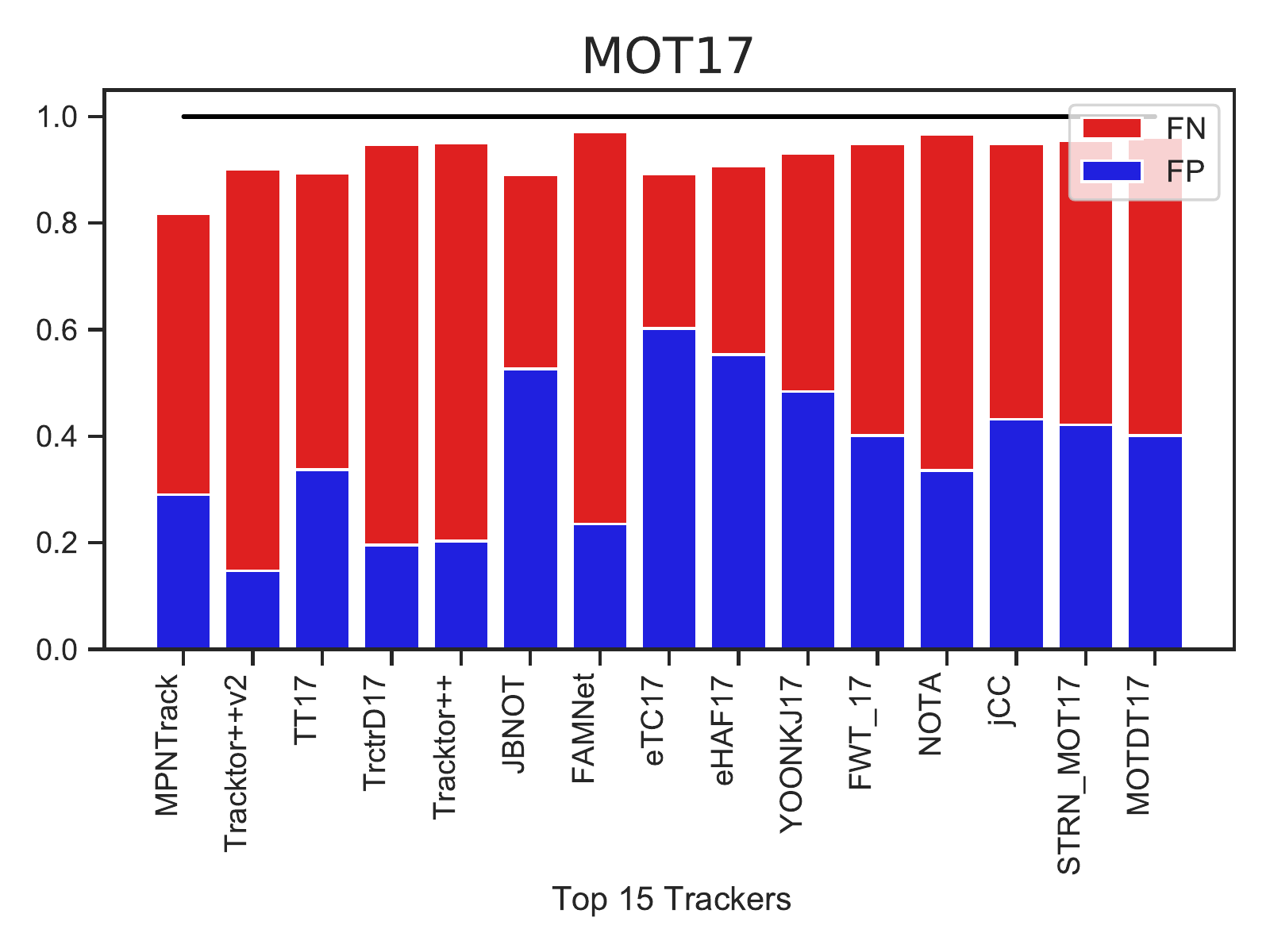}


    %
    \caption{Detailed error analysis. The plots show the error ratios for trackers \wrt detector (taken at the lowest confidence threshold), for two types of errors: false positives (FP) and false negatives (FN). Values above $1$ indicate a higher error count for trackers than for detectors. Note that most trackers concentrate on removing false alarms provided by the detector at the cost of eliminating a few true positives, indicated by the higher FN count.}
    \label{fig:FNFP}
\end{figure*}

\subsection{Runtime Analysis}
\label{sec:runtime}
Different methods require a varying amount of computational resources to track multiple targets.
Some methods may require large amounts of memory while others need to be executed on a GPU. 
For our purpose, we ask each benchmark participant to provide the number of seconds required to produce the results on the entire dataset, regardless of the computational resources used. %
It is important to note that the resulting numbers are therefore only indicative of each approach and are not immediately comparable to one another.

Figure~\ref{fig:MOTA-runtime} shows the relationship between each submission's performance measured by MOTA and its efficiency in terms of frames per second, averaged over the entire dataset. 
There are two observations worth pointing out. 
First, the majority of methods are still far below real-time performance, which is assumed at 25 Hz. 
Second, the average processing rate $\sim 5$ Hz does not differ much between the different sequences, which suggests that the different object densities (9 ped./fr. in \MOTOLD and 26 ped./fr. in \MOTNEW/\MOTLAST) do not have a large impact on the speed the models. One explanation is that novel learning methods have an efficient forward computation, which does not vary much depending on the number of objects. This is in clear contrast to classic methods that relied on solving complex optimization problems at inference, which increased computation significantly as the pedestrian density increased.
However, this conclusion has to be taken with caution because the runtimes are reported by the users on a trust base and cannot be verified by us.

\subsection{Error Analysis}
\label{sec:error-analysis}

As we now, different applications have different requirements, e.g., for surveillance it is critical to have few false negatives, while for behavior analysis, having a false positive can mean computing wrong motion statistics. In this section, we take a closer look at the most common errors made by the tracking approaches. This simple analysis can guide researchers in choosing the best method for their task. 
In~\Fig~\ref{fig:FNFP}, we show the number of false negatives (FN, blue) and false positives (FP, red) created by the trackers on average with respect to the number of FN/FP of the object detector, used as an input. %
A ratio below $1$ indicates that the trackers have improved in terms of FN/FP over the detector. 
We show the performance of the top 15 trackers, averaged over sequences. We order them according to MOTA from left to right in decreasing order.

We observe all top-performing trackers reduce the amount of FPs and FNs compared to the public detections. 
While the trackers reduce FPs significantly, FNs are decreased only slightly. %
Moreover, we can see a direct correlation between the FN and tracker performance, especially for \MOTNEW and \MOTLAST datasets, since the number of FNs is much larger than the number of FPs. 
The question is then, why are methods not focusing on reducing FNs? It turns out that ``filling the gaps`` between detections, what is commonly thought trackers should do, is not an easy task.

It is not until 2018 that we see methods drastically decreasing the number of FNs, and as a consequence, MOTA performance leaps forward. As shown in~\Fig~\ref{fig:evolutionMOTA}, this is due to the appearance of learning-based tracking-by-regression methods \citep{tracktor, mpntrack, stam16, dman}.
Such methods decrease the number of FNs the most by effectively using image evidence not covered by detection bounding boxes and regressing targets to areas where they are visible but missed by detectors. 
This brings us back to the common wisdom that trackers should be good at ``filling the gaps`` between detections.

Overall, it is clear that \MOTLAST still presents a challenge both in terms of detection as well as tracking. 
It will require significant further future efforts to bring performance to the next level. 
In particular, the next challenge that future methods will need to tackle is bridging large occlusion gaps, which can not be naturally resolved by methods performing target regression, as these only work as long as the target is (partially) visible.

 \section{Conclusion and Future Work}
 \label{sec:conclusion}
 
We have introduced \MOTChallenge, a standardized benchmark for a fair evaluation of single-camera multi-person tracking methods. We presented its first two data releases with about 35,000 frames of footage and almost 700,000 annotated pedestrians.
Accurate annotations were carried out following a strict protocol, and extra classes such as vehicles, sitting people, reflections, or distractors were also annotated in the second release to provide further information to the community.  

We have further analyzed the performance of 101 trackers; 73 \MOTOLD, 74 \MOTNEW, and 57 on \MOTLAST obtaining several insights. 
In the past, at the center of vision-based MOT were methods focusing on global optimization for data association. Since then, we observed that large improvements were made by hand-crafting strong affinity measures and leveraging deep learning for learning appearance models, used for better data association. 
More recent methods moved towards directly regressing bounding boxes, and learning to adapt target appearance models online. 
As the most promising recent trends that hold a large potential for future research, we identified the methods that are going in the direction of learning to track objects in an end-to-end fashion, combining optimization with learning.

We believe our Multiple Object Tracking Benchmark and the presented systematic analysis of existing tracking algorithms will help identify the strengths and weaknesses of the current state of the art and shed some light into promising future research directions.

\noindent{\bf Acknowledgements.} We would like to specially acknowledge 
Siyu Tang, Sarah Becker, Andreas Lin, and Kinga Milan for their help in 
the annotation process. 
We thank Bernt Schiele for helpful discussions
and important insights into benchmarking. 
IDR gratefully acknowledges the support of the 
Australian Research Council through FL130100102. 
LLT acknowledges the support of the Sofja Kovalevskaja Award from the Humboldt Foundation, endowed by the Federal Ministry of Education and Research. DC acknowledges the support of the ERC Consolidator Grant {\em 3D Reloaded}.


\bibliographystyle{abbrvnat}      

\bibliography{refs, abbrev_short}
\appendix
\clearpage 
\addappheadtotoc
\appendixpage

\section{Benchmark Submission}
\label{sec:benchmark-structure}
Our benchmark consists of the database and evaluation server on one hand, and the website as the user interface on the other. 
It is open to everyone who respects the submission policies (see next section). Before participating, every user is required to create an account, providing an institutional and not a generic e-mail address\footnote{For accountability and to prevent abuse by using several email accounts.}.

After registering, the user can create a new tracker with a unique name and enter all additional details. It is 
mandatory to indicate: 
\begin{itemize}
\item the full name and a brief description of the method 
\item a reference to the publication of the method, if already existing, 
\item whether the method operates online or on a batch of frames and whether the source code is publicly available,
\item whether only the provided or also external training and detection 
data were used. 
\end{itemize}
After creating all details of a new tracker, it is possible to assign open challenges to this tracker and submit results to the different benchmarks. To participate in a challenge the user has to provide the following information for each challenge they want to submit to:
\begin{itemize}
\item name of the challenge in which the tracker will be participating,
\item a reference to the publication of the method, if already existing, 
\item the \emph{total} runtime in seconds for computing the results for the test sequences and the hardware used, and 
\item whether only provided data was used for training, or also data from other sources were involved.
\end{itemize}
The user can then submit the results to the challenge in the format described in~\Sec~\ref{sec:data-format}.
The tracking results are automatically evaluated and appear on the user's profile. 
The results are \emph{not} automatically displayed in the public ranking table. 
The user can decide at any point in time to make the results public. Results can be published anonymously,~\eg, to enable a blind review process for a corresponding paper. 
In this case, we ask to provide the venue and the paper ID or a similar unique reference.
We request that a proper reference to the method's description is added upon acceptance of the paper. 
Anonymous entries are hidden from the benchmark after six months of inactivity.

The trackers and challenge meta information such as description, project page, runtime, or hardware can be edited at any time. 
Visual results of all public submissions, as well as annotations and detections, can be viewed and downloaded on the individual result pages of the corresponding tracker. 

\subsection{Submission Policy}
The main goal of this benchmark is to provide a platform that allows for
objective performance comparison of multiple target tracking 
approaches on real-world data. 
Therefore, we introduce a few simple guidelines that must be followed by all participants.

\noindent {\bf Training.} Ground truth is only provided for the training sequences.
It is the participant's own responsibility to find the best setting
using \emph{only} the training data.
The use of additional training data must be indicated during submission and will be visible in the public ranking table.
The use of ground truth labels on the test data is strictly forbidden. This or any other misuse of the benchmark will lead to the deletion of the participant's account and their results.

\noindent {\bf Detections.} We also provide a unique set of detections
(see \Sec \ref{sec:detections}) for each sequence. 
We expect all tracking-by-detection algorithms to use the given detections. 
In case the user wants to present results with another set of
detections or is not using detections at all, this should be clearly stated during submission and will also be displayed in the results table.

\noindent {\bf Submission frequency.} 
Generally, we expect one single submission for a particular method per benchmark.
If for any reason the user needs to re-compute and re-submit the results (\eg due to a bug discovered in the implementation), they may do so after a waiting period of 72 hours 
after the last submission to submit to the same challenge with any of their trackers.
This policy should discourage the use of the
benchmark server for training and parameter tuning on the test data. The number of submissions is counted and displayed for each method. We allow a maximum number of 4 submissions per tracker and challenge.  We allow a user to create several tracker instances for different tracking models. However, a user can only create a new tracker every 30 days.
Under \emph{no} circumstances must anyone create a second account and 
attempt to re-submit in order to bypass the waiting period. 
Such behavior will lead to the deletion of the accounts and exclusion of the user from participating in the benchmark.

\subsection{Challenges and Workshops}
We have two modalities for submission: the general open-end challenges and the special challenges.
The main challenges, 2D MOT 2015, 3D MOT 2015, MOT16, and MOT17 are always open for submission and are nowadays the standard evaluation platform for multi-target tracking methods submitting to computer vision conferences such as CVPR, ICCV or ECCV.

Special challenges are similar in spirit to the widely known PASCAL VOC series~\citep{Everingham15ijcv}, or the 
ImageNet competitions~\citep{Russakovsky15ijcv}. 
Each special challenge is linked to a workshop. 
The first edition of our series was the WACV 2015 Challenge that consisted of six outdoor sequences with both moving and static cameras, followed by the 2nd edition held in conjunction with ECCV 2016 on which we evaluated methods on the new \MOTNEW sequences. The \MOTLAST sequences were presented in the Joint Workshop on Tracking and Surveillance in conjunction with the Performance Evaluation of Tracking and Surveillance (PETS)~\citep{Ferryman09pets, Ferryman10avss} benchmark at the Conference on Vision and Pattern Recognition (CVPR) in 2017.
The results and winning methods were presented during the respective workshops.
Submission to those challenges is open only for a short period of time, \ie, there is a fixed submission deadline for all participants.
Each method must have an accompanying paper presented at the workshop. The results of the methods are kept {hidden} until the date of the workshop itself when the winning method is revealed and a prize is awarded.
\section{MOT 15}
We have compiled a total of 22 sequences, of which we use half for training and half for testing. The annotations of the testing sequences are not released in order to avoid (over)fitting of the methods to the specific sequences. Nonetheless, the test data contains over 10 minutes of footage and 61440 annotated bounding boxes, therefore, it is hard for researchers to over-tune their algorithms on such a large amount of data. This is one of the major strengths of the benchmark. We classify the sequences according to:

We classify the sequences according to: \\

\begin{itemize}
    \item \textit{Moving or static camera}: the camera can be held by a person, placed on a stroller~\citep{Ess08cvpr} or on a car~\citep{Geiger12kitti}, or can be positioned fixed in the scene.
    \item \textit{Viewpoint}: the camera can overlook the scene from a high position, a medium position (at pedestrian's height), or at a low position. 
    \item \textit{Weather}: the illumination conditions in which the sequence was taken. Sequences with strong shadows and saturated parts of the image make tracking challenging, while night sequences contain a lot of motion blur, which is often a problem for detectors. Indoor sequences contain a lot of reflections, while the sequences classified as normal do not contain heavy illumination artifacts that potentially affect tracking.\\
\end{itemize}

\begin{figure}[htpb]
    \centering
    \includegraphics[width=1\linewidth]{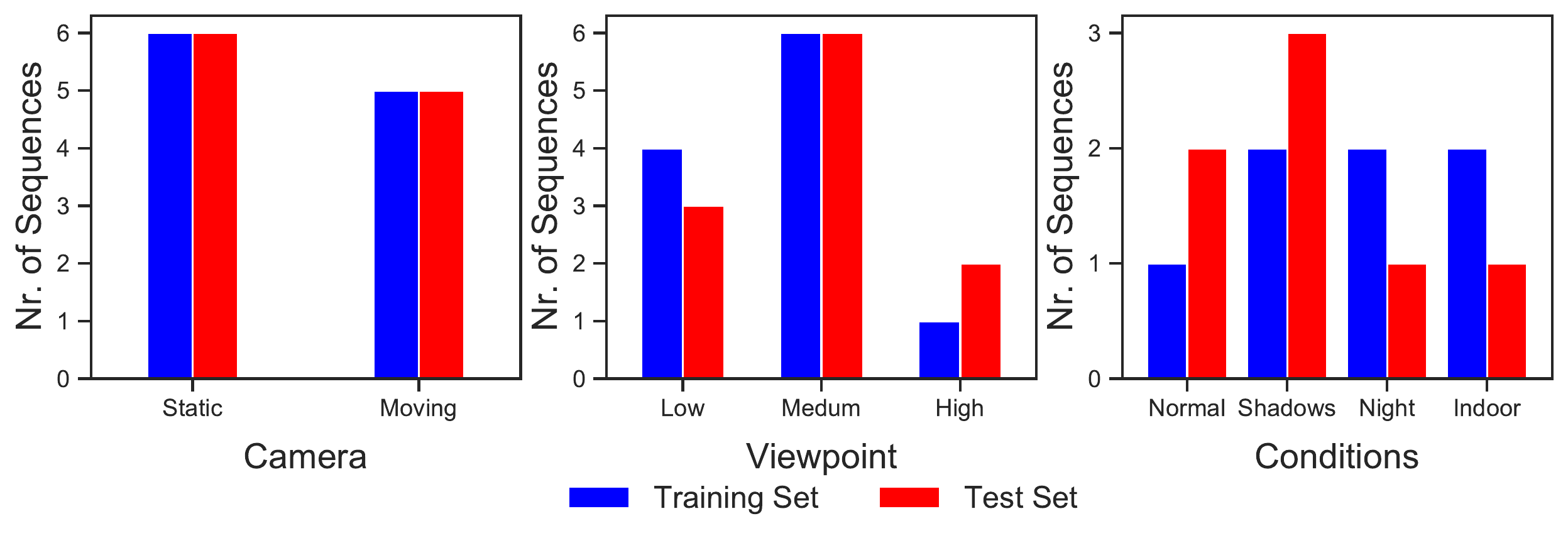}
    \caption{Comparison histogram between training and testing sequences of \textit{static} vs. \textit{moving} camera, camera \textit{viewpoint}: low, medium or high, \textit{conditions}: normal, shadows, night or indoor.}
    \label{fig:2Ddata}
\end{figure}

We divide the sequences into training and testing to have a balanced distribution, as shown in Figure~\ref{fig:2Ddata}. 
\begin{table*}[tbp]
    \begin {center}
     \caption{Overview of the sequences currently included in the \MOTOLD benchmark.}
    \label{tab:dataoverview15}
    \tiny
    \begin{adjustbox}{max width=\textwidth}
    \begin{tabular}{l| c| c| c| c| c| c| c |c | c | c}
    \toprule
    \multicolumn{11}{c}{\bf Training sequences} \\ 
    Name & FPS & Resolution & Length & Tracks & Boxes & Density & 3D & Camera & Viewpoint & Conditions \\ 
    \midrule
    TUD-Stadtmitte~\citep{Andriluka:2010:CVPR} & 25 & 640x480 & 179 (00:07) & 10 & 1156 & 6.5 & yes  & static & medium & normal \\
    TUD-Campus~\citep{Andriluka:2010:CVPR} & 25 & 640x480 & 71 (00:03) & 8 & 359 & 5.1 & no & static & medium & normal \\
    PETS09-S2L1~\citep{Ferryman10avss} & 7 & 768x576 & 795 (01:54) & 19 & 4476 & 5.6 & yes  & static & high & normal  \\
    ETH-Bahnhof~\citep{Ess08cvpr} & 14 & 640x480 & 1000 (01:11) & 171 & 5415 & 5.4 & yes  & moving  & low & normal \\
    ETH-Sunnyday\citep{Ess08cvpr} & 14 & 640x480 & 354 (00:25) & 30 & 1858 & 5.2 & yes  & moving  & low & shadows \\
    ETH-Pedcross2\citep{Ess08cvpr} & 14 & 640x480 & 840 (01:00) & 133 & 6263 & 7.5 & no & moving  & low & shadows \\
    ADL-Rundle-6 (new) & 30 & 1920x1080 & 525 (00:18) & 24 & 5009 & 9.5 & no & static & low & indoor  \\
    ADL-Rundle-8 (new) & 30 & 1920x1080 & 654 (00:22) & 28 & 6783 & 10.4 & no & moving  & medium & night \\
    KITTI-13~\citep{Geiger12kitti} & 10 & 1242x375 & 340 (00:34) & 42 & 762 & 2.2 & no & moving  & medium & shadows  \\
    KITTI-17~\citep{Geiger12kitti} & 10 & 1242x370 & 145 (00:15) & 9 & 683 & 4.7 & no & static & medium & shadows \\
    Venice-2 (new) & 30 & 1920x1080 & 600 (00:20) & 26 & 7141 & 11.9 & no & static & medium & normal \\
    \midrule %
    \multicolumn{3}{c|}{\bf Total training} & {\bf 5503 (06:29)} & {\bf 500} & {\bf 39905} & {\bf 7.3} & & & &   \\
    \bottomrule %
    %
    %
    \multicolumn{11}{c}{\vspace{1em}} \\
    \toprule
    \multicolumn{11}{c}{\bf Testing sequences} \\ 
    Name & FPS & Resolution & Length & Tracks & Boxes & Density & 3D &  Camera & Viewpoint & Conditions \\ 
    \midrule
    TUD-Crossing~\citep{PoseTrack} & 25 & 640x480 & 201 (00:08) & 13 & 1102 & 5.5 & no & static & medium & normal \\   
    PETS09-S2L2~\citep{Ferryman10avss} & 7 & 768x576 & 436 (01:02) & 42 & 9641 & 22.1 & yes  & static & high & normal \\
    ETH-Jelmoli~\citep{Ess08cvpr} & 14 & 640x480 & 440 (00:31) & 45 & 2537 & 5.8 & yes  & moving  & low & shadows \\
    ETH-Linthescher~\citep{Ess08cvpr} & 14 & 640x480 & 1194 (01:25) & 197 & 8930 & 7.5 & yes  & moving  & low & shadows \\
    ETH-Crossing~\citep{Ess08cvpr} & 14 & 640x480 & 219 (00:16) & 26 & 1003 & 4.6 & no & moving  & low & normal \\
    AVG-TownCentre~\citep{Benfold:2011:CVPR} & 2.5 & 1920x1080 & 450 (03:45) & 226 & 7148 & 15.9 & yes  & static & high & normal \\
    ADL-Rundle-1 (new) & 30 & 1920x1080 & 500 (00:17) & 32 & 9306 & 18.6 & no & moving  & medium & normal \\
    ADL-Rundle-3 (new) & 30 & 1920x1080 & 625 (00:21) & 44 & 10166 & 16.3 & no & static & medium & shadows \\
    KITTI-16~\citep{Geiger12kitti} & 10 & 1242x370 & 209 (00:21) & 17 & 1701 & 8.1 & no & static & medium & shadows \\
    KITTI-19~\citep{Geiger12kitti} & 10 & 1242x374 & 1059 (01:46) & 62 & 5343 & 5.0 & no & moving  & medium & shadows  \\ 
    Venice-1 (new) & 30 & 1920x1080 & 450 (00:15) & 17 & 4563 & 10.1 & no & static & medium & normal \\ 
    \midrule
    \multicolumn{3}{c|}{\bf Total testing} & {\bf 5783 (10:07)} & {\bf 721} & {\bf 61440} & {\bf 10.6} & & & & \\
    \bottomrule 
    \end{tabular}
    \end{adjustbox}
    \end{center}

\end{table*}

\subsection{Data format}
\label{sec:data-format}

All images were converted to JPEG and named sequentially to a 6-digit file name (\eg~000001.jpg). 
Detection and annotation files are simple comma-separated value (CSV) files. 
Each line represents one object instance, and it contains 10 values as shown in~\Tab~\ref{tab:dataformat}.

The first number indicates in which frame the object appears, while the second number identifies that object as belonging to a trajectory by assigning a unique ID (set to $-1$ in a detection file, as no ID is assigned yet). 
Each object can be assigned to only one trajectory.
The next four numbers indicate the position of the bounding box of the pedestrian in 2D image coordinates. The position is indicated by the top-left corner as well as the width and height of the bounding box.
This is followed by a single number, which in the case of detections denotes their confidence score.
The last three numbers indicate the 3D position in real-world coordinates of the pedestrian. This position represents the feet of the person. In the case of 2D tracking, these values will be ignored and can be left at $-1$.
\begin{table*}[hbt]
\begin {center}
\caption{Data format for the input and output files, both for detection and annotation files.}
\label{tab:dataformat}
\tiny
 \begin{tabular}{ c | c| p{13cm}}
 \toprule
     \bf Position & \bf Name & \bf Description\\ 
          \hline 
    1 & Frame number & Indicate at which frame the object is present\\
 2 & Identity number & Each pedestrian trajectory is identified by a
 unique ID ($-1$ for detections)\\
 3 & Bounding box left &  Coordinate of the top-left corner of the pedestrian bounding box\\
  4 & Bounding box top & Coordinate of the top-left corner of the pedestrian bounding box \\
 5 & Bounding box width & Width in pixels of the pedestrian bounding box\\
 6 & Bounding box height & Height in pixels of the pedestrian bounding box\\
 7 & Confidence score & Indicates how confident the detector is that this instance is a pedestrian. 
 For the ground truth and results, it acts as a flag whether the entry is to be considered.  \\
  8 &  $x$ & 3D $x$ position of the pedestrian in real-world
  coordinates ($-1$ if not available)\\
 9 & $y$ &  3D $y$ position of the pedestrian in real-world coordinates ($-1$ if not available)\\
 10 & $z$ & 3D $z$ position of the pedestrian in real-world
 coordinates  ($-1$ if not available)\\
      \bottomrule 
    \end{tabular}
  \end{center}
    
\end{table*}

\noindent An example of such a detection 2D file is:

\begin{samepage}
    \begin{center}
    \begin{footnotesize}
    \texttt{1, -1, 794.2, 47.5, 71.2, 174.8, 67.5, -1, -1, -1}\nopagebreak\\
    \texttt{1, -1, 164.1, 19.6, 66.5, 163.2, 29.4, -1, -1, -1}\nopagebreak\\
    \texttt{1, -1, 875.4, 39.9, 25.3, 145.0, 19.6, -1, -1, -1}\nopagebreak\\
    \texttt{2, -1, 781.7, 25.1, 69.2, 170.2, 58.1, -1, -1, -1}\nopagebreak\\
    \end{footnotesize}
    \end{center}
\end{samepage}

\noindent For the ground truth and results files, the 7$^\text{th}$ value (confidence score) acts as a flag whether the entry is to be considered. A value of 0 means that this particular instance is ignored in the evaluation, while a value of 1 is used to mark it as active. 
An example of such an annotation 2D file is:

\begin{samepage}
    \begin{center}
    \begin{footnotesize}
    \texttt{1, 1, 794.2, 47.5, 71.2, 174.8, 1, -1, -1, -1}\nopagebreak\\
    \texttt{1, 2, 164.1, 19.6, 66.5, 163.2, 1, -1, -1, -1}\nopagebreak\\
    \texttt{1, 3, 875.4, 39.9, 25.3, \enspace35.0, 0, -1, -1, -1}\nopagebreak\\
    \texttt{2, 1, 781.7, 25.1, 69.2, 170.2, 1, -1, -1, -1}\nopagebreak\\
    \end{footnotesize}
    \end{center}
\end{samepage}

\noindent In this case, there are 2 pedestrians in the first frame of the sequence, with identity tags 1, 2. The third pedestrian is too small and therefore not considered, which is indicated with a flag value (7$^\text{th}$ value) of 0. In the second frame, we can see that pedestrian 1 remains in the scene. Note, that since this is a 2D annotation file, the 3D positions of the pedestrians are ignored and therefore are set to -1.
All values including the bounding box are 1-based, \ie the top left corner corresponds to $(1,1)$.
s

To obtain a valid result for the entire benchmark, a separate CSV file following the format described above must be created for each sequence and called\newline \texttt{``Sequence-Name.txt''}. All files must be compressed into a single zip file that can then be uploaded to be evaluated.

\section{MOT16 and MOT17 Release}
Table~\ref{tab:dataoverview16} presents an overview of the \MOTNEW and \MOTLAST dataset.
\subsection{Annotation rules}
\label{sec:anno-rules}

We follow a set of rules to annotate every moving person or vehicle within each sequence with a bounding box as accurately as possible. %
In this section, we define a clear protocol that was obeyed throughout the entire dataset annotations of \MOTNEW and \MOTLAST to guarantee consistency. 

\subsubsection{Target class}
In this benchmark, we are interested in tracking moving objects in videos. 
In particular, we are interested in evaluating multiple people tracking algorithms. Therefore, people will be the center of attention of our annotations. 
We divide the pertinent classes into three categories:
\begin{enumerate}[(i)]
    \item {\it moving} or {\it standing} pedestrians;
    \item people that are {\it not in an upright position} or artificial representations of humans; and
    \item {\it vehicles} and {\it occluders}.
\end{enumerate}

In the first group, we annotate all moving or standing (upright) pedestrians that appear in the field of view and can be determined as such by the viewer. People on bikes or skateboards will also be annotated in this category (and are typically found by modern pedestrian detectors). Furthermore, if a person \emph{briefly} bends over or squats, \eg to pick something up or to talk to a child, they shall remain in the standard \emph{pedestrian} class.
The algorithms that submit to our benchmark are expected to track these targets.

In the second group, we include all people-like objects whose exact classification is ambiguous and can vary depending on the viewer, the application at hand, or other factors. 
We annotate all static people that are not in an upright position, \eg sitting, lying down. 
We also include in this category any artificial representation of a human that might fire a detection response, such as mannequins, pictures, or reflections. People behind glass should also be marked as distractors.
The idea is to use these annotations in the evaluation such that an algorithm is neither penalized nor rewarded for tracking, \eg, a sitting person or a reflection. 

\begin{figure}
    \centering
    \includegraphics[width=0.6\linewidth]{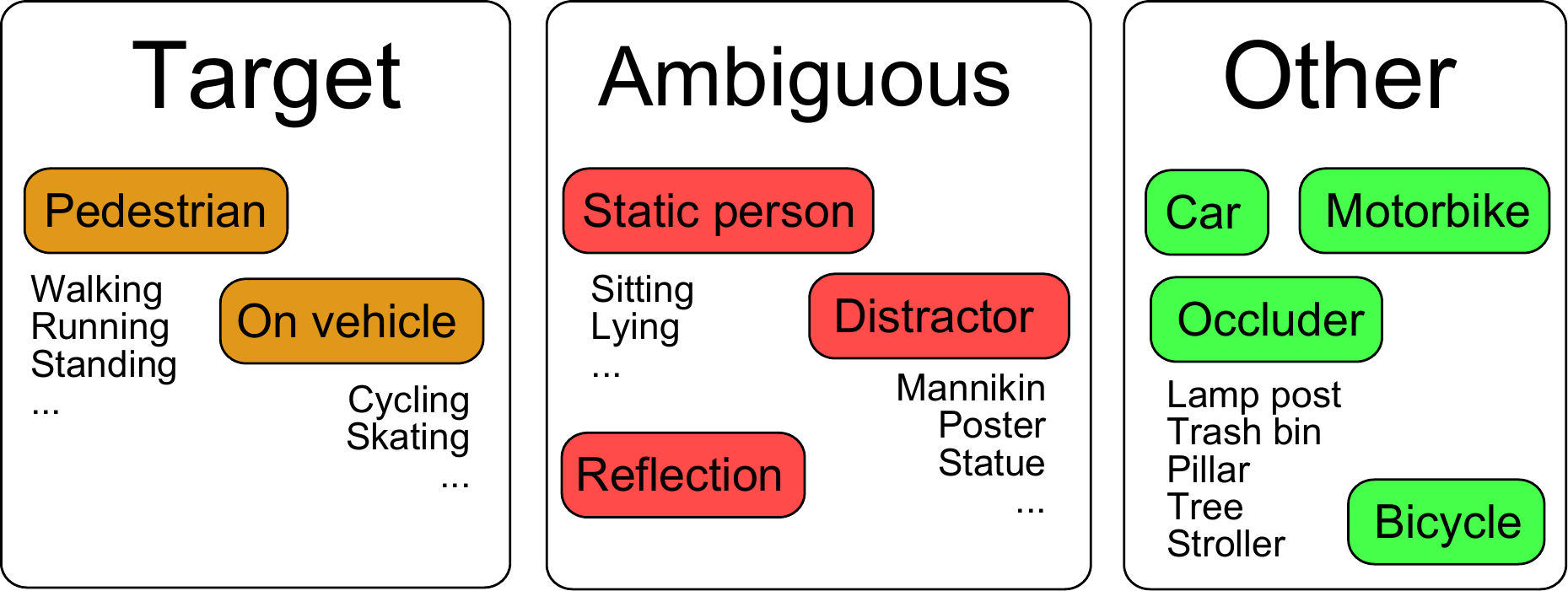}
    \includegraphics[width=0.38\linewidth]{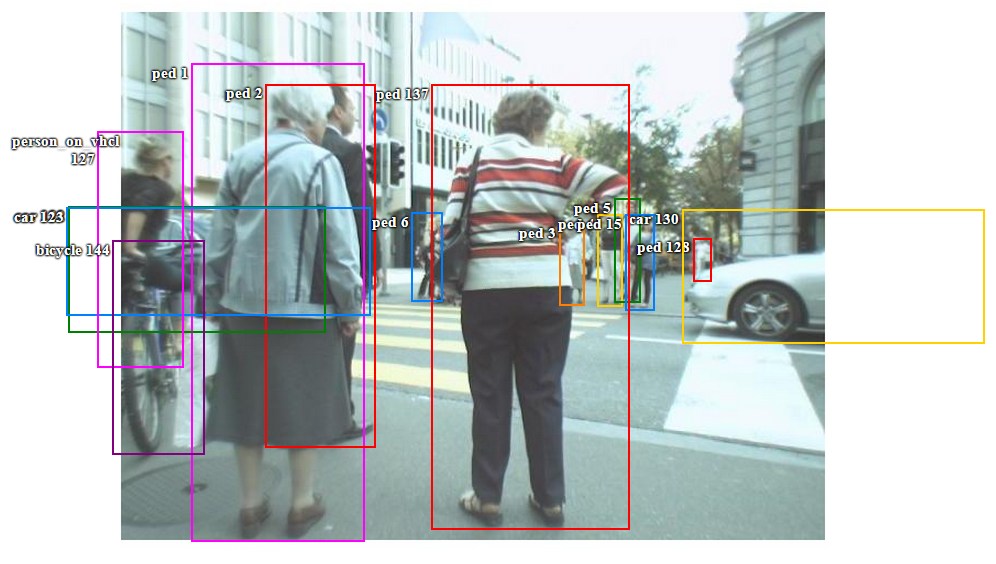}
    \caption{Left: An overview of annotated classes. The classes in orange will be the central ones to evaluate on. The red classes include ambiguous cases such that neither recovering nor missing will be penalized in the evaluation. The classes in green are annotated for training purposes and for computing the occlusion level of all pedestrians. Right: An exemplar of an annotated frame. Note how partially cropped objects are also marked outside of the frame. Also note that the bounding box encloses the entire person but not \eg the white bag of Pedestrian 1 (bottom left).}
    \label{fig:class}
\end{figure}

In the third group, we annotate all moving vehicles such as cars, bicycles, motorbikes and non-motorized vehicles (\eg strollers), as well as other potential occluders. These annotations will not play any role in the evaluation, but are provided to the users both for training purposes and for computing the level of occlusion of pedestrians. Static vehicles (parked cars, bicycles) are not annotated as long as they do not occlude any pedestrians.
The rules are summarized in \Tab~\ref{tab:instructions}, and in \Fig~\ref{fig:class} we present a diagram of the classes of objects we annotate, as well as a sample frame with annotations. 

\begin{table}[t]
\begin{tabular}{lp{.75\linewidth}}
\toprule
\textbf{What?} & \textit{Targets}: all upright people including\\
& + walking, standing, running pedestrians\\
& + cyclists, skaters\\ [1em]
& \textit{Distractors}: static people or representations\\
& + people not in upright position (sitting, lying down)\\
& + reflections, drawings or photographs of people\\
& + human-like objects like dolls, mannequins\\[1em]
& \textit{Others}: moving vehicles and other occluders\\
& + Cars, bikes, motorbikes\\
& + Pillars, trees, buildings\\
\hline
\textbf{When?} & Start as early as possible.\\
& End as late as possible.\\
& Keep ID as long as the person is inside the field of view and its path can be determined unambiguously.\\
\hline
\textbf{How?} & The bounding box should contain all pixels belonging to that person and at the same time be as tight as possible.\\
\hline
\textbf{Occlusions} & Always annotate during occlusions if the position can be determined unambiguously. \\
& If the occlusion is very long and it is not possible to determine the path of the object using simple reasoning (\eg constant velocity assumption), the object will be assigned a new ID once it reappears. \\
\bottomrule
\end{tabular}
\caption{Annotation rules.}
\label{tab:instructions}
\end{table}

\subsubsection{Bounding box alignment}
The bounding box is aligned with the object's extent as accurately as possible. 
It should contain all object pixels belonging to that instance and at the same time be as tight as possible.
This implies that a walking side-view pedestrian will typically have a box whose width varies periodically with the stride, while a front view or a standing person will maintain a more constant aspect ratio over time. If the person is partially occluded, the extent is estimated based on other available information such as expected size, shadows, 
reflections, previous and future frames and other cues. 
If a person is cropped by the image border, the box is estimated beyond the original frame to represent the entire person and to estimate the level of cropping. 
If an occluding object cannot be accurately enclosed in one box (\eg a tree with branches or an escalator may require a large bounding box where most of the area does not belong to the actual object), then several boxes may be used to better approximate the extent of that object.

Persons on vehicles are only annotated separately from the vehicle when clearly visible. 
For example, children inside strollers or people inside cars are not annotated, while motorcyclists or bikers are.

\subsubsection{Start and end of trajectories}
The box (track) appears as soon as the person's location and extent can be determined precisely. 
This is typically the case when $\approx 10 \%$ of the person becomes visible.
Similarly, the track ends when it is no longer possible to pinpoint the exact location. 
In other words, the annotation starts as early and ends as late as possible such that the accuracy is not forfeited. 
The box coordinates may exceed the visible area. 
A person leaving the field of view and re-appearing at a later point is assigned a new ID.

\subsubsection{Minimal size}
Although the evaluation will only take into account pedestrians that have a minimum height in pixels, 
annotations contain all objects of all sizes as long as they are distinguishable by the annotator. 
In other words, all targets are annotated independently of their sizes in the image.

\subsubsection{Occlusions}
There is no need to explicitly annotate the level of occlusion. This value is be computed automatically using the annotations. We leverage the assumption that for two or more overlapping bounding boxes the object with the lowest y-value of the bounding box is closest to the camera and therefore occlude the other object behind it.
Each target is fully annotated through occlusions as long as its extent and location can be determined accurately. 
If a target becomes completely occluded in the middle of a sequence and does not become visible later, the track is terminated (marked as `outside of view'). If a target reappears after a prolonged period such that its location is ambiguous during the occlusion, it is assigned a new ID.

\subsubsection{Sanity check}
Upon annotating all sequences, a ``sanity check'' is carried out to ensure that no relevant entities are missed. 
To that end, we run a pedestrian detector on all videos and add all high-confidence detections that correspond to either humans or distractors to the annotation list. 

\begin{figure}[htpb]
     \centering
     \includegraphics[width=1\linewidth]{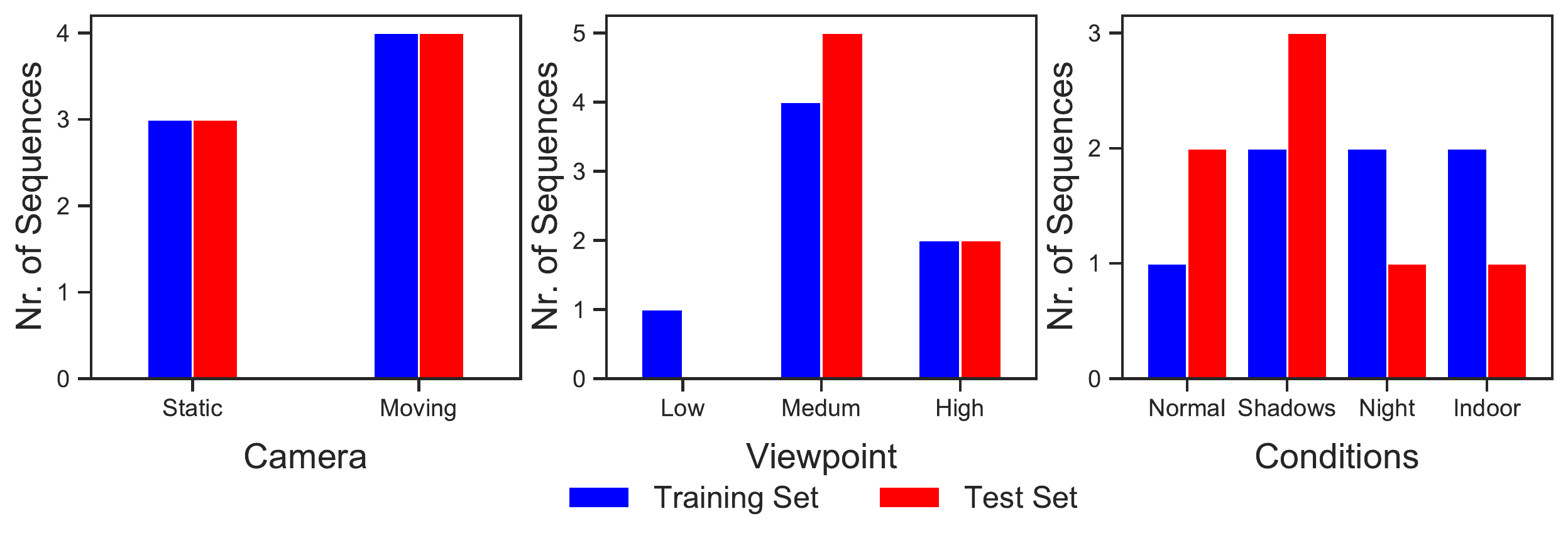}
    \caption{Comparison histogram between training and testing sequences of \MOTNEW/\MOTLAST: camera: static vs. moving camera, viewpoint: low, medium or high,  conditions: normal, shadows, night or indoor.}
    \label{fig:2Ddata16}
\end{figure}

\begin{table*}[htbp]
\begin{center}
    \caption{Overview of the types of annotations currently found in the \MOTNEW /\MOTLAST benchmark.}
\label{tab:dataclasses}
\begin{adjustbox}{max width=\textwidth}
\begin{tabular}{lcc|c|c|c|c|c|c|c|c|c|c}
 \toprule
 \multicolumn{13}{c}{\bf Annotation classes} \\ 

 Sequence & Pedestrian & Person on vehicle & Car & Bicycle & Motorbike & Vehicle (non-mot.) & Static person & 
 Distractor  & Occluder (ground) & Occluder (full)      & Refl.   & Total \\
  \midrule
MOT16/17-01 & 6,395/6,450&346&0/0&341&0&0&4,790/5,230&900&3,150/4,050&0&0/0&15,922/17,317\\ 
 MOT16/17-02 & 17,833/18,581&1,549&0/0&1,559&0&0&5,271/5,271&1,200&1,781/1,843&0&0/0&29,193/30,003\\ 
 MOT16/17-03 & 104,556/104,675&70&1,500/1,500&12,060&1,500&0&6,000/6,000&0&24,000/24,000&13,500&0/0&163,186/163,305\\ 
 MOT16/17-04 & 47,557/47,557&0&1,050/1,050&11,550&1,050&0&4,798/4,798&0&23,100/23,100&18,900&0/0&108,005/108,005\\ 
 MOT16/17-05 & 6,818/6,917&315&196/196&315&0&11&0/0&16&0/235&0&0/0&7,671/8,013\\ 
 MOT16/17-06 & 11,538/11,784&150&0/0&118&0&0&269/269&238&109/109&0&0/299&12,422/12,729\\ 
 MOT16/17-07 & 16,322/16,893&0&0/0&0&0&0&2,023/2,023&0&1,920/2,420&0&0/131&20,265/21,504\\ 
 MOT16/17-08 & 16,737/21,124&0&0/0&0&0&0&1,715/3,535&2,719&6,875/6,875&0&0/0&28,046/34,253\\ 
 MOT16/17-09 & 5,257/5,325&0&0/0&0&0&0&0/514&1,575&1,050/1,050&0&948/1,947&8,830/10,411\\ 
 MOT16/17-10 & 12,318/12,839&0&25/25&0&0&0&1,376/1,376&470&2,740/2,740&0&0/0&16,929/17,450\\ 
 MOT16/17-11 & 9,174/9,436&0&0/0&0&0&0&0/82&306&596/596&0&0/181&10,076/10,617\\ 
 MOT16/17-12 & 8,295/8,667&0&0/0&0&0&0&1,012/1,036&763&1,394/1,710&0&0/953&11,464/13,272\\ 
 MOT16/17-13 & 11,450/11,642&0&4,484/4,918&103&0&0&0/0&4&2,542/2,733&680&0/122&19,263/20,202\\ 
 MOT16/17-14 & 18,483/18,483&0&1,563/1,563&0&0&0&712/712&47&4,062/4,062&393&0/0&25,260/25,294\\ 
 \midrule 
 Total & 292,733/300,373&2,430&8,818/9,252&26,046&2,550&11&27,966/30,846&8,238&73,319/75,523&33,473&948/3,633&476,532/492,375\\ 

 \bottomrule
 \end{tabular}
 \end{adjustbox}
  \end{center}

\end{table*}

\begin{table*}[tbp]
    \begin {center} %
        \caption{Overview of the sequences currently included in the \MOTNEW /\MOTLAST benchmark.}
    \label{tab:dataoverview16}
    \tiny %
   \begin{adjustbox}{max width=\textwidth}
    \begin{tabular}{l| c| c| r| r| r| r| c |c | c}
    \toprule %
    \multicolumn{10}{c}{\bf Training sequences} \\  %
    Name & FPS & Resolution & Length & Tracks & Boxes & Density & Camera & Viewpoint & Conditions \\ %
    \midrule %
    MOT16/17-02 (new) & 30 & 1920x1080 & 600 (00:20) & 54/62 & 17,833/18,581 & 29.7/31.0  & static  & medium & cloudy \\
    MOT16/17-04 (new) & 30 & 1920x1080 & 1,050 (00:35) & 83/83 & 47,557/47557 & 45.3/45.3  & static & high & night  \\
    MOT16/17-05~\citep{Ess08cvpr} & 14 & 640x480 & 837 (01:00) & 125/133 & 6,818/6,917 & 8.1/8.3 & moving & medium & sunny \\
    MOT16/17-09 (new) & 30 & 1920x1080 & 525 (00:18) & 25/26 & 5,257/5,325 & 10.0/10.1  & static & low & indoor	\\
    MOT16/17-10 (new) & 30 & 1920x1080 & 654 (00:22) & 54/57 & 12,31812,839 & 18.8/19.6  & moving & medium & night \\
    MOT16/17-11 (new) & 30 & 1920x1080 & 900 (00:30) & 69/75 & 9,174/9,436 & 10.2/10.5 & moving & medium & indoor \\
    MOT16/17-13 (new) & 25 & 1920x1080 & 750 (00:30) & 107/110 & 11,450/11,642 & 15.3/15.5  & moving & high & sunny  \\
    \midrule %
    \multicolumn{3}{c|}{\bf Total training} & {\bf 5,316 (03:35)} & {\bf 517}/{\bf 546}& {\bf 110,407}/{\bf 112,297} & {\bf 20.8}/{\bf 21.1} & & &    \\
    \bottomrule
    %
    %
    \multicolumn{10}{c}{\vspace{1em}} \\ %
    \toprule %
    \multicolumn{10}{c}{\bf Testing sequences} \\ %
    Name & FPS & Resolution & Length & Tracks & Boxes & Density & Camera & Viewpoint & Conditions \\ %
    \midrule %
    MOT16/17-01 (new) & 30 & 1920x1080 & 450 (00:15) & 23/24 & 6,395/6,450 & 14.2/14.3  & static  & medium & cloudy \\
    MOT16/17-03 (new) & 30 & 1920x1080 & 1,500 (00:50) & 148/148 & 104,556/104,675 & 69.7/69.8  & static & high & night \\
    MOT16/17-06~\citep{Ess08cvpr} & 14 & 640x480 & 1,194 (01:25) & 221/222 & 11,538/11,784 & 9.7/9.9 & moving & medium & sunny  \\
    MOT16/17-07 (new) & 30 & 1920x1080 & 500 (00:17) & 54/60 & 16,322/16.893 & 32.6/33.8  & moving & medium & shadow \\
    MOT16/17-08 (new) & 30 & 1920x1080 & 625 (00:21) & 63/76 & 16,737/21,124 & 26.8/33.8  & static & medium & sunny \\
    MOT16/17-12 (new) & 30 & 1920x1080 & 900 (00:30) & 86/91 & 8,295/8,667 & 9.2/9.6 & moving & medium & indoor \\
    MOT16/17-14 (new) & 25 & 1920x1080 & 750 (00:30) &164/164 & 18,483/18,483 & 24.6/24.6  & moving & high & sunny \\
    \midrule %
    \multicolumn{3}{c|}{\bf Total testing} & {\bf 5,919 (04:08)} & {\bf 759}/{\bf 785} & {\bf 182,326}/{\bf 188,076} & {\bf 30.8}/{\bf 31.8} & & &  \\
    \bottomrule %
    \end{tabular} %
    \end{adjustbox}
    \end{center} %

\end{table*}
\begin{table*}[hbt]
    \begin {center}
        \caption{Detection bounding box statistics.}
    \label{tab:detMOT16}
    \tiny
    \begin{adjustbox}{max width=\linewidth}
    \begin{tabular}{l |r r | r r|r r | r r }
    \toprule 
      & \multicolumn{2}{c}{MOT16} & \multicolumn{6}{|c}{MOT17} \\
      \midrule
    & \multicolumn{2}{c}{DPM} & \multicolumn{2}{|c}{DPM} & \multicolumn{2}{c}{FRCNN} &  \multicolumn{2}{c}{SDP} \\
    
    \bf Seq & \bf nDet. & \bf nDet./fr. & \bf nDet. & \bf nDet./fr. & \bf nDet. & \bf nDet./fr. & \bf nDet. & \bf nDet./fr.  \\
    \midrule 
MOT16/17-01 &3,775 & 8.39& 3,775 & 8.39&5,514& 12.25&5,837& 12.97 \\ 
MOT16/17-02 &7,267 & 12.11& 7,267 & 12.11&8,186& 13.64&11,639& 19.40 \\ 
MOT16/17-03 &85,854 & 57.24& 85,854 & 57.24&65,739& 43.83&80,241& 53.49 \\ 
MOT16/17-04 &39,437 & 37.56& 39,437 & 37.56&28,406& 27.05&37,150& 35.38 \\ 
MOT16/17-05 &4,333 & 5.20& 4,333 & 5.20&3,848& 4.60&4,767& 5.70 \\ 
MOT16/17-06 &7,851 & 6.58& 7,851 & 6.58&7,809& 6.54&8,283& 6.94 \\ 
MOT16/17-07 &11,309 & 22.62& 11,309 & 22.62&9,377& 18.75&10,273& 20.55 \\ 
MOT16/17-08 &10,042 & 16.07& 10,042 & 16.07&6,921& 11.07&8,118& 12.99 \\ 
MOT16/17-09 &5,976 & 11.38& 5,976 & 11.38&3,049& 5.81&3,607& 6.87 \\ 
MOT16/17-10 &8,832 & 13.50& 8,832 & 13.50&9,701& 14.83&10,371& 15.86 \\ 
MOT16/17-11 &8,590 & 9.54& 8,590 & 9.54&6,007& 6.67&7,509& 8.34 \\ 
MOT16/17-12 &7,764 & 8.74& 7,764 & 8.74&4,726& 5.32&5,440& 6.09 \\ 
MOT16/17-13 &5,355 & 7.22& 5,355 & 7.22&8,442& 11.26&7,744& 10.41 \\ 
MOT16/17-14 &8,781 & 11.71& 8,781 & 11.71&10,055& 13.41&10,461& 13.95 \\ 
\midrule 
 Total &  215,166 & 19.19&215,166& 19.19&177,780& 15.84&211,440& 18.84 \\ 
    \bottomrule 
    \end{tabular}
    \end{adjustbox}
    \end{center}

\end{table*}

\subsection{Data format}
\label{sec:data-format16}

All images were converted to JPEG and named sequentially to a 6-digit file name (\eg~000001.jpg). Detection and annotation files are simple comma-separated value (CSV) files. 
Each line represents one object instance and contains 9 values as shown in~\Tab~\ref{tab:dataformat16}.

The first number indicates in which frame the object appears, while the second number identifies that object as belonging to a trajectory by assigning a unique ID (set to $-1$ in a detection file, as no ID is assigned yet). 
Each object can be assigned to only one trajectory.
The next four numbers indicate the position of the bounding box of the pedestrian in 2D image coordinates. 
The position is indicated by the top-left corner as well as the width and height of the bounding box.
This is followed by a single number, which in the case of detections denotes their confidence score.
The last two numbers for detection files are ignored (set to -1).

\begin{table*}[htpb]
    \begin {center}
    \tiny
    \begin{tabular}{c | c| p{13cm}}
    \toprule
    \bf Position & \bf Name & \bf Description\\ 
    \hline 
    1 & Frame number & Indicate at which frame the object is present\\
    2 & Identity number & Each pedestrian trajectory is identified by a
    unique ID ($-1$ for detections)\\
    3 & Bounding box left &  Coordinate of the top-left corner of the pedestrian bounding box\\
    4 & Bounding box top & Coordinate of the top-left corner of the pedestrian bounding box \\
    5 & Bounding box width & Width in pixels of the pedestrian bounding box\\
    6 & Bounding box height & Height in pixels of the pedestrian bounding box\\
    7 & Confidence score & DET: Indicates how confident the detector is that this instance is a pedestrian. \hspace{5cm} GT: It acts as a flag whether the entry is to be considered (1) or ignored (0).  \\
    8 &  Class & GT: Indicates the type of object annotated  \\
    9 & Visibility & GT: Visibility ratio, a number between 0 and 1 that says how much of that object is visible. Can be due to occlusion and due to image border cropping. \\
    \bottomrule 
    \end{tabular}
    \end{center}
    \caption{Data format for the input and output files, both for detection (DET) and annotation/ground truth (GT) files.}
    \label{tab:dataformat16}
\end{table*}
\begin{table}[hbt]
    \begin {center}
    \tiny
    \begin{tabular}{l | c}
    \toprule 
    \bf Label & \bf ID\\ 
    \hline 
    Pedestrian & 1 \\
    Person on vehicle & 2 \\
    Car & 3 \\
    Bicycle & 4 \\
    Motorbike & 5 \\
    Non motorized vehicle & 6 \\
    Static person & 7 \\
    Distractor & 8 \\
    Occluder & 9 \\
    Occluder on the ground & 10 \\
    Occluder full & 11 \\
    Reflection & 12 \\
    \bottomrule 
    \end{tabular}
    \end{center}
    \caption{Label classes present in the annotation files and ID appearing in the 7$^\text{th}$ column of the files as described in \Tab~\ref{tab:dataformat16}.}
    \label{tab:labelclass}
\end{table}

\noindent An example of such a 2D detection file is:

\begin{samepage}
    \begin{center}
    \begin{footnotesize}
    \texttt{1, -1, 794.2, 47.5, 71.2, 174.8, 67.5, -1, -1}\nopagebreak\\
    \texttt{1, -1, 164.1, 19.6, 66.5, 163.2, 29.4, -1, -1}\nopagebreak\\
    \texttt{1, -1, 875.4, 39.9, 25.3, 145.0, 19.6, -1, -1}\nopagebreak\\
    \texttt{2, -1, 781.7, 25.1, 69.2, 170.2, 58.1, -1, -1}\nopagebreak\\
    \end{footnotesize}
    \end{center}
\end{samepage}

\noindent For the ground truth and result files, the 7$^\text{th}$ value (confidence score) acts as a flag whether the entry is to be considered. 
A value of 0 means that this particular instance is ignored in the evaluation, while a value of 1 is used to mark it as active. 
The 8$^\text{th}$ number indicates the type of object annotated, following the convention of~\Tab~\ref{tab:labelclass}. 
The last number shows the visibility ratio of each bounding box. 
This can be due to occlusion by another static or moving object, or to image border cropping.
 
\noindent An example of such an annotation 2D file is:

\begin{samepage}
    \begin{center}
    \begin{footnotesize}
    \texttt{1, 1, 794.2, 47.5, 71.2, 174.8,  1,  1, 0.8}\nopagebreak\\
    \texttt{1, 2, 164.1, 19.6, 66.5, 163.2,  1,  1, 0.5}\nopagebreak\\
    \texttt{2, 4, 781.7, 25.1, 69.2, 170.2, 0, 12, 1.}\nopagebreak\\
    \end{footnotesize}
    \end{center}
\end{samepage}

\noindent In this case, there are 2 pedestrians in the first frame of the sequence, with identity tags 1, 2. 
In the second frame, we can see a reflection (class 12), which is to be considered by the evaluation script and will neither count as a false negative nor as a true positive, independent of whether it is correctly recovered or not.
All values including the bounding box are 1-based, \ie the top left corner corresponds to $(1,1)$.

To obtain a valid result for the entire benchmark, a separate CSV file following the format described above must be created for each sequence and called \newline
\texttt{``Sequence-Name.txt''}. All files must be compressed into a single ZIP file that can then be uploaded to be evaluated.

\section{Implementation details of the evaluation}
\label{sec:eval_appendix}

In this section, we detail how to compute false positives, false negatives, and identity switches, which are the basic units for the evaluation metrics presented in the main paper.
We also explain how the evaluation deals with special non-target cases: people behind a window or sitting people.

\subsection{Tracker-to-target assignment}
\label{sec:tracker-assignment}
There are two common prerequisites for quantifying the performance of a tracker. 
One is to determine for each hypothesized output, whether it is a true positive (TP) that describes an actual (annotated) target, or whether the output is a false alarm (or false positive, FP). 
This decision is typically made by thresholding based on a defined distance (or dissimilarity) measure $\dismeas$ between the coordinates of the true and predicted box placed around a target (see~\Sec~\ref{sec:distance-measure}). 
A target that is missed by any hypothesis is a false negative (FN). 
A good result is expected to have as few FPs and FNs as possible. 
Next to the absolute numbers, we also show the false positive ratio measured by the number of false alarms per frame (FAF), sometimes also referred to as false positives per image (FPPI) in the object detection literature.

%
\begin{figure*}[t]
\centering
\def\svgwidth{1\linewidth}
\input{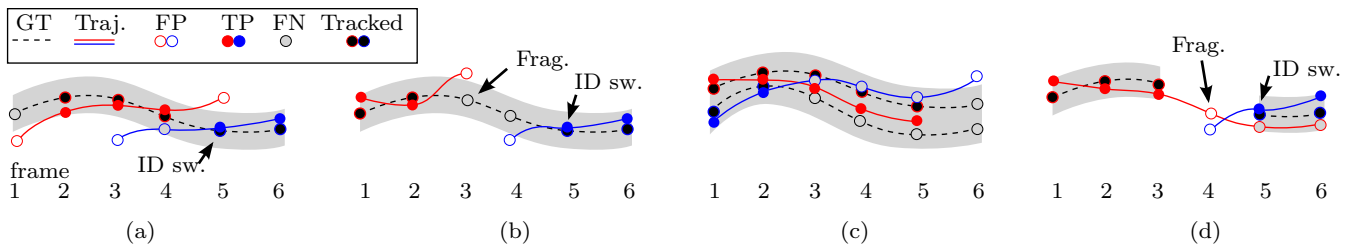}
\caption
{
Four cases illustrating tracker-to-target assignments. 
(a) An ID switch occurs when the mapping switches from the previously assigned red track 
to the blue one. 
(b) A track fragmentation is counted in frame 3 because the target is tracked in frames 1-2, then interrupts, and then reacquires its `tracked' status at a later point. A new (blue) track hypothesis also causes an ID switch at this point. 
(c) Although the tracking results are reasonably good an optimal single-frame assignment in frame 1 is propagated through the sequence, causing 5 missed targets (FN) and 4 
false positives (FP). Note that no fragmentations are counted in frames 
3 and 6 because tracking of those targets is not resumed at a later point. 
(d) A degenerate case illustrating that target re-identification is not handled correctly. 
An interrupted ground-truth trajectory will typically cause a fragmentation. 
Also note the less intuitive ID switch, which is counted because blue is the closest target in frame 5 that is not in conflict with the mapping in frame 4. 
}
\label{fig:mapping}
\end{figure*}

The same target may be covered by multiple outputs. 
The second prerequisite before computing the numbers is then to establish the correspondence between all annotated and hypothesized objects under the constraint that a true object should be recovered at most once, and that one hypothesis cannot account for more than one target. 

For the following, we assume that each ground-truth trajectory has one unique start and one unique endpoint, \ie, that it is not fragmented. 
Note that the current evaluation procedure does not explicitly handle target re-identification. 
In other words, when a target leaves the field-of-view and then reappears, it is treated as an unseen target with a new ID. 
As proposed in~\citep{Stiefelhagen06CLE}, the optimal matching is found using Munkres (a.k.a.~Hungarian) algorithm. 
However, dealing with video data, this matching is not performed independently for each frame, but rather considering a temporal correspondence. 
More precisely, if a ground-truth object $i$ is matched to hypothesis $j$ at time $t-1$ \emph{and} the distance (or dissimilarity) between $i$ and $j$ in frame $t$ is below $\simthresh$, then the correspondence between $i$ and $j$ is carried over to frame $t$ even if there exists another hypothesis that is closer to the actual target. 
A mismatch error (or equivalently an identity switch, IDSW) is counted if a ground-truth target $i$ is matched to track $j$ and the last known assignment was $k \ne j$. 
Note that this definition of ID switches is more similar to~\citep{LiCVPR2009} and stricter than the original one~\citep{Stiefelhagen06CLE}. 
Also note that, while it is certainly desirable to keep the number of ID switches low, their absolute number alone is not always expressive to assess the overall performance, but should rather be considered concerning the number of recovered targets. 
The intuition is that a method that finds twice as many trajectories will almost certainly produce more identity switches. 
For that reason, we also state the relative number of ID switches, which is computed as IDSW / Recall.

These relationships are illustrated in~\Fig~\ref{fig:mapping}. 
For simplicity, we plot ground-truth trajectories with dashed curves, and the tracker output with solid ones, where the color represents a unique target ID. 
The grey areas indicate the matching threshold (see \Sec~\ref{sec:target_like_annotations}). 
Each true target that has been successfully recovered in one particular frame is represented with a filled black dot with a stroke color corresponding to its matched hypothesis. False positives and false negatives are plotted as empty circles. 
See figure caption for more details. 

After determining true matches and establishing correspondences it is now possible to compute the metrics. 
We do so by concatenating all test sequences and evaluating the entire benchmark. 
This is in general more meaningful than averaging per-sequences figures because of the large variation on the number of targets per sequence.

\subsection{Distance measure}
\label{sec:distance-measure}

The relationship between ground-truth objects and a tracker output is established using bounding boxes on the image plane. 
Similar to object detection~\citep{Everingham15ijcv}, the intersection over union (a.k.a. the Jaccard index) is usually employed as the similarity criterion, while the threshold $\simthresh$ is set to 
$0.5$ or $50\%$.

\subsection{Target-like annotations}
\label{sec:target_like_annotations}
People are a common object class present in many scenes, but should we track all people in our benchmark?
For example, should we track static people sitting on a bench? Or people on bicycles? How about people behind a glass?  
We define the target class of \MOTNEW and \MOTLAST as \textit{all upright people, standing or walking, that are reachable along the viewing ray without a physical obstacle.} 
For instance, reflections or people behind a transparent wall or window are excluded.
We also exclude from our target class people on bicycles (riders) or other vehicles. 

\begin{figure}
    \centering
    \includegraphics[height=1.6cm]{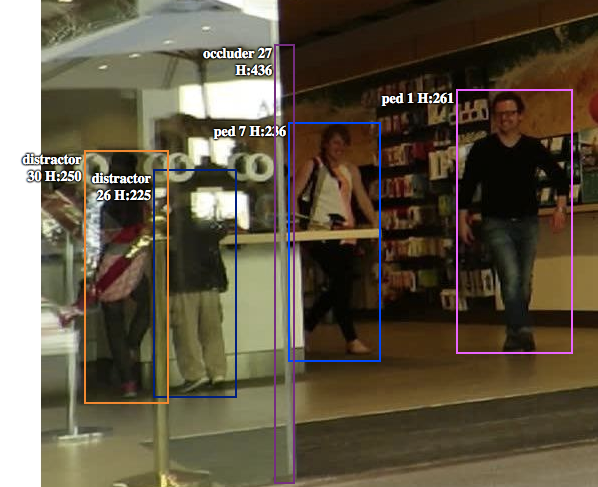}
    \includegraphics[height=1.6cm]{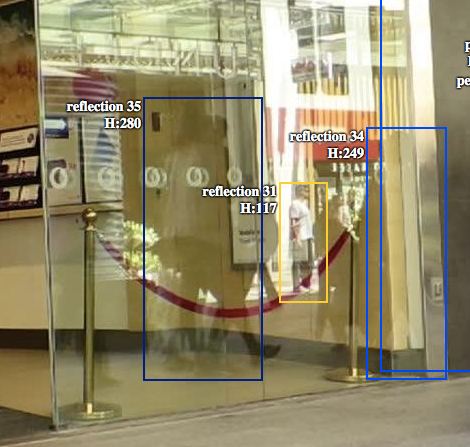}
    \includegraphics[height=1.6cm]{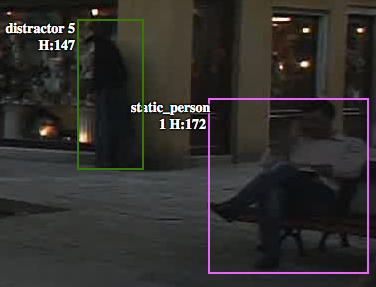}
    \includegraphics[height=1.6cm]{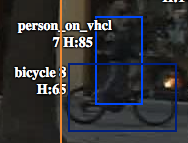}
    \caption{The annotations include different classes of objects similar to the target class, a pedestrian in our case. We consider these special classes (distractor, reflection, static person and person on vehicle) to be so similar to the target class that a tracker should neither be penalized nor rewarded for tracking them in the sequence.}
    \label{fig:distractors}
\end{figure}

For all these cases where the class is very similar to our target class (see Fig.~\ref{fig:distractors}), we adopt a similar strategy as in~\citep{Mathias14ECCV}. 
That is, a method is neither penalized nor rewarded for tracking or not tracking those similar classes. 
Since a detector is likely to fire in those cases, we do not want to penalize a tracker with a set of false positives for properly following that set of detections, \ie, of a person on a bicycle. 
Likewise, we do not want to penalize with false negatives a tracker that is based on motion cues and therefore does not track a sitting person. 

To handle these special cases, we adapt the tracker-to-target assignment algorithm to perform the following steps:

\begin{enumerate}
    \item At each frame, all bounding boxes of the result file are matched to the ground truth via the Hungarian algorithm.
    \item All result boxes that overlap more than the matching threshold ($>50\%$) with one of these classes (distractor, static person, reflection, person on vehicle) excluded from the evaluation.
    \item During the final evaluation, {\it only} those boxes that are annotated as {\it pedestrians} are used.
\end{enumerate}

\end{document}